\documentclass[9pt, a4paper, logo, twocolumn, copyright]{deepmind}

\pdftrailerid{redacted}

\makeatletter
\renewcommand\bibentry[1]{\nocite{#1}{\frenchspacing\@nameuse{BR@r@#1\@extra@b@citeb}}}
\makeatother

\newcommand{\printauthorlist}{{%
  \expandafter\let\csname \endcsname\@gobble
  \AB@authlist}
}

\usepackage{dsfont}
\usepackage{dm-colors}
\usepackage{subcaption}
\hypersetup{
    colorlinks=true,
    allcolors=dmblue500,
    pdftitle={Open-Ended learning leads to Generally Capable Agents},
    pdfpagemode=FullScreen,
    }

\usepackage{mathtools}
\usepackage{dsfont}
\usepackage[dvipsnames]{xcolor}
\usepackage{booktabs}
\usepackage{xfrac}
\usepackage{bbm}

\usepackage{apxproof}

\usepackage{algorithm}
\usepackage{algorithmicx}

\usepackage[most]{tcolorbox}
\usepackage{xparse}
\usepackage{lipsum}
\usepackage{changepage}
\usepackage{enumitem}
\usepackage{wrapfig}
\usepackage[export]{adjustbox}

\usepackage{caption}
\newcommand{\squishlist}{
   \begin{list}{$\bullet$}
    { \setlength{\itemsep}{0pt}      \setlength{\parsep}{3pt}
      \setlength{\topsep}{3pt}       \setlength{\partopsep}{0pt}
      \setlength{\leftmargin}{1.5em} \setlength{\labelwidth}{1em}
      \setlength{\labelsep}{0.5em} } }

\newcommand{\squishlisttwo}{
   \begin{list}{$\bullet$}
    { \setlength{\itemsep}{0pt}    \setlength{\parsep}{0pt}
      \setlength{\topsep}{0pt}     \setlength{\partopsep}{0pt}
      \setlength{\leftmargin}{2em} \setlength{\labelwidth}{1.5em}
      \setlength{\labelsep}{0.5em} } }

\newcommand{\squishend}{
    \end{list}  }

\newtheoremrep{thm}{Theorem}[section]
\newtheoremrep{prop}{Proposition}[section]
\newtheoremrep{cor}{Corollary}[section]
\newtheoremrep{obs}{Observation}[section]
\newtheoremrep{defn}{Definition}[section]

\DeclareMathAlphabet{\mathpzc}{OT1}{pzc}{m}{n}

\newcommand{\xlandspace}{\aleph}
\newcommand{\game}{\mathbf{G}}
\newcommand{\goal}{\mathbf{g}}
\newcommand{\option}{\widehat{\mathbf{g}}}
\newcommand{\world}{\mathbf{w}}
\newcommand{\gamespace}{\mathfrak{G}}
\newcommand{\goalspace}{\mathcal{G}}
\newcommand{\worldspace}{\mathcal{W}}
\newcommand{\valuehead}{\mathbf{v}}
\newcommand{\goatvaluehead}{\widehat{\mathbf{v}}}
\newcommand{\policy}{\pi}
\newcommand{\policyspace}{\Pi}
\newcommand{\population}{\boldsymbol{\Pi}}
\newcommand{\xlandtask}{\mathbf{x}}
\newcommand{\statespace}{\mathcal{S}}
\newcommand{\state}{\mathbf{s}}
\newcommand{\observation}{\mathbf{o}}
\newcommand{\observationhistory}{\mathbf{h}}

\newcommand{\predicate}{\phi}

\newcommand{\ed}{\kappa}
\newcommand{\coop}{\mathrm{coop}}
\newcommand{\comp}{\mathrm{comp}}
\newcommand{\balance}{\mathrm{bal}}

\newcommand{\reward}{r}
\newcommand{\return}{R}

\newcommand{\normperf}{\widehat{\mathrm{perf}}}
\newcommand{\perf}{\mathrm{perf}}
\newcommand{\perc}{\mathrm{perc}}
\newcommand{\normaliser}{\mathrm{norm}}

\newcommand{\goatattention}{\mathrm{att}}

\newcommand{\goatvalue}{f_\mathbf{V}}
\newcommand{\stopgradient}[1]{\left \llbracket #1 \right \rrbracket}

\newcommand{\evaltrain}{validation}
\newcommand{\evalvalid}{test}

\def\figref#1{Figure~\ref{#1}}
\def\secref#1{Section~\ref{#1}}

\usepackage[authoryear, sort&compress, round]{natbib} 

\graphicspath{{figures/}}
\title{Open-Ended Learning Leads to Generally Capable Agents}
\reportnumber{} 

\author{{\bf Open-Ended Learning Team}*} 
\author{Adam Stooke}
\author{Anuj Mahajan}
\author{Catarina Barros}
\author{Charlie Deck}
\author{Jakob Bauer}
\author{Jakub Sygnowski}
\author{Maja Trebacz}
\author{Max Jaderberg}
\author{Michael Mathieu}
\author{Nat McAleese}
\author{Nathalie Bradley-Schmieg}
\author{Nathaniel Wong}
\author{Nicolas Porcel}
\author{Roberta Raileanu}
\author{Steph Hughes-Fitt}
\author{Valentin Dalibard}
\author{Wojciech Marian Czarnecki}

\affil{\textbf{DeepMind, London, UK}}

\banner{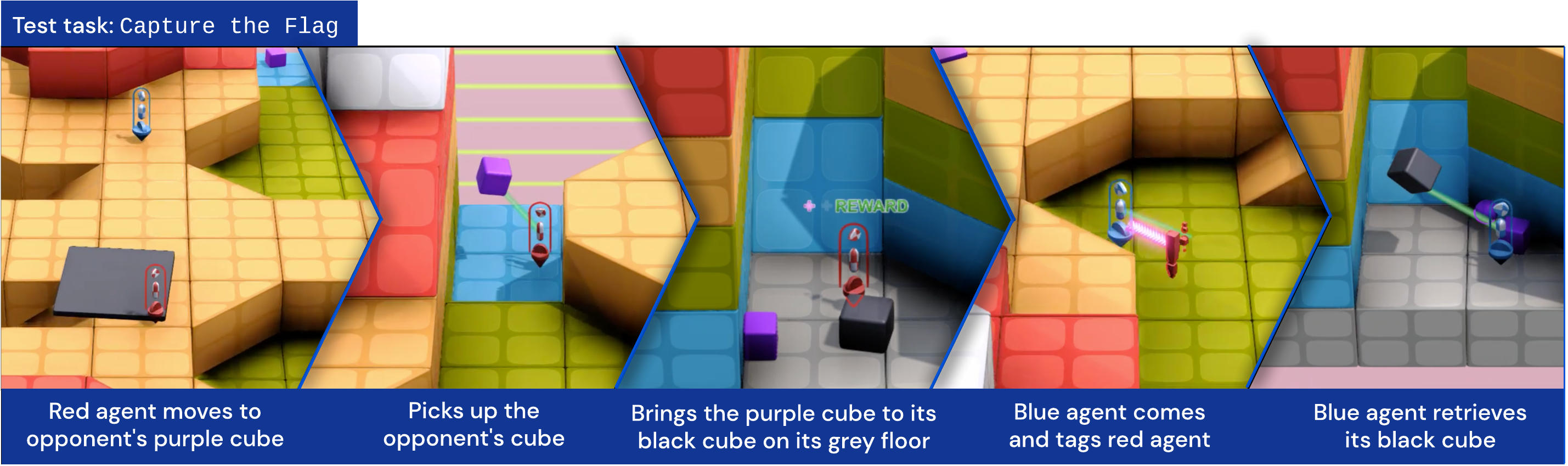}{Example zero-shot behaviour of an agent playing a Capture the Flag task at test time. The agent has trained on 700k games, but has never experienced any Capture the Flag games before in training. 
The red player's goal is to put both the purple cube (the opponent's cube) and the black cube (its own cube) onto its base (the grey floor), while the blue player tries to put them on the blue floor -- the cubes are used as flags.
The red player finds the opponent's cube, brings it back to its cube at its base, at which point reward is given to the agent.
Shortly after, the opponent, played by another agent, tags the red player and takes the cube back.
}

\begin{abstract}
Artificial agents have achieved great success in individual challenging simulated environments, mastering the particular tasks they were trained for, with their behaviour even generalising to maps and opponents that were never encountered in training. 
In this work we create agents that can perform well beyond a single, individual task, that exhibit much wider generalisation of behaviour to a massive, rich space of challenges.
We define a universe of tasks within an environment domain and demonstrate the ability to train agents that are generally capable across this vast space and beyond. 
The environment is natively multi-agent, spanning the continuum of competitive, cooperative, and independent games, which are situated within procedurally generated physical 3D worlds.
The resulting space is exceptionally diverse in terms of the challenges posed to agents, and as such, even measuring the learning progress of an agent is an open research problem.
We propose an iterative notion of improvement between successive generations of agents, rather than seeking to maximise a singular objective, allowing us to quantify progress despite tasks being incomparable in terms of achievable rewards. 
Training an agent that is performant across such a vast space of tasks is a central challenge, one we find that pure reinforcement learning on a fixed distribution of training tasks does not succeed in.
We show that through constructing an open-ended learning process, which dynamically changes the training task distributions and training objectives such that the agent never stops learning, we achieve consistent learning of new behaviours.
The resulting agent is able to score reward in every one of our humanly solvable evaluation levels, with behaviour generalising to many held-out points in the universe of tasks.
Examples of this zero-shot generalisation include good performance on Hide and Seek, Capture the Flag, and Tag.
Through analysis and hand-authored probe tasks we characterise the behaviour of our agent, and find interesting emergent heuristic behaviours such as trial-and-error experimentation, simple tool use, option switching, and cooperation.
Finally, we demonstrate that the general capabilities of this agent could unlock larger scale transfer of behaviour through cheap finetuning. 
A summary \href{https://deepmind.com/blog/article/generally-capable-agents-emerge-from-open-ended-play}{blog post can be found here} and a \href{https://youtu.be/lTmL7jwFfdw}{video catalogue of results here}. 
\end{abstract}

\begin{document}

\maketitle

\begin{figure*}[ht]
    \centering
    \begin{tabular}{cc}
    \includegraphics[width=\linewidth]{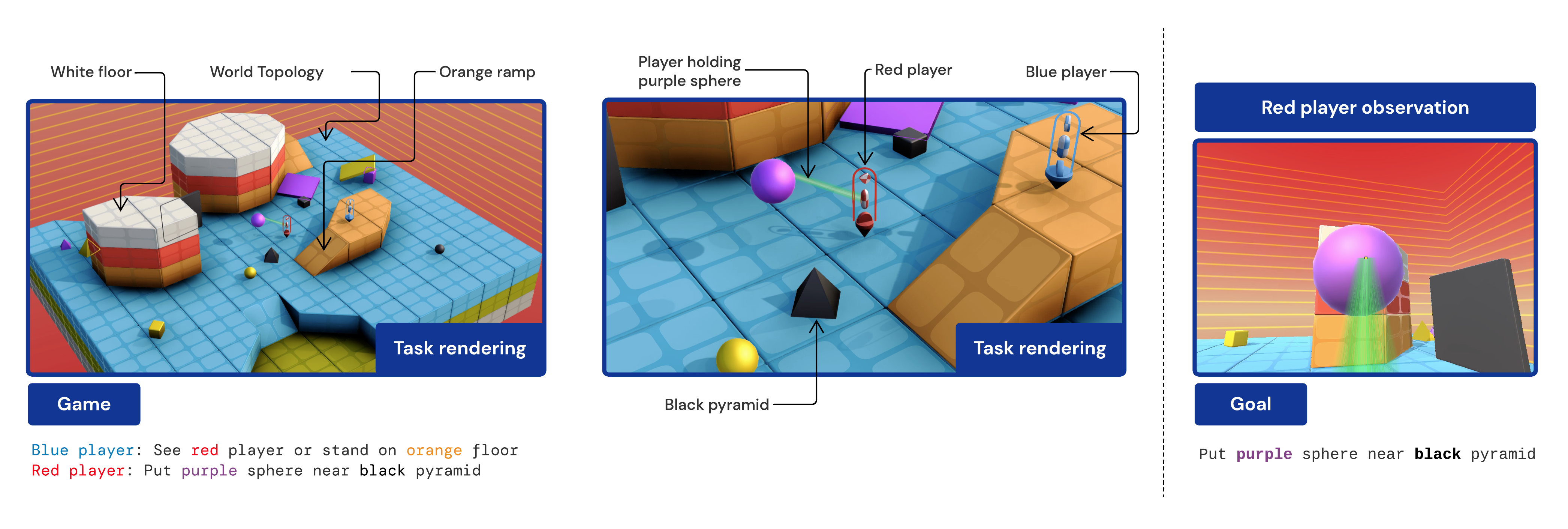}
    \end{tabular}
    \caption{\textbf{(Left \& Center)} An instance of a \emph{task} within the XLand environment space, composed of the \emph{world} -- the layout of the topology, initial object and player positions, and player gadgets -- as well as the \emph{game} -- the specification of rewarding states for each player in this task. \textbf{(Right)} The observation of the red player consisting of the first-person view and the goal of the player.}
    \label{fig:unity}
\end{figure*}

\section{Introduction}
\let\thefootnote\relax\footnotetext{*Authors ordered alphabetically by first name. More details in \hyperref[sec:author]{Author Contributions}. Correspondence to \texttt{jaderberg@deepmind.com}}
Over recent years, deep reinforcement learning (deep RL) has repeatedly yielded highly performant artificial agents across a range of training domains~\citep{silver2017mastering,rubiks_cited,mirhoseini2021graph}. The marriage of expressive neural network architectures, together with scalable and general reinforcement learning algorithms to train these networks, has resulted in agents that can outperform humans on the complex simulated games they were trained on~\citep{mnih2015human}. In addition, through \emph{multi-agent} deep RL, agents have also demonstrated impressive robustness to held-out opponents -- opponents that were never encountered during training~\citep{jaderberg2019human}. Some of the most salient examples include robustness to the top human professional players~\citep{silver2016mastering,vinyals2019grandmaster,berner2019dota}. However, these agents are often constrained to play only the games they were trained for -- whilst the exact instantiation of the game may vary (\emph{e.g.} the layout, initial conditions, opponents) the goals the agents must satisfy remain the same between training and testing. Deviation from this can lead to catastrophic failure of the agent.

In this work we move towards creating an artificial agent whose behaviour generalises beyond the set of games it was trained on, an agent which is robust and generally capable across a vast evaluation space of games. By training an agent effectively across a massively multi-task continuum we obtain a neural network policy that exhibits general heuristic behaviours, allowing it to score reward in all humanly solvable tasks in our held-out evaluation task set. In addition, we see the agent being capable in tasks that not only are explicitly held-out from training, but lie far outside of its training distributions, including versions of hide and seek~\citep{baker2019emergent} and capture the flag~\citep{jaderberg2019human}.

To produce a vast and diverse continuum of training and evaluation tasks we develop an environment space, dubbed \emph{XLand}, that permits procedural generation of rich 3D worlds and multiplayer games (described by the goals of the players). These span both two- and three-player tasks, highly competitive and completely cooperative as well as mixtures of both, balanced and imbalanced games, and strategically deep games (\emph{e.g.} Capture the Flag or XRPS, see \secref{sec:game-diversity}). The capabilities asked of players include visual scene understanding, navigation, physical manipulation, memory, logical reasoning, and theory of mind. 

To train agents in this environment space, we first define a multi-dimensional measure of performance, \emph{normalised score percentiles}, which characterises agent performance and robustness across the evaluation task space. We create an open-ended training process to iteratively improve the spectrum of normalised score percentiles. The training process uses deep RL at its core with an attention-based neural network architecture allowing implicit modelling of goals of the game which are provided to the agent. The training tasks consumed by the agent are dynamically generated in response to the agent's performance, with the generating function constantly changing to keep a population of agents improving across all percentiles of normalised score. This population training is repeated multiple times sequentially, each generation of agents bootstrapping their performance from previous generations with policy distillation, each generation of agents contributing new policies to train against in this multiplayer environment, and each generation redefining the normalised score percentiles as the frontier of performance across task space is advanced. From experimental results we demonstrate the clear benefit of each component of this learning process, with the dynamic task generation being particularly important for learning compared to uniform sampling from task space.

\begin{figure*}[ht!]
\includegraphics[width=\linewidth]{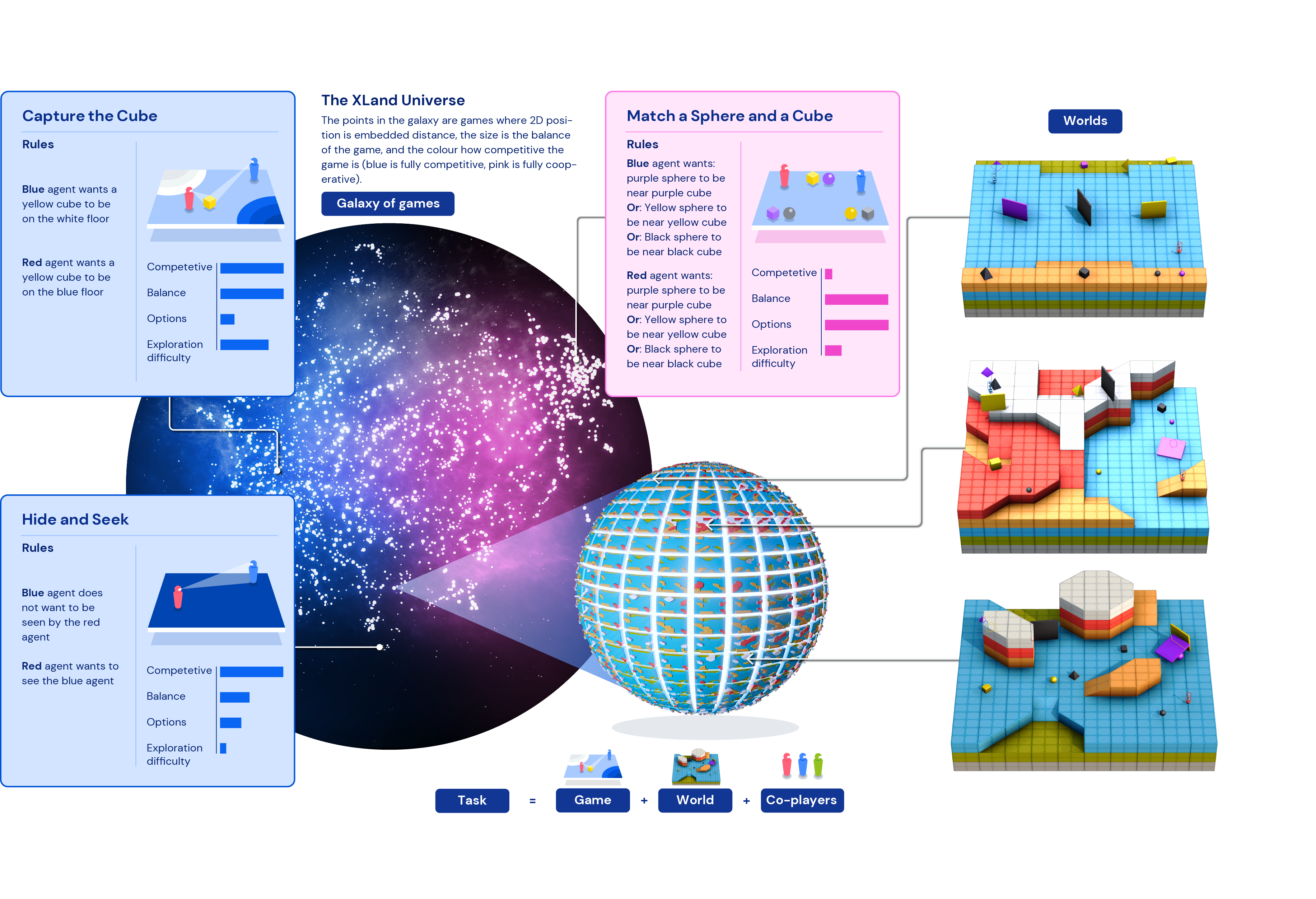}
\caption{Visualisation of the XLand environment space. \textbf{(Left)} Each dot corresponds to a single \emph{game} and is positioned by a 2D UMAP embedding of distance between games, with the size of the dot corresponding to the \emph{balance} of the game, and the colour representing competitiveness of the game (from blue -- completely competitive, to purple -- completely cooperative). \textbf{(Right)} Each game can be played on a myriad of \emph{worlds}, which we can smoothly mutate to traverse a diverse set of physical challenges. \textbf{(Bottom)} An XLand \emph{task} consists of combining a game with a world and co-players.}
\label{fig:xland_space}
\end{figure*}

The result of this training process is an agent that is generally capable across the held-out evaluation space. Qualitatively, we observe the agent exhibiting behaviours that are generally applicable, rather than optimal for any specific task. Examples of such behaviours include: experimentation through directed exploration until the agent recognises a rewarding state has been achieved; seeking another player out to gather information of its state irrespective of its goal; and tagging another player if it is holding an object that is related to the agent's goal irrespective of that player's intention. We also probe quantitatively the behaviour of agents in test-time multi-agent situations and see evidence of cooperation emerging with training. In addition to the agent exhibiting zero-shot capabilities across a wide evaluation space, we show that finetuning on a new task for just 100 million steps (around 30 minutes of compute in our setup) can lead to drastic increases in performance relative to zero-shot, and relative to training from scratch which often fails completely.

The paper is organised as follows: first we introduce the XLand environment space in \secref{sec:environment} followed by an exploration of the quantitative properties of this environment space in \secref{sec:envproperties}. In \secref{sec:goal} we introduce the goal, metric, and evaluation space we use to measure progress in the open-ended environment. In \secref{sec:training} we detail the different components of our learning system and how these work together. \secref{sec:results} describes the experimental results, dynamics, and analysis of the produced agent. Finally, \secref{sec:related} gives an overview of some related works, followed by the conclusions of this work in \secref{sec:conclusions}.
All proofs and experimental details can be found in the Appendices.

\section{XLand Environment Space}
\label{sec:environment}
To promote the emergence of general behaviour in reinforcement learning, we seek an environment that exhibits dimensions of consistency across tasks, as well as dimensions of smooth variation. 

The development of an environment exhibiting smooth vastness with consistency is central to this work, and as such, we introduce the \emph{XLand} environment space. 
XLand is a 3D environment consisting of static topology together with dynamic objects simulated by rigid-body physics, with multiple players (controllable by both humans or agents) perceiving first person observations and egocentric movement akin to DM-Lab~\citep{beattie2016deepmind} and Quake III: Arena~\citep{jaderberg2019human}. Players have different world-affecting gadgets at their disposal, are able to carry and hold dynamic objects, and receive reward at each timestep based on the state of the environment: relationships between players, objects, and topology. The environment is developed and simulated using the Unity framework from \cite{ward2020using}, with an example task seen in \figref{fig:unity}.

Consistency in this environment comes from: players always having the same control interface, observation specification, gadget dynamics, and movement dynamics; objects being simulated with similar physical properties; and a limited set of topological building blocks. However, the remainder of the environment properties are vastly but also smoothly variable: the layout and structure of topological building blocks, the positions of objects, the lighting, and crucially the specification of rewarding states for each player. Finally, from a single player's perspective, the policies of the co-players can be vastly but smoothly variable. 

The XLand task space, from the perspective of the target player (\emph{e.g.} an agent), denoted as $\xlandspace$, is a Cartesian product of all possible worlds $\world \in \worldspace$, games $\game \in \gamespace$ (defined as one goal $\goal_i \in \goalspace$ for each of the $n$ players), and the policies $\policy_i \in \policyspace$ of each of the remaining $n-1$ players (the players of the game not including the target player).
Formally
$$
\xlandspace := \worldspace \times \bigcup_{n=1}^\infty \left [ \goalspace^{n} \times \policyspace^{n-1} \right ].
$$
Under this definition, each XLand task $$\xlandtask = (\world, (\goal_1, \dots, \goal_n), (\policy_2, \dots, \policy_n)) \in \xlandspace$$ can be seen as a regular POMDP over a simulation state space $\statespace$.
For notational simplicity we often refer to the policy of the target player as either $\policy$ or $\policy_1$.
At each timestep $t$, each player $\policy_i$ receives its player-centric observations $\observation_{t}^{i} := (f_i(\state_t), \goal_i)$, where $f_i$ extracts a pixel-based render of the state of the world from the perspective of player $i$ and also provides the prioperception readings (\emph{e.g.} whether a player is holding something). Note, the reward from the environment is not included in player observations. 
Based on these observations, an action $\mathbf{a}_t^i$ of each player is sampled from its corresponding policy $\mathbf{a}_t^i \sim \policy_i(\observationhistory_t^i)$, where $\observationhistory_t^i = (\observation_1^i, \dots, \observation_t^i)$ is a sequence of observations perceived so far.
The initial state of the simulation is uniquely identified by $\world$. The simulation is terminated after a fixed number of $T=900$ iterations (two minutes when simulated in real-time).
The transition function comes from the simulation's physics engine that calculates the new state $\state_{t+1}$ from its current state $\state_t$ given the simultaneous actions of all the players involved in a specific task $(\mathbf{a}_t^i)_{i=1}^{n}$, analogously to other multi-agent real-time environments~\citep{vinyals2019grandmaster,berner2019dota,jaderberg2019human}. From the perspective of a single player (such as a learning agent), actions of all the co-players can be seen as part of the transition function, and thus the whole process relies only on $\mathbf{a}_t^1$, the action of the target player.
The reward function $\reward_t: \statespace \rightarrow \{0,1\}$ returns 1 if and only if a player's goal is satisfied in the current simulation state.
Consequently, on a given task, a player's goal is to maximise the expected future discounted\footnote{For notation simplicity we will omit the dependence of all returns/values on the discount factor value $\gamma$.} number of timesteps in which its goal is satisfied
$$
\mathbf{V}_\policy(\xlandtask) := \mathbb{E} \left [ \return_\policy(\xlandtask) \right ] = \mathbb{E} \left [ \sum_{t=1}^T \gamma^t \reward_t \right ].
$$

We will now describe in more detail the makeup of the XLand environment, separating out the initial conditions of the physical environment space, \emph{worlds}, from the specification of rewarding states for each player, \emph{games}. We will highlight the vastness and smoothness of these components of XLand, and finally how these components combine and interact to form a vast and complex space of \emph{tasks}, \figref{fig:xland_space}. 

\subsection{World Space}
Tasks in XLand are embedded within 3D physically simulated worlds, an example of which shown in \figref{fig:unity}. The layout of the topology, the initial locations of the objects, the initial locations of the players, and the gadgets at each players' disposal are central to the behaviour being asked of a capable player in this task. For example, consider the simple game consisting of a single player, which receives reward when the player is near a purple sphere. If the player is initially located next to the purple sphere, the player needs to simply stand still. If the purple sphere is initially located out of sight of the player, the player must search for the object. The topology could provide navigational challenges to this search, requiring analysis of connected paths and memory to quickly find the object. The physical interaction between the initial location of the sphere and the topology or other objects could cause the sphere to roll, requiring the player to intercept the sphere once it is found, and if the player has a freeze gadget this would allow the player to stop the sphere rolling by freezing its motion.

The initial condition of a simulated world defines the possible challenges faced by a player somewhat independently of the game, the goals of the players. As such, we define the world $\world$ as the initial state of the simulated 3D world and its constituents, the state at the beginning of each episode of play. The three main components of a world are the topology, objects, and players. Worlds are procedurally generated~\citep{shaker2016procedural}.

\paragraph{Topology}
A world in XLand contains a static topology which defines the unmovable landscape that is navigated by the players, surrounded by four walls which enclose the rectangular playable area, with variable lighting conditions. The topology is generated by first selecting a rectangular size of the world which encloses a grid, and subsequently placing a number of predefined 3D topological tiles. These tiles can be placed in any arrangement but cannot violate local neighbourhood connectivity constraints, ensuring that the arrangement of 3D tiles forms congruent and connected playable regions.

\paragraph{Objects}
Objects are elements of XLand worlds that are dynamic -- they undergo physics simulation and can be manipulated by players. Each world defines a specified initial location for each movable object as well as its orientation, shape, colour and size. Object instances vary in size, colour, and shape. There are three colours -- black, purple, yellow -- and four shapes -- cube, sphere, pyramid, slab.

\paragraph{Players}
The players of the game, which can be controlled by agents, are given initial positions in the same manner as objects. Players are coloured, and in this work we consider up to three players, each being assigned a unique colour of either blue, red, or green. In addition, each player is assigned a gadget: either the freeze gadget or the tagging gadget. The freeze gadget can be used by a player only on an object and has the effect of freezing the dynamics of the object so that it remains static and unmovable for 5 seconds, before becoming dynamic again and undergoing physics simulation as normal. The tagging gadget can be used by a player on an object or another player and has the effect of removing the object or player from the world for 3 seconds, before the object or player is returned to the world at its initial location, rather than the location at which it was removed.

An instance of a world $\world$ is therefore a particular topology, combined with a set of objects with locations, and a particular set of players with locations and gadgets. An agent playing in a world $\world$ will always experience identical initial conditions. 

Our process of generating worlds leads to a vast and smooth space of worlds, with these properties explored further in \secref{sec:worldproperties}. More details of this process can be found in \secref{sec:app-procgen} and \figref{fig:worldgen}.

\subsection{Game Space}

Whilst a world defines the initial state of the simulated physical space for the players to act in, a task requires a \emph{game} for these players to act towards. A game $\game$ consists of a goal $\goal_i \in \goalspace$ for each of the $n$ players, $\game = (\goal_1, \ldots, \goal_n)$. A goal defines the reward function for the associated player, and each player is tasked with acting in a way to maximise their total reward, while perceiving only their own goal (and not seeing goals of the co-players).

The state of our simulated environment $\state \in \statespace$ describes the physical world the players interact with. $\state$ consists of the positions of all the objects, players, their orientations, velocities, \emph{etc}. We define a set of atomic predicates $\predicate_j : \statespace \rightarrow \{0,1\}$ in the form of a physical relation applied to some of the entities present in the state. These relations include: being near, on, seeing, and holding, as well as their negations, with the entities being objects, players, and floors of the topology. An example predicate could be \texttt{near(purple sphere, opponent)}, which is going to return 1 if and only if one of the co-players is currently close to a purple sphere.
With the set of possible predicates fixed, a goal of a player can be represented by a set of options (disjunctions) over sets of necessary predicates for this option (conjunctions). Consequently, an example goal could look like
$$
\goal = \underset{\text{option 1}}{\underbrace{(\predicate_{j_1} \wedge \predicate_{j_2})}} \vee \underset{\text{option 2}}{\underbrace{(\predicate_{j_2} \wedge \predicate_{j_3} \wedge \predicate_{j_4})}}
$$
which, for some example predicates, could mean \emph{``Hold a purple sphere ($\predicate_{j_1}$) while being near a yellow sphere ($\predicate_{j_2}$) \textbf{or} be near a yellow sphere ($\predicate_{j_2}$) while seeing an opponent ($\predicate_{j_3}$) who is not holding the yellow sphere ($\predicate_{j_4}$)''}. This is a canonical representation of Boolean formulas, the \textit{disjunctive normal form} (DNF), which can express any Boolean formula~\citep{davey2002introduction}. 
The corresponding reward function $\reward_\goal(\state)$ follows the transformation of disjunctions becoming sums, and conjunctions becoming products, \emph{i.e.} for a goal $\goal := \bigvee_{i=1}^k [\bigwedge_{j=1}^{n_i} \predicate_{ij}]$:
$$
\reward_\goal(\state) = \max_{i=1}^k \left [ \min_{j=1}^{n_i} \predicate_{ij}(\state) \right ] = \min \left ( \sum_{i=1}^k  \prod_{j=1}^{n_i} \predicate_{ij}(\state), 1 \right ).
$$

A simple example in our game space would be the game of hide and seek. The two-player version of the game consists of two goals $(\goal_\text{seek}, \goal_\text{hide})$ where the goal of one player consists of just one option of one predicate, $\goal_\text{seek}=\predicate_{\text{seek}} = \texttt{see(me, opponent)}$, and the goal of the co-player is  $\goal_\text{hide}=\predicate_{\text{hide}} = \texttt{not(see(opponent, me))}$.

This general construction of games allows us to represent a vast number of highly diverse games, ranging from simple games of finding an object to complex, strategically deep games.
Importantly, the space of games is also smooth, allowing for gradual transition between games. These properties are explored in \secref{sec:gameproperties}.

\subsection{Task Space}
 A task in XLand $\xlandtask$ is the combination of a world $\world$, a game $\game$ and the policies of the \emph{co-players} $(\policy_2, \dots, \policy_n)$. With this view, despite its clearly multi-agent nature, we can view each task as a standard single-player problem for $\policy_1$.

The combination of a world, a game, and co-players can interact in complex ways to shape the space of optimal behaviours required of the player. Consider the example game where the player has a goal consisting of two options \emph{``Hold a purple sphere  \textbf{or} hold a yellow sphere''} and there is one co-player with the identical goal. If the game is played in a fully open world where initially both rewarding objects are visible, the challenge to obtain the optimal behaviour is to choose to navigate to the closest object. If the paths to each object are occluded along the route, the optimal behaviour might require memory to reach its goal object, remembering the path to take. If the world is such that only one of the objects is initially visible but out of reach on a higher floor, the optimal behaviour may be to manipulate another object to reach the goal object. Now consider the variation of co-player policies. If the co-player picks up the purple sphere and moves away quickly, the optimal behaviour of the player may be to ignore the purple sphere and navigate to hold the yellow sphere. However, if the co-player seeks out the player and uses its tagging gadget on sight, hindering the player's ability to navigate to the goal object, the optimal behaviour of the player may be to avoid being seen or to tag the co-player itself, before navigating to a goal object. 

A result of this complex interaction is that the cross product of a set of worlds, games, and co-player policies creates a set of tasks with challenges -- optimal behaviours of the player -- which is larger than the sum of the number of worlds, games, and co-player policies.

\section{Environment Properties}
\label{sec:envproperties}
The previous section introduced the XLand environment space and its tasks' construction from worlds, games, and co-players. In this section we analytically and empirically explore some of the properties of this space, focusing on world and game properties independently. In both cases we explain how these components give rise to the properties of vastness, diversity, and smoothness.

\subsection{World Properties}
\label{sec:worldproperties}
The worlds are high dimensional objects consisting of topology, object locations, and player locations. To highlight the characteristics of worlds, we can describe a world in terms of the navigational challenges it poses due to the topology and the objects.

Our worlds are all grid-aligned, with varied dimensions of each single \emph{tile}, some of which (ramps) one can use to navigate to a higher level. We consider two world representations: first, the height map, $\tau(\world) : \worldspace \rightarrow [0,1]^{w \times h}$ where $w, h$ are the width and height of the world respectively, and each element in $\tau(\world)$ is the height of the top of the tile at the location of the element. The second representation is a world topology graph, representing navigation paths.
\begin{defn}[World topology graph]
For a given world $\world$, we define a directed graph $G_\world = (V_\world, E_\world)$ where each tile of a world is represented as a vertex, and an edge exists between two vertices $v_i$ and $v_j$ if and only if it is possible for a player to travel between the two neighbouring tiles in a straight line (they are on the same level, the height of $v_j$ is lower so the agent can fall to it, or $v_j$ is an accessible ramp leading to a higher level). 
\end{defn}
Given this graph, we can define various proxy measures of navigational complexity by looking at the distribution of paths between every pair of vertices.
\begin{defn}[Shortest paths distribution]
For a given $\world$ we define $\rho_\mathrm{sp}(\world)$ as a distribution of lengths of shortest paths between every pair of vertices in $G_\world$.
\end{defn}
\begin{defn}[Resistance distances distribution]
For a given $\world$ we define $\rho(\world)$ as a distribution of resistance distances~\citep{klein1993resistance} between every pair of vertices in $G_\world$, where a resistance distance between $v_i$ and $v_j$ is given by $\Gamma_{ii} + \Gamma_{jj} - \Gamma_{ij} - \Gamma_{ji}$ for $\Gamma = (L + \tfrac{1}{w\cdot h} \mathbf{1}_{w\cdot h \times w\cdot h})^\dagger$, $L$ being the Laplacian matrix of $G_\world$ and $\dagger$ being the Moore-Penrose pseudoinverse~\citep{penrose1955generalized}.
\end{defn}

\subsubsection{World Vastness}
\begin{figure}[t]
    \centering
    \includegraphics[width=0.85\linewidth]{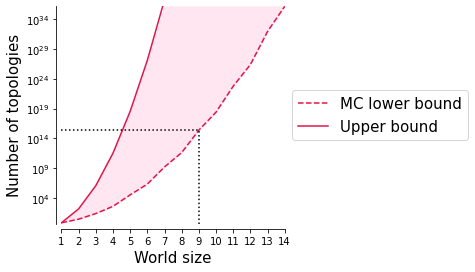}
    \caption{Visualisation of the bounds of the number of possible world topologies of shape $(n,n)$ as a function of a world size $n$. See \secref{app:worldcount} for details.  
    }
    \label{fig:world_bound}
\end{figure}

Let us start by discussing the vastness of worlds by looking at how many topographies are possible in XLand. In principle, every combination of tiles could be utilised, creating $N_\mathrm{floors}^{w \cdot h } \cdot N_\mathrm{tiles}$ possibilities. However, as discussed previously, constraints on tile placements exist to ensure ramps connect levels and there are accessible playable regions. Consequently, it is reasonable to count the number of world topologies where all ramps are properly connected, and that have at least 50\% of the world fully accessible (there exists a way to go from every point to any other point within the accessible area). We estimate a lower bound to this quantity with Monte Carlo sampling, and present results in \figref{fig:world_bound} (see \secref{app:worldcount} for details). For worlds of size 9 by 9 tiles, (9,9), we have more that $10^{16}$ unique topologies (corrected for 8 possible symmetries) -- a vast space of worlds.

\subsubsection{World Smoothness}
\begin{figure}[t]
    \centering
    \includegraphics[width=0.475\linewidth]{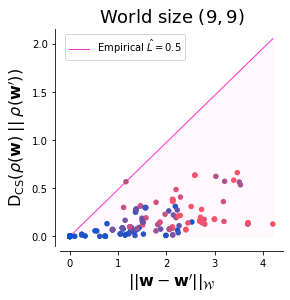}
    \includegraphics[width=0.475\linewidth]{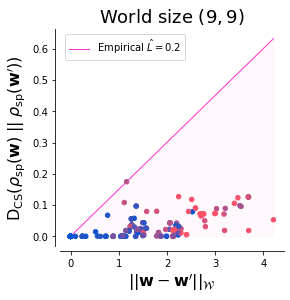}\\
    \includegraphics[width=0.475\linewidth]{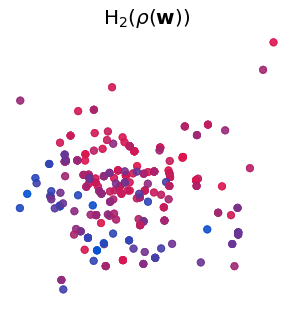}
    \includegraphics[width=0.475\linewidth]{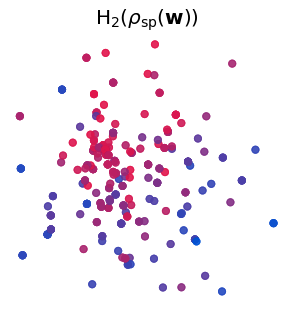}
    \caption{An empirical visualisation of the world space smoothness. We take a set of worlds of size (9,9) and then apply local mutations up to 30 times. \textbf{(Top)} Each dot on the plot represents one pair of mutated worlds, with the x-axis showing the $L_2$ distance in tile space, and the y-axis showing the Cauchy-Schwarz Divergence between distributions of reachability graph resistances $\rho(\world)$ (left) and shortest path distances $\rho_\mathrm{sp}(\world)$ (right). The pink line represents the empirical smoothness coefficient. The colour of each dot encodes the number of mutations between the pair of worlds, from 1 (blue) to 30 (red). \textbf{(Bottom)} We linearly embed each of the worlds, trying to find a linear projection where the entropy of the corresponding distribution (in colour) can be well described by a distance from the center of the projection. One can see how small changes in the world space (position) lead to small deviations of the entropy (colour).}
    \label{fig:world_smooth}
\end{figure}
\begin{figure*}[t]
    \centering
    \includegraphics[width=\linewidth]{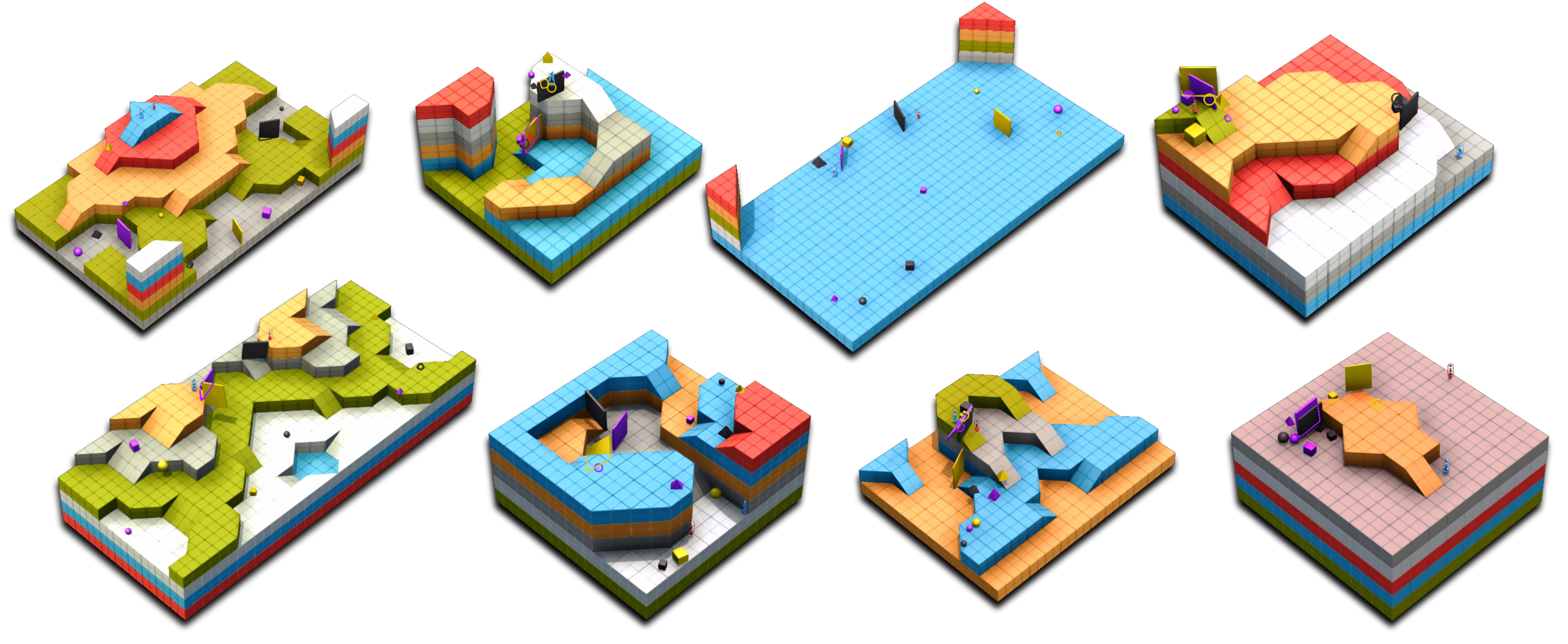}
    \caption{An example array of worlds from the XLand environment space.}
    \label{fig:lotsofworlds}
\end{figure*}
We hypothesise that small changes in the world topography lead to small changes in the overall navigational complexity.
To formalise this claim we take a set of 9 by 9 worlds, and then apply local changes to each of them, in the form of moving tiles around, changing floors, \emph{etc}.
Given this set of mutated worlds, we plot the relation between the change in the topography 
$$\|\world - \world'\|_\worldspace := 
\sqrt{\sum_{i,j=1}^{w,h} [\tau(\world)_{ij} - \tau(\world')_{ij}]^2 }$$
and the Cauchy-Schwarz Divergence~\citep{nielsen2012closed}
\begin{equation*}
\begin{aligned}
\mathrm{D}_\mathrm{CS}(p, q) &:= - \mathrm{H}_2(p) - \mathrm{H}_2(q) + 2\mathrm{H}^\times_2(p,q)\\
& := \log \int p^2 (x) dx + \log \int q^2(x) dx\\
& \phantom{:=}  - 2\log \int p(x)q(x) dx,
\end{aligned}
\end{equation*}
between the corresponding shortest paths distributions $\rho_\mathrm{sp}$ and resistance distances distributions $\rho$. The top row of \figref{fig:world_smooth} shows that there is a visible linear bound in the change in the paths distributions, suggesting L-Lipschitzness.

To further confirm this claim, we take the same set of worlds and find a linear projection (\secref{app:worldprojection}) that embeds our worlds in
a 2-dimensional space, with each point coloured by its corresponding Renyi's quadratic entropy $\mathrm{H}_2$ of the distribution of paths over its navigation graph, \figref{fig:world_smooth}~(bottom). We can see that the world space appears smooth.

\subsubsection{World Diversity}
The world topology, jointly with object and player positions, allow one to express arbitrary navigation challenges, including various types of mazes, but also complex maps with difficult to access regions, and occluded visibility similar to the maps used in competitive first-person video games~\citep{jaderberg2019human}, see \figref{fig:lotsofworlds}. 

\begin{figure}[t]
    \centering
    \includegraphics[width=\linewidth]{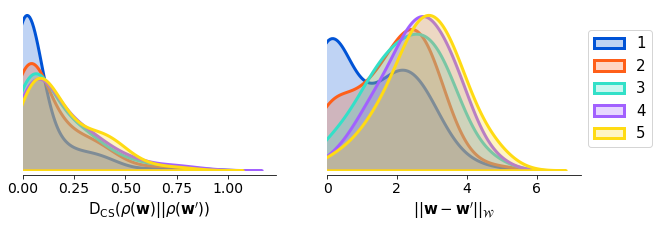}
    \caption{The distributions of distances between two worlds, $\world$ and $\world'$, with different number of local tile mutations between them (colour). The distances are the navigational resistance (left) and topology distance (right). With only a few mutations the characteristics of the world can change a lot.}
    \label{fig:worldchars}
\end{figure}

To illustrate diversity, one can see that the Cauchy-Schwarz Divergence between resistance distances distributions $\rho$ as well as topology distances can be increased with relatively few local tile mutations (see \figref{fig:worldchars}). This confirms that, despite being a relatively smooth space, our world space spans diverse worlds, which can be found with local search methods (\emph{e.g.} evolutionary algorithms).

\subsection{Game Properties}
\label{sec:gameproperties}
Once multiple goals are combined to form a game, new complexity emerges -- the ways in which the objectives of players interact, compete, and affect each other. This complex interaction is central to the fields of Game Theory and multi-agent RL~\citep{shoham2007if,balduzzi2019open}. To characterise the properties of games, we focus our analysis on three dimensions of games: the number of options, exploration difficulty, and their cooperative/competitive/balance aspects.

The first property is the number of options in a given goal (and a game). Having multiple options for getting a reward in each timestep encourages players to be constantly evaluating the environment state, assessing which option is the more rewarding one. 

To define the more complex game properties, recall that every goal is a Boolean expression over a set of $d$ predicates $\predicate_j$. Let us define $\predicate: \mathcal{S} \rightarrow \{0,1\}^{d}$, a mapping that assigns each simulation state $\state$ to a binary vector of $d$ predicate truth values. A goal is simply a mapping from $\predicate(\statespace)$ to $\{0,1\}$, labelling which predicate states are rewarding. 
We denote by $N_\predicate := \#\{\predicate(\state): \state \in \statespace\}$ the size of the predicate state space. 
We define a distance metric between two goals $\goal_i$ and $\goal_j$ as 
$$
\| \goal_i - \goal_j \|_\goalspace := \frac{\#\{ \predicate(\state) : \reward_{\goal_i}(\state) \neq \reward_{\goal_j}(\state) \}}{N_\predicate} \in [0, 1].
$$
This distance between two goals is the fraction of different predicate evaluations where one goals is rewarding, and the other goal is not. Analogously, between two games
$$
\| \game_i - \game_j \|_\gamespace := \tfrac{1}{n} \sum_{k=1}^{n} \| (\game_i)_k - (\game_j)_k \|_\goalspace \in [0, 1].
$$
This leads to the following observation.
\begin{obs}
$(\goalspace, \| \cdot \|_\goalspace)$ and $(\gamespace, \| \cdot \|_\gamespace)$ are metric spaces.
\end{obs}
\noindent In particular we have
$$
\goal_i \equiv \goal_j \iff \reward_{\goal_i} = \reward_{\goal_j} \iff \| \goal_i - \goal_j \|_\goalspace =0 $$
$$
\game_i \equiv \game_j \iff  \| \game_i - \game_j \|_\gamespace =0 .
$$

This allows us to define the next game property: exploration difficulty.
\begin{defn}{Exploration difficulty} of a game is the fraction of predicate states in which no player is being rewarded.
$$
\ed(\game) = \ed((\goal_1, \dots, \goal_n))= \frac{ \#\{ \predicate(\state): \forall_k \reward_{\goal_k}(\state) = 0 \} }{N_\predicate}
$$
we will also call the unnormalised exploration difficulty the quantity
$$
\hat{\ed}(\game) :=N_\predicate \ed(\game).
$$
\end{defn}
One simple interpretation of this quantity is: assuming each of the predicates is independent and equally probable to be (dis-)satisfied at a given timestep, then $1-\ed(\goal)$ describes the probability of at least one player getting a reward. Consequently, we will refer to goals as trivial if $\ed(\goal) \in \{0, 1\}$, since these are goals where every policy is an optimal policy (similarly we say a game is trivial from the perspective of the main agent if $\ed(\goal_1) \in \{0, 1\}$).
\begin{proprep}
For every goal $\goal$ where $\ed(\goal) = 0$ or $\ed(\goal) = 1$ every policy is optimal.
\end{proprep}
\begin{proof}
Lets assume that $\ed(\goal_1) = z \in \{0,1\}$ this means that for every state $\state$ we have $\reward_\goal(\state) = z$. Consequently, for every policy $\policy$ we have $\mathbf{V}_\policy(\xlandtask) = \mathbf{V}_\policy((\world, (\goal_1, \dots, \goal_n), (\policy_1, \dots, \policy_n))) = Tz$, and so in particular $\forall_\policy \mathbf{V}_\policy(\xlandtask) = Tz = \max_\policy \mathbf{V}_\policy(\xlandtask) = \mathbf{V}^*(\xlandtask)$.
\end{proof}

Given exploration difficulty, we now define a new property -- the notion of cooperativeness -- that will assign a number between 0 and 1 to each game, where a game of cooperativeness 1 is going to be one where all players always get rewards jointly, and cooperativeness 0 when they can never both get a reward at the same timestep.
\begin{defn}{Cooperativeness} is the fraction of predicate states in which all the players are being rewarded compared to the number of predicate states in which at least one of them is.
$$
\coop(\game) = \coop((\goal_1, \dots, \goal_n)) = \frac{
\#\{ \predicate(\state): \forall_k \reward_{\goal_k}(\state) = 1\}
}{
N_\predicate - \hat{\ed}(\game)
}
$$
\end{defn}

Symmetrically, competitiveness can be expressed as $\comp(\game) = 1 - \coop(\game)$ or more explicitly with the following definition.
\begin{defn}{Competitiveness} is the fraction of predicate states in which some but not all players are being rewarded compared to the number of predicate states in which at least one of them is.
$$
\comp((\goal_1, \dots, \goal_n)) = \frac{
\#\{ \predicate(\state): \max_k \reward_{\goal_k}(\state) \neq \min_k \reward_{\goal_k}(\state)\}
}{
N_\predicate-\hat{\ed}(\game)
}
$$
\end{defn}

Finally, let us introduce the property of balance of a game. In game design, the issue of one player of the game having a constant advantage is a common one, referred to as an imbalance. Whilst fully symmetric, simultaneous moves games are fully balanced by construction, it is a complex problem to assess the degree of balance when the game is not symmetric, \emph{i.e.} when the goals of each player are different. 
\begin{defn}Balance with respect to game transformations $\Xi \supset \{ \mathrm{identity}\}$ is the maximal cooperativeness of the game when goals are transformed with elements of $\Xi$:
\begin{equation*}
\begin{aligned}
\balance(\game) = \max_{\xi \in \Xi} \coop(\xi(\game)).
\end{aligned}
\end{equation*}
\end{defn}
With the above definition it is easy to note that when $\Xi = \{\mathrm{identity}\}$ then balance is equivalent to cooperativeness. Consequently, balance can be seen as a relaxation of the notion of cooperation, under the assumption that some aspects of game rules are equivalent (equally hard). For XLand we note that colours of objects should have negligible effect on the complexity of a task, meaning that satisfying a predicate \texttt{hold(me,yellow sphere)} should be equally hard as \texttt{hold(me,purple sphere)}. Consequently, we use $\Xi$ to be the set of all bijective recolourings of objects in goals that are consistent across the entire goal.

\subsubsection{Game Vastness}
\begin{figure}
    \centering
    \includegraphics[width=\linewidth]{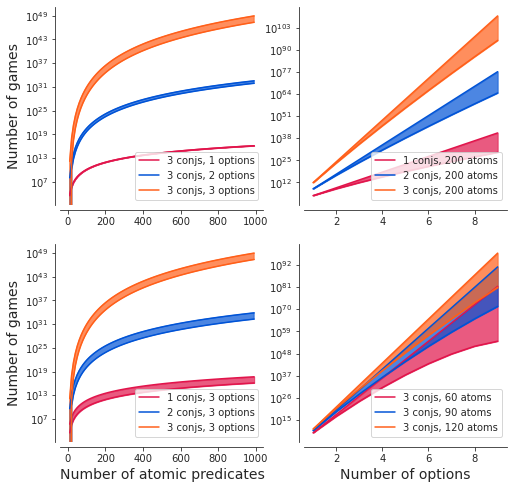}
    \caption{Bounds on the number of 2-player games provided by Theorem~\ref{thm:game_number} as functions of the number of options and atomic predicates. Our evaluation space (\secref{sec:evalset}) spans approximately 200 atomic predicates and up to 3 options.}
    \label{fig:game_number_bounds}
\end{figure}

Let us denote the number of unique atomic predicates as $n_a$, the number of options a goal consists of as $n_o$ and the number of predicates in each option as $n_c$. There are exactly $n_a^{n_c \cdot n_o}$ goals that differ in terms of their string representation, however many goals are equivalent such that $$\goal_i \equiv \goal_j \iff \reward_{\goal_i} = \reward_{\goal_j}.$$ For example, the goal of \emph{seeing a purple sphere or not seeing a purple sphere} is equivalent to the goal of \emph{holding a yellow cube or not holding a yellow cube}, both corresponding to $\reward(\state) = 1$. Counting the exact number of unique $\reward$ functions that emerge from $n_i$ options each being a conjunction of $n_c$ out of $n_a$ predicates is a hard combinatorial problem, but under the assumption that each atomic predicate (apart from their negations) is independently solvable we can provide a lower bound of 
the number of unique goals.
\begin{thmrep}
\label{thm:game_number}
Under the assumption that each atomic predicate that does not involve negation is independently solvable, the number of unique $n$-player games $N_\gamespace$ with respect to the reward functions they define satisfies:
$$
\frac{1}{n!} \left [ \frac{1}{n_o!} \prod_{i=1}^{n_o} \left ( { n_a/2 - i\cdot n_c \choose n_c} 2^{n_c} \right ) \right ]^n \leq N_\gamespace \leq {n_a \choose n_c}^{n \cdot n_o}.
$$
\end{thmrep}
\begin{proof}
The high level idea for the lower bound is to count the number of games where inside each goal every predicate is unique (and can only repeat across players). 

First, let us prove the statement for two goals, $\goal_i$ and $\goal_j$, where each has a corresponding set of predicates used $(\predicate_{i o_k c_l})_{k,l=1}^{n_o,n_k}$ and $(\predicate_{j o_k c_l})_{k,l=1}^{n_o,n_c}$, each being lexicographically sorted over options (indexed by $o_k$, and over predicates inside each option (indexed by $c_l$), so that the option and alternatives orderings are irrelevant.

If the two goals are different, this means that there exists $k^*, l^*$ such that $\predicate_{ik^*l^*} \neq \predicate_{jk^*l^*}$. Let's take the smallest such $k^*$ and a corresponding smallest $l^*$. This means that there exists an option in one the goals, that the other goal does not possess. Without loss of generality, let us assume it is an option of $\goal_i$, meaning that $\neg \exists_{k'} \predicate_{j k'} = \predicate_{i k^*}$. Since this option uses unique predicates across the goal, let us define $s^*$ as a simulation state such that all the predicates of this option are true, while all the other predicates are false. Then we have $$\reward_{\goal_i}(s^*) = 1 \neq 0 = \reward_{\goal_j}(s^*)$$
proving that $\reward_{\goal_i} \neq \reward_{\goal_j}$ and thus $\goal_i \not\equiv \goal_j$.

The only thing left is to count such goals. For that, let us note that this is an iterative process, where for each $i$th of $n_o$ options we can pick $n_c$ out of $n_a/2 - i \cdot n_c$ predicates to be used (since we already picked $i \cdot n_c$ before, and we are not picking negations). Once we picked the predicates, each of them can be either itself or its negations, which introduces the $2^{n_c}$ factor. And since the process is order variant, we need to simply divide by the number of permutations of length $n_o$, leading to
$$
 \frac{1}{n_o!} \prod_{i=1}^{n_o} \left ( { n_a/2 - i c\dot n_c \choose n_c} 2^{n_c}\right )
$$
and completing the lower bound proof.

The upper bound comes from simply noting that every reward function that comes from a Boolean expression with $n_o$ alternatives, each being a conjunction of $n_c$ out of $n_a$ predicates has a corresponding Boolean expression of this form, and thus we can just count how many such expressions are there:
$$
{ n_a \choose n_c}^{n \cdot n_o},
$$
completing the upper bound proof.
\end{proof}
\figref{fig:game_number_bounds} shows these bounds as functions of the number of options, atoms and conjunctions. As an example we see that with 3 options, each a conjunction of 3 predicates, using a set of 200 atomic predicates (the approximate number available in practice) gives us more than $10^{37}$ unique 2-player games (composed of more than $10^{18}$ goals) -- a vast space of games.

\subsubsection{Game Smoothness}
For our XLand task space to exhibit smoothness, the game space itself must be smooth: if we change our games by a small amount, the game properties of interest should only change by a small amount. We show that the interesting properties of our games are L-Lipschitz functions. 
\begin{proprep}
\label{prop:ed_l}
Exploration difficulty is a $1$-Lipschitz function, meaning that for any $\game_i, \game_j$ we have
$$
\| \ed(\game_i) - \ed(\game_j) \| \leq \| \game_i - \game_j \|_\gamespace .
$$
\end{proprep}
\begin{proof}
We will show this with a proof by contradiction. Let us assume that the negation holds, meaning that there are two such games that
$$
\| \ed(\game_i) - \ed(\game_j) \| > \| \game_i - \game_j \|_\gamespace,
$$
This would mean that
\begin{equation*}
\begin{aligned}
\|
& \#\{ \predicate(\state): \forall_k \reward_{(\game_i)_k}(\state) = 0 \} - 
 \#\{ \predicate(\state): \forall_k \reward_{(\game_j)_k}(\state) = 0 \}
\| \\
&>
\| \game_i - \game_j \|_\gamespace N_\predicate.
    \end{aligned}
\end{equation*}
The left hand side of the inequality measures the difference in the number of non-rewarding states.
The right hand side measures the difference in the number of states that simply have a different reward (and thus already includes those counted on the left hand side). 
Clearly the left hand side cannot be strictly bigger than the right.
Contradiction.
\end{proof}

\begin{thmrep}
\label{thm:coop_l}
$\coop((\cdot, \goal'))$ is a $\tfrac{1}{1-k}$-Lipschitz function wrt. $\|\cdot \|_\goalspace$ for any $\goal$ such that $\ed((\goal,\goal')) = k$.
\end{thmrep}
\begin{proof}
Let as assume that 
$$
\| \goal_i - \goal_j \|_\goalspace = \tfrac{z}{N_\predicate}.
$$
From the definition of the metric this means there are exactly $z$ predicate states where one of them is rewarding and the other is not.

Let us denote by $y$ number of predicate states where both $\goal_i$ and $\goal'$ are rewarded.
Then the number of predicate states where $\goal_j$ and $\goal'$ are rewarded has to belong to $(y-z, y+z)$.
Now by denoting $\hat{k} = k N_\predicate$ we have
\begin{equation*}
\begin{aligned}
\|\coop(\goal_i, \goal')  - \coop(\goal_j,\goal')\| (N_\predicate-\hat{k}) &\leq z\\
&= \| \goal_i - \goal_j \|_\goalspace N_\predicate,
\end{aligned}
\end{equation*}
and thus
\begin{equation*}
\begin{aligned}
&\|\coop(\goal_i, \goal')  - \coop(\goal_j,\goal')\| \leq \frac{N_\predicate}{N_\predicate-\hat{k}}  \| \goal_i - \goal_j \|_\goalspace\\
&= \frac{N_\predicate}{N_\predicate-k N_\predicate} \| \goal_i - \goal_j \|_\goalspace= \frac{1}{1-k}\| \goal_i - \goal_j \|_\goalspace
\end{aligned}
\end{equation*}
It is natural to ask if the restriction imposed is not empty, but it is easy to prove that in the vicinity of any game there is another one satisfying said restriction.

\begin{proprep}For any game $\game = (\goal,\goal')$ where $\ed(\game)=k>0$ there exists a goal $\goal''$ such that 
$
\ed((\goal, \goal'')) = k 
$
and it is in vicinity of the previous game in the sense that $$ \|(\goal,\goal'') - (\goal,\goal')\|_\gamespace = \frac{1}{2N_\predicate}.$$
\end{proprep}

Without loss of generality let us assume that $\goal$ has at least one rewarding predicate state $\predicate(\state^*)$. If  $\predicate(\state^*)$ is also rewarding for $\goal'$ then we define $\goal''$ as an exact copy of $\goal'$, but set  $\predicate(\state^*)$ to be non rewarding thus the distance between the two is 1. If it was not rewarding in $\goal'$ we symmetrically make it rewarding in $\goal''$, again moving by 1 in the game space. The resulting game $(\goal,\goal'')$  has  $\ed((\goal,\goal''))=k$ since we did not add any new rewarding predicate states.

\end{proof}
\noindent In a natural way the same is true for competitiveness.
\begin{obs}
$\comp((\cdot, \goal'))$ is a $\tfrac{1}{1-k}$-Lipschitz function wrt. $\|\cdot \|_\goalspace$ for any $\goal$ such that $\ed((\goal,\goal')) = k$.
\end{obs}
Therefore, if we change one of the goals by a small amount, we have an upper bound on the change in exploration difficulty, cooperativeness, and competitiveness of the whole game.

\begin{figure}[t]
    \centering
    \begin{tabular}{cc}
    \includegraphics[width=0.45\linewidth]{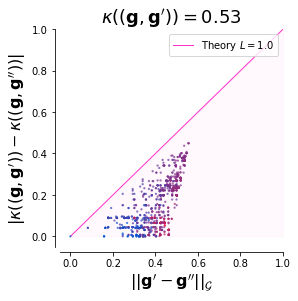} &
    \includegraphics[width=0.45\linewidth]{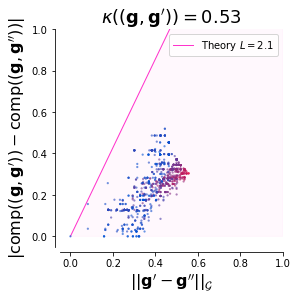} \\
    \includegraphics[width=0.45\linewidth]{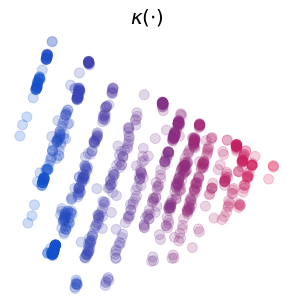} &
    \includegraphics[width=0.45\linewidth]{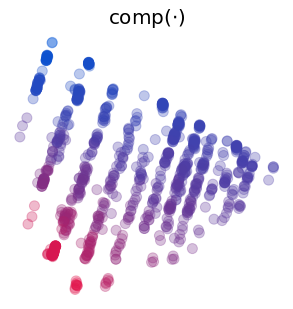} 
    \end{tabular}
    \caption{Empirical confirmation of game space smoothness with respect to exploration difficulty ($\ed$, left) and competitiveness ($\comp$, right). We took a single game, then created 1000 similar games by performing simple edits on one of its goals ($\goal'$) creating a new one ($\goal''$)-- removal of an option/relation, adding a new option/relation, substitution of a relation, \emph{etc}. {\bf(Top)} Each of these games corresponds to a point, with the x-axis being its distance from one randomly selected anchor game $(\game, \game')$ (with the exploration difficulty reported in the title), and on the y-axis the difference in its $\comp$ or $\ed$ (colour corresponds symmetrically to $\ed$ on $\comp$ plot and vice versa). The pink line is the upper bound from Proposition~\ref{prop:ed_l} and Theorem~\ref{thm:coop_l}. {\bf(Bottom)} The matrix of pairwise distances between these games is computed, and PCA used to embed them on a plane, followed by representing $\comp$ and $\ed$ with a point colour. In both cases one can see very smooth transitions. }
    \label{fig:game_smooth}
\end{figure}

\figref{fig:game_smooth} verifies these properties empirically by showing the relation between the distance in game space compared to the change in competitiveness and exploration difficulty. We also provide a 2D projection of sample games using PCA, showing that these two properties are visible, suggesting they explain a lot of variance in game space. These examples show analytically and empirically the smoothness of game space.

\subsubsection{Game Diversity}
\label{sec:game-diversity}
We have shown that the game space consists of vastly many games, and that small changes in their definitions lead to small changes in properties of interest. One missing aspect is to show how diverse this game space is, that eventually, after taking many small steps, one can change a game into a wildly different one.

\begin{thmrep}
For every two player game $\game$ such that $\hat \ed(\game) = k$ and a desired change in competitiveness $m \in (-\comp(\game), 1-\comp(\game) )$ such that $k|m| \in \mathbb{N}$ there exists a $\game'$ such that 
$
\comp(\game') = \comp(\game) + m
$ 
and
$
\|\game-\game'\|_\mathcal{\gamespace} \leq \tfrac{k|m|}{2}.
$
\end{thmrep}
\begin{proof}
Let us first assume that $m>0$, consequently $\comp(\game)$ is smaller than 1, which means that if we look at $\game=(\goal_1,\goal_2)$ we can find at least $k \cdot (1-m)$ predicate states, where $\reward_{\goal_1}(\predicate(\state)_i) = \reward_{\goal_2}(\predicate(\state)_i)$. Let us define
$$
\goal'_2(\predicate(\state)) := \left \{ \begin{matrix}
1-\goal_2(\predicate(\state)) \text{ if } \predicate(\state) \in \{\predicate(\state)_i\}_{i=1}^{mk} \\
\goal_2(\predicate(\state)) \text{ otherwise }\\
\end{matrix}
\right.
$$
By construction $\ed((\goal_1, \goal_2)) = \ed((\goal_1, \goal'_2))$ and 
$\comp((\goal_1,\goal_2)) + m = \comp((\goal_1, \goal'_2)) $ and $\| (\goal_1,\goal_2) - (\goal_1,\goal'_2) \| = \tfrac{km}{2}$. Proof for $m<0$ is analogous.
\end{proof}

To see qualitatively the diversity of games, we present a few examples of games showcasing a range of challenges imposed on players.

\paragraph{Simple navigation task}
XLand games include simple challenges such as a player being tasked with finding an object of interest and grabbing it. Tasks like this challenge navigational skills, perception, and basic manipulation.
\begin{equation*}
\begin{aligned}
\goal_1 &:= \texttt{hold(me, yellow sphere)}\\
\goal_2 &:= \texttt{near(me, yellow pyramid)}\\
\ed(\game) &= \tfrac{1}{4}\;\;\;\;
\comp(\game) = \tfrac{2}{3}\;\;\;\;
\balance(\game) = \tfrac{7}{15}\\
\end{aligned}
\end{equation*}

\paragraph{Simple cooperation game}
Setting the goal of both players to be identical gives a fully cooperative, balanced game, which challenges a player's ability to navigate and manipulate objects, but also to synchronise and work together.
\begin{equation*}
\begin{aligned}
\goal_1 &:= \texttt{near(yellow pyramid, yellow sphere)}\\
\goal_2 &:= \texttt{near(yellow pyramid, yellow sphere)}\\
\ed(\game) &= \tfrac{1}{2}\;\;\;\;
\comp(\game) = 0\;\;\;\;
\balance(\game) = 1\\
\end{aligned}
\end{equation*}

\paragraph{Hide and Seek}A well known game of hiding and seeking, that has been used in the past as a source of potentially complex behaviours~\citep{baker2019emergent}. This is an example of a simple, fully competitive, imbalanced game in XLand.
\begin{equation*}
\begin{aligned}
\goal_1 &:= \texttt{see(me, opponent)}\\
\goal_2 &:= \texttt{not(see(opponent, me))}\\
\ed(\game) &= 0\;\;\;\;
\comp(\game) = 1\;\;\;\;
\balance(\game) = \tfrac{1}{3}\\
\end{aligned}
\end{equation*}

\paragraph{Capture the Cube} The competitive game of Capture the Flag has been shown to be as a rich environment for agents to learn to interact with a complex 3d world, coordinate and compete~\citep{jaderberg2019human}. Each player must get the flag (for example represented as a cube) to their base floor to score reward. An example one-flag instantiation of this game in XLand (with a supporting world) is
\begin{equation*}
\begin{aligned}
\goal_1 &:= &&\texttt{on(black cube, blue floor)} \wedge \\  &&&\texttt{not(on(black cube, red floor))}\\
\goal_2 &:= &&\texttt{on(black cube, red floor)} \wedge \\  &&&\texttt{not(on(black cube, blue floor))}\\
\ed(\game) &=&& \tfrac{1}{4}\;\;\;\;
\comp(\game) = 1\;\;\;\;
\balance(\game) = 1\\
\end{aligned}
\end{equation*}
\paragraph{XRPS} A final example is that of XRPS games, inspired by the study of non-transitivites in games leading to strategic depth~\citep{vinyals2019grandmaster,czarnecki2020}. We give each player three options to choose from, each one being explicitly countered by exactly one other option. A player can choose to pick up a yellow sphere, but it will get a reward if and only if an opponent is not holding a purple sphere; if it picks up a purple sphere the reward will be given if and only if the opponent does not pick up a black sphere, and so on. With these cyclic rules, players are encouraged not only to navigate and perceive their environment, but also to be aware of opponent actions and strategies, and to try to actively counter potential future behaviours, leading to potentially complex, time-extended dynamics.
\begin{equation*}
\begin{aligned}
\widehat{\goal}_\mathrm{rock} &:=&& \texttt{hold(me,yellow sphere)} \wedge \\
&&&\texttt{not(hold(opponent,yellow sphere))}\wedge\\
&&&\texttt{not(hold(opponent,purple sphere))}\\
\widehat{\goal}_\mathrm{paper} &:=&& \texttt{hold(me,purple sphere)}\wedge \\
&&&\texttt{not(hold(opponent,purple sphere))}\wedge\\
&&&\texttt{not(hold(opponent,black sphere))}\\
\widehat{\goal}_\mathrm{scissors} &:=&& \texttt{hold(me,black sphere)}\wedge \\
&&&\texttt{not(hold(opponent,black sphere))} \wedge\\
&&&\texttt{not(hold(opponent,yellow sphere))}\\
\goal_1 &:= &&\widehat{\goal}_\mathrm{rock} \vee \widehat{\goal}_\mathrm{paper} \vee \widehat{\goal}_\mathrm{scissors}\\
\goal_2 &:=&& \widehat{\goal}_\mathrm{rock} \vee \widehat{\goal}_\mathrm{paper} \vee \widehat{\goal}_\mathrm{scissors}\\
\ed(\game) &=&& \tfrac{1}{4}\;\;\;\;
\comp(\game) = 1\;\;\;\;
\balance(\game) = 1\\
\end{aligned}
\end{equation*}

\section{Goal and Metric}
\label{sec:goal}

\begin{figure*}[t]
    \centering
    \includegraphics[width=\textwidth]{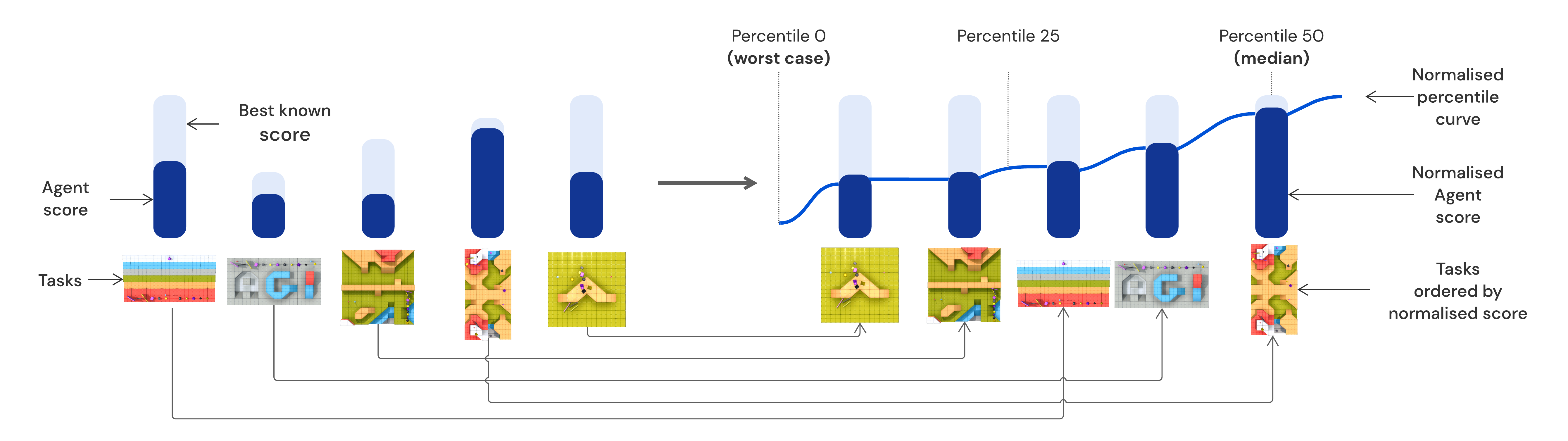}
    \caption{The process of computing normalised percentiles. Tasks vary significantly in terms of their complexity, some have much higher values of optimal policies than others. We normalise the performance of the agent by an estimate of an optimal policy score -- using the Nash equilibrium of trained agents -- providing a normalised score, which after ordering creates a normalised percentile curve. This can be iteratively updated as new trained agents are created.}
    \label{fig:normpercentiles}
\end{figure*}
In \secref{sec:environment} we introduced the XLand environment and explored some of the properties of this space in \secref{sec:envproperties} such as the vastness, diversity, and smoothness across tasks. We now turn our attention to training an agent on XLand.

To train an agent $\policy$ in an episodic environment such as XLand, one generally aims to maximise the expected return of the agent
$$
\mathbf{V}_\policy(\mathrm{P}_\policy) := 
\mathbb{E}_{\mathrm{P}_\policy(\xlandspace)} \left [ \return_\policy(\xlandtask) \right ].
$$
where $\mathrm{P}_\policy$ is an agent-specific distribution over tasks. 

A challenge in evaluating the performance of an agent in this massively multitask environment comes from the fact that each task can be of completely different complexity. The optimal value 
$$\mathbf{V}^*(\xlandtask) := \max_\policy \mathbf{V}_\policy(\xlandtask)$$ of one task\footnote{$\mathbf{V}_\policy(\xlandtask)$ is interpreted as an expectation over the Dirac delta distribution around $\xlandtask$.} can be of a different order of magnitude than the optimal value of another task $\mathbf{V}^*(\xlandtask')$, \emph{i.e.} $\mathbf{V}^*(\xlandtask) \gg \mathbf{V}^*(\xlandtask').$ Consequently, simply averaging the agent's value across all tasks to form a single score will overemphasise tasks with large optimal values. Even if one was able to sensibly normalise value per-task, with a big enough set of tasks, averaging will remove relevant information regarding agent failures. For example, averaging will not surface an agent's failure modes on some tasks if these tasks do not occupy a big part of the task space~\citep{balduzzi2018re}. This becomes an even bigger issue if there is no particular ground truth test distribution 
of interest, but rather our goal is to find a policy that is generally capable. 

A Game Theoretic solution would be to focus on the infimum performance (the worst-case scenario~\citep{nash1950equilibrium}), since performance on this task will always lower bound any expectation over a distribution defined over the same set. Unfortunately, the infimum suffers from not providing any notion of progress or learning signal if there are any tasks that are simply impossible or extremely hard.

\subsection{Normalised Percentiles}
In this work we seek to create \emph{generally capable} agents in the whole XLand task space. General capability is not strictly defined but has some desiderata:
\begin{itemize}
    \item Agents should catastrophically fail on as few tasks as possible.
    \item Agents should be competent on as many tasks as possible.
    \item Broad ability is preferred over narrow competency.
\end{itemize}
These desiderata cannot be encapsulated by a single number describing an agent's performance, as they do not define a total order~\citep{balduzzi2019open}.
We move away from characterising agents purely by expected return, and instead consider the distribution of returns over a countable task space. 
However, for large task spaces this is a very high-dimensional object. In addition, due to the drastically different return scales of tasks, returns cannot be compared and one needs knowledge of each individual task to interpret the significance of reward. 
Naturally, one could normalise the return per task by the return of the optimal policy on each specific task. However, in practice:
\begin{itemize}
    \item an optimal policy is not known a priori,
    \item we want to use these normalisers over the entire environment space, which means that we need to know a single optimal policy for the entire space, and then normalise by its score on each task. 
\end{itemize}
In order to address these issues we follow two practical simplifications. 

First, to address the need of having one optimal policy for the entire space, we compute the normalisation factor for each game independently, and then combine them into a global normaliser. 

Second, even with the above simplification we do not have access to an optimal policy per game. However, we can take ideas from multi-agent training algorithms that eventually converge to a Nash equilibrium~\citep{heinrich2016deep,marris2021multi,mcmahan2003planning}. We iteratively build a set of agents that are capable of solving a specific goal, and use the best mixture (Nash equilibrium) of them as a normalising constant. As training progresses and our agents become better at satisfying the goal, they will beat the existing Nash equilibrium and improve the normalising constant. This dynamic provides us with an iterative notion of improvement for a multi-task environment, rather than a fixed numerical quantity to describe progress. It is akin to theoretical results showing that in multi-agent problems it is impossible to have a fixed objective, because finding better agents and improving the quality of evaluation are the same problem~\citep{garnelo2021pick}. These normalisers give us a normalised score per task.

Finally, to mitigate the problem of having a high-dimensional normalised score distribution, we characterise the distribution in terms of the percentiles of normalised score, up to a maximum of the 50th percentile (median normalised score):
\begin{equation*}
\begin{aligned}
\perf(\policy | \goal, \population_t) &:= \min_{ (\policy_j, \goal_j)} \mathbb{E}_{\world}\left [ \return_\policy(\world, (\goal, \goal_2, ... \goal_n), (\policy_2, ... \policy_n)) \right ] 
\\
\normaliser(\goal | \population_t) &:= \max_{\policy} \perf(\policy | \goal, \population_t) = \mathrm{NashValue}(\goal | \population_t)
\\
\normperf(\policy | \goal, \population_t) &:=  \frac{
\perf(\policy | \goal, \population_t) 
}{
\normaliser(\goal | \population_t)
} \in [0,1]
\\
\perc(\policy | \population_t)[k] &:= \mathcal{P}_k( \normperf(\policy | \goal, \population_t) ), \;\; \text{ for } k \in \{0, \dots 50\}
\end{aligned}
\end{equation*}
where $\mathcal{P}_k$ is the $k$th percentile and both min and max operations over policies operate over convex combinations\footnote{$\return_{\alpha\policy + (1-\alpha)\policy'}(\xlandtask) := \alpha \return_{\policy}(\xlandtask) + (1-\alpha)\return_{\policy'}(\xlandtask)$} of policies from a corresponding population $\population_t$. \figref{fig:normpercentiles} illustrates this process. Each agent's performance is described as 51 numbers between 0 and 1, with each number being the normalised score at each integer percentile in the range of 0 to 50 (inclusive), which forms a non-decreasing sequence $$\perc(\policy)[k+1] \geq \perc(\policy)[k].$$

One can read out various human interpretable quantities from this representation, e.g. $\perf(\policy)[0]$ is the infimum -- the normalised score an agent obtains on the hardest game; $\perf(\policy)[50]$ is the median normalised performance; the smallest $k$ such that $\perf(\policy)[k] > 0$ informs us that an agent scores any reward in at least $(100-k)\%$ of games (and thus provides a notion of coverage/participation).

We say an agent $\policy$ is better than agent $\policy'$ if and only if it achieves at least as good a score for every percentile, and on at least one percentile it achieves a strictly better score, formally:
\begin{equation*}
\begin{aligned}
\policy \succeq_{\population_t} \policy' \iff& \forall_k \perc(\policy | \population_t)[k] \geq \perc(\policy'| \population_t)[k]\\
\policy \succ_{\population_t} \policy' \iff& \exists_k \perc(\policy | \population_t)[k] > \perc(\policy' | \population_t)[k] \\&\wedge \policy \succeq_{\population_t} \policy'.
\end{aligned}
\end{equation*}

Let us refer to our desiderata -- if agent $\policy$ fails catastrophically (never achieves any reward) on fewer tasks than $\policy'$ then it will have non-zero values on a larger number of percentiles, and thus captured in our notion of being better. Conversely, if catastrophic failures are more common, then $\policy$ will not be considered better (it can be non-comparable or worse). The notion of competency refers to the fraction of the score obtained by the Nash equilibrium over known policies, and thus similarly by being competent on more tasks, $\policy$ will increase its values on smaller percentiles. Finally, a narrow competency will be visible in low scores over low percentiles, and despite high scores being obtained on high percentiles -- such an agent will not be considered better. In addition, cutting our percentiles at 50 means that an agent that is an expert on less than half of the games, but does not score any points on remaining ones, will be considered worse than an agent of broader ability. 

To summarise, we propose to use the following tools to measure and drive progress of general capabilities of agents:
\begin{itemize}
    \item to normalise performance by the estimated highest obtainable score,
    \item to iteratively improve the estimate of the highest obtainable score,
    \item to evaluate agents across normalised score percentiles, creating a multi-dimensional performance descriptor,
    \item to require Pareto dominance over said descriptor to guarantee improvements with respect to our desiderata.
\end{itemize}

\subsection{Evaluation Task Set}
\label{sec:evalset}
The normalised percentile metric described in the previous section provides a way to compare agents and drive learning with a lens towards general capability. However, this metric is still evaluated with respect to a distribution of tasks $\mathrm{P}_\xlandspace$. The XLand task space as defined in \secref{sec:environment} is prohibitively large, and as such we need to create a manageable evaluation task set against which to assess agents' general capability.

Given a budget number of evaluation tasks (\emph{e.g.} on the order of thousands), arbitrarily sampling tasks from $\xlandspace$ could risk critically underrepresenting the vastness and diversity of the underlying task space, with aliasing also hiding the smoothness property. As such, we define an evaluation task space that samples tasks spanning a smaller but representative subspace of XLand tasks, and skew sampling to ensure uniform coverage of interesting world and game features. Finally, we combine these evaluation worlds and games with pretrained evaluation policies to give us an evaluation task set.

\paragraph{Evaluation worlds}
For evaluation, we want a set of worlds that expose agents to a range of topological challenges. We use a world-agent co-evolution process (\secref{sec:app-procgen}, \figref{fig:worldevo}), saving the training distribution of worlds created at each point in training of this process. This gives a collection of worlds where the earlier-created worlds are generally topologically simpler than those created later in training. Uniformly sampling this collection of worlds with respect to the creation time gives a set of worlds spanning the range of topological complexity (as defined by an agent learning to find an object). We also randomly apply reflections and resampling of object positions to this set of worlds. Finally, we add additional Wave Function Collapse~\citep{wavefunctioncollapse} generated worlds, biased towards specific topological elements that we observe rarely: ones containing single central islands and door-like bottlenecks separating play areas. The gadget of each player is uniformly sampled and the colour ordering of each player randomly permuted. Exactly 12 objects are placed into each evaluation world, one of each colour-shape combination.

\paragraph{Evaluation games}
In the game space, we look to create a set of evaluation games that span a large range of complexity and expressivity, but are still logically simple enough for quick human understanding. Therefore, representing the goals of the game in their \emph{disjunctive normal form}, we restrict the evaluation games to have at most three options per goal, with each option composed of at most three predicates, and a maximum of six unique predicates used across all goals. Only two- and three-player games are considered in the evaluation set. Additionally, we ensure the evaluation set of games spans the range of competitiveness and balance (defined in \secref{sec:envproperties}) -- we create discrete buckets in competitiveness-balance space, with some buckets corresponding to the extreme values of these measures. Evaluation games are sampled such that there is an equal number of games per competitiveness-balance bucket, and per competitiveness-balance bucket an equal number of games across the different number of options and predicates in the game. We also remove trivial games (\emph{i.e.} where $\ed({\goal_1}) \in \{0,1\}$). The result is an evaluation set of games which is uniform across balance buckets, competitiveness buckets, number of options, and number of predicates.

\paragraph{Evaluation co-players}
Each evaluation task must include policies to act as the co-players of the task, leaving a single player slot available for evaluation of an agent in the task. For the purposes of this work, we use a collection of pretrained agents. These include a \texttt{noop} agent that always emits the noop action (corresponding to not moving) and a \texttt{random} agent that emits an action uniformly sampled from the whole action space. In addition, we use agents trained on simpler incarnations of the evaluation space, as well as sub-spaces of evaluation space (\emph{e.g.} an agent trained only on single predicate games). These agents were generated during earlier phases of the research project.

We combine the evaluation worlds, games, and co-players to get \evalvalid{} and \evaltrain{} sets. We first generate the \evalvalid{} set of evaluation tasks. Next the \evaltrain{} set of evaluation tasks is generated in an identical manner, however explicitly holding out all games and worlds within a certain distance from the \evalvalid{} set (\secref{sec:holdout}) and likewise holding out all \evalvalid{} set co-players except for the trivially generated \texttt{noop} and \texttt{random} policies. In addition, all hand-authored tasks (\secref{sec:handauthored}) are held out from all evaluation task sets. The \evalvalid{} task set consists 1678 world-game pairs played with all 7 co-players for a total of 11746 tasks. The \evaltrain{} task set consists of 2900 world-game pairs played with a growing number of co-players: \texttt{noop}, \texttt{random} and an extra player per previous generation of training.

\begin{figure*}[t]
    \centering
    \includegraphics[width=\linewidth]{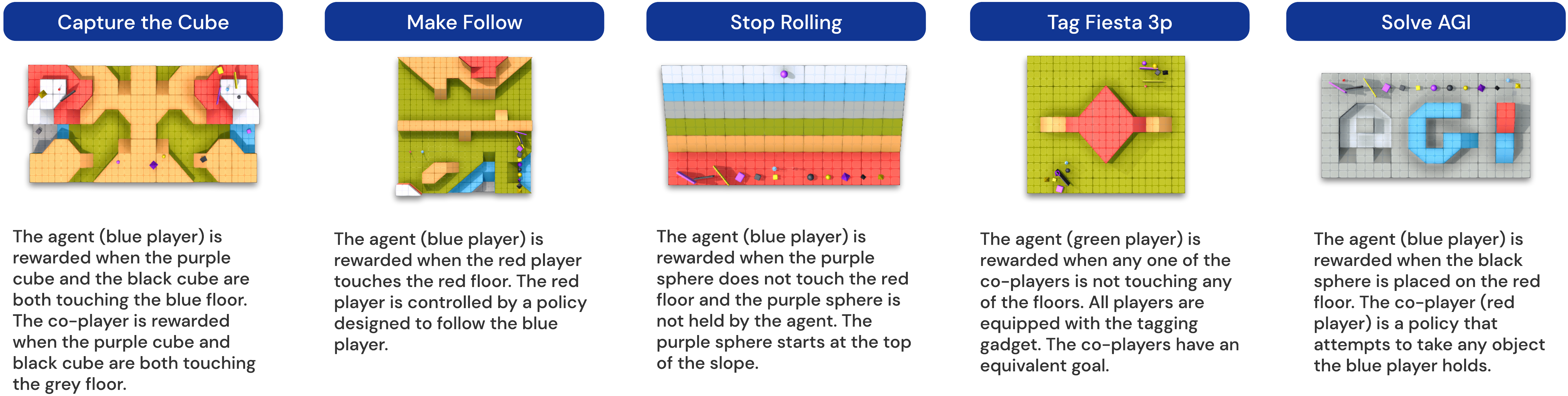}
    \caption{Five examples of some of the 42 tasks in the hand-authored evaluation task set. A full list of hand-authored evaluation tasks is given in Table~\ref{tab:handauthoredlist2}.}
    \label{fig:handauthored-tasks}
\end{figure*}

\subsection{Hand-authored Task Set}
\label{sec:handauthored}
The evaluation set of tasks described previously covers a diverse subspace of XLand, however the automatic generation of these tasks can make interpretation of successful policies difficult -- it can be hard to know what challenges an individual task poses. We created a hand-authored set of tasks which act as interpretable evaluation tasks. In addition, many of these hand-authored evaluation tasks are out-of-distribution or represent challenges that are extremely rare to be seen in a sample from the evaluation set, and thus further test ability of an agent to generalise. Examples of the 42 tasks in the hand-authored task set can be found in \figref{fig:handauthored-tasks} (full list is provided in Table~\ref{tab:handauthoredlist}), and include well known tasks such as \emph{Capture the Flag}, \emph{Hide and Seek}, and \emph{King of the Hill} which have been projected into XLand space. Other examples include physical challenges such as \emph{Stop Rolling} and \emph{Tool Use}. The hand-authored task set is also held out from all training.

\section{Learning Process}
\label{sec:training}
We now turn our attention to the learning process. We seek agents that are generally capable in the XLand space. 
As one of the proxies to this goal, we want agents that can zero-shot generalise to tasks from the \evalvalid{}{} set, and use normalised percentiles computed on the \evalvalid{} set as the performance metric to encapsulate this. 

\noindent Our training process consists of three main components: 
\begin{enumerate}
    \item Deep RL to update the neural network of a single agent. Deep RL optimises an agent to maximise expected return across a distribution of training tasks given.
    \item Dynamic task generation with population based training to provide the distribution of training tasks for a population of agents. The task distributions are changed throughout training and are themselves optimised to improve the population's normalised percentiles on the \evaltrain{} set.
    \item Generational training of populations of agents to chain together multiple learning processes with different objectives. Agents are trained with different learning objectives per generation, with each subsequent population bootstrapping behaviour off the previous generation of agents, to improve \evaltrain{} normalised percentiles with each subsequent generation.
\end{enumerate}
\noindent We will now describe these three components in more detail.

\subsection{Deep Reinforcement Learning}
\label{sec:agent}
An agent playing on an XLand task $\xlandtask$ takes in high-dimensional observations $\observation_t$ at each timestep, and produce a policy from which actions are sampled $\mathbf{a}_t \sim \policy_t$, allowing the agent to maximise the collected reward on the task. We use a neural network to parameterise the policy, and train this network using the V-MPO RL algorithm~\citep{vmpo}. Similarly to the the original V-MPO implementation, we use single-task PopArt normalisation of the value function~\citep{popart1, hessel2019multi}. At each weight update,
the network parameterising $\policy$ is updated in the direction to maximise the expected discounted return on the instantaneous task distribution $\mathbf{V}_\policy(\mathrm{P}_\policy)$. 

The per-timestep observation $\observation_t$ the neural network takes as input consists of an RGB image from the agent player's point-of-view $\observation^\text{RGB}_t$, proprioception values corresponding to the forces relating to the agent's player holding an object $\observation^\text{prio}_t$, as well as the goal $\goal$ of the agent's player in the task $\xlandtask$.
A recurrent neural network processes this information to produce a value prediction $\valuehead_t$ and policy $\policy_t$, from which a single action $\mathbf{a}_t$ is sampled.

\paragraph{Goal attention network}
The recurrent neural network incorporates an architecture that is tailored towards the structure of $\mathbf{V}^*$ the value of an optimal policy for a given task. For simplicity let us write
$$
\mathbf{V}^*(\goal) :=  \max_\policy \mathbf{V}_\policy(\goal)
$$ 
to denote the value of the optimal policy when we hold the world, other goals, and co-players fixed.
\begin{thmrep}[Value Consistency]
\label{thm:value_consistency}
For a goal $\goal := \bigvee_{o=1}^k [\bigwedge_{c=1}^{n_o} \predicate_{oc}]$ we have 
$$
\mathbf{V^*}(\goal_l)
\leq
\mathbf{V^*}(\goal)
\leq
\mathbf{V^*}(\goal_u)
$$
for
$
\goal_l := \bigvee_{o=1}^{k-1} \left [\bigwedge_{c=1}^{n_o} \predicate_{oc} \right ],
\goal_u := \bigvee_{o=1}^{k} \left [\bigwedge_{c=1}^{n'_o} \predicate_{oc} \right ]
$
where $n'_o \geq n_o$.
\end{thmrep}
\begin{proof}
Since $\goal_l$ differs from $\goal$ by simply missing the $k$th option, this means that the corresponding reward function 
$$
\reward_{\goal_l}(\state) = \max_{o=1}^{k-1} \left [ \min_{c=1}^{n_o}\predicate_{oc}(\state) \right ] \leq \max_{o=1}^{k} \left [ \min_{c=1}^{n_o}\predicate_{oc}(\state) \right ] = \reward_\goal(\state).
$$
Consequently
$\mathbf{V^*}(\goal_l)
\leq
\mathbf{V^*}(\goal).
$
Analogously, $\goal_u$ differs from $\goal$ by potentially having additional predicates in each options, this means that the corresponding reward function 
$$
\reward_{\goal}(\state) = \max_{o=1}^{k} \left [ \min_{c=1}^{n_o}\predicate_{oc}(\state) \right ] \leq \max_{o=1}^{k} \left [ \min_{c=1}^{n'_o}\predicate_{oc}(\state) \right ] = \reward_{\goal_u}(\state).
$$
Consequently,
$
\mathbf{V^*}(\goal)
\leq
\mathbf{V^*}(\goal_u).
$
\end{proof}
This property says that for each game we can easily construct another game providing an upper or lower bound of the optimal value, by either selecting a subset of options or a superset of conjunctions.
Therefore, with $\goal := \option_1 \vee \dots \vee \option_{n_o}$ we have $$\mathbf{V}^*(\goal) \geq \max_{i=1}^{n_o} \mathbf{V}^*(\option_i).$$ 
By putting $\option_0 := \goal$ we can consequently say that
$$
\mathbf{V}^*(\goal) =  \max_{i=0}^{n_o} \mathbf{V}^*(\option_i),
$$
the optimal value for $\goal$ is the maximum of values of the subgames consisting of each of the options $\option_i$ and the full goal $\goal$ itself.

\begin{figure*}[t]
    \centering
    \includegraphics[width=\textwidth]{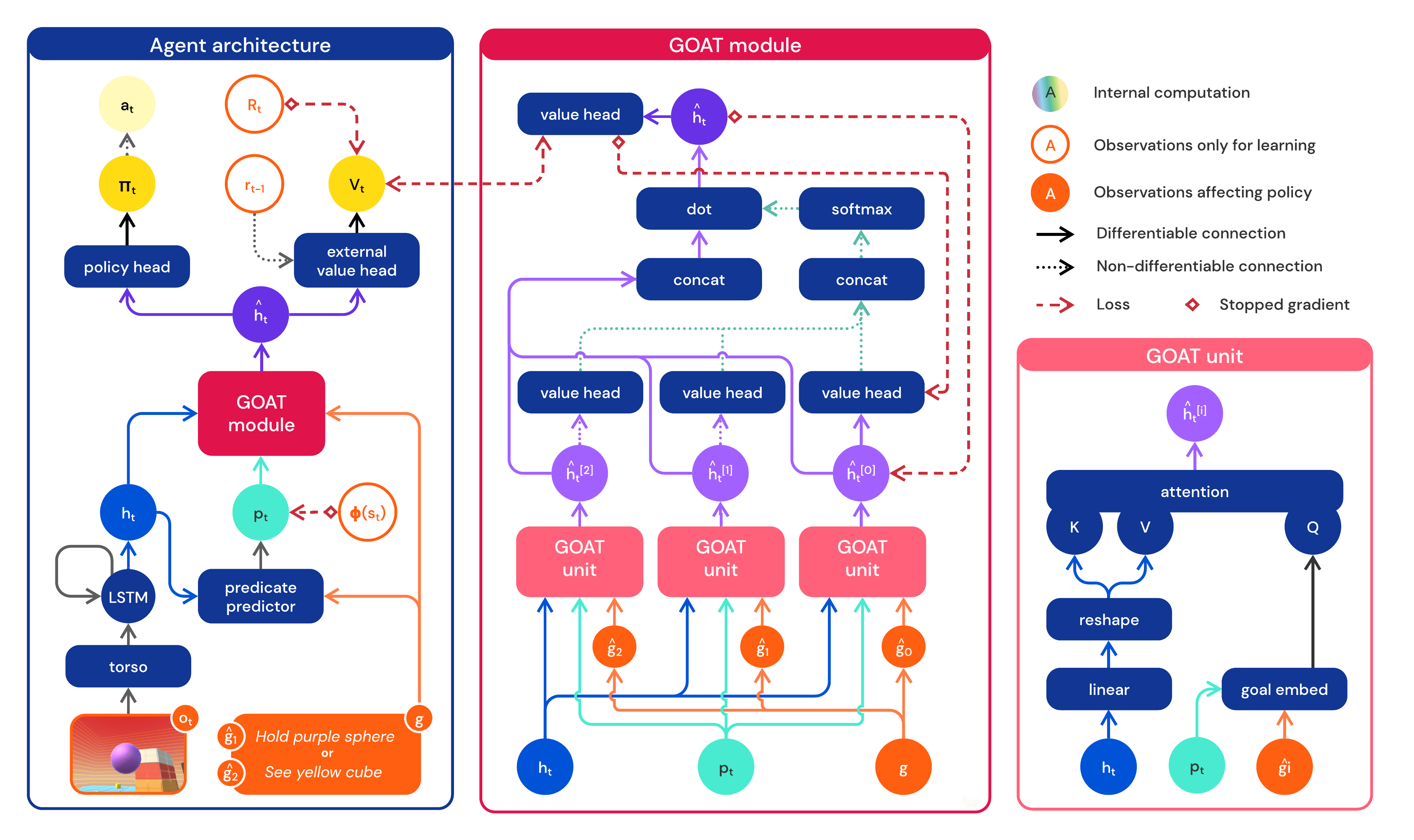}
    \caption{A schematic of the neural network used to parameterise the agent's policy. The input observations $\observation_t$ consist of RGB images and the proprioception, and the agent also receives its goal $\goal$. The agent processes the observations through the torso and a recurrent core to produce $\mathbf{h}_t$, which is used for the predicate predictor, producing $\mathbf{p}_t$. The recurrent core output, the predicate predictor output, and the goal is passed to the GOAT module. The GOAT module (see \secref{sec:agent}) attends to a specific part of the recurrent representation based on the current goal of the agent, and performs logical analysis of the goal using \emph{value consistency} (see Theorem~\ref{thm:value_consistency}). The goal embedding and predicate predictor architectures are provided in \figref{fig:goal_network}. Modules with the same names share weights (\emph{i.e.} each value head, as well as each GOAT unit). }
    \label{fig:network}
\end{figure*}
We encode this property explicitly in the neural network architecture. At each timestamp, the agent produces an internal hidden state embedding using the history of its observations but no knowledge of the goal. Separately, the goal is embedded and used as a query for an attention mechanism that produces a goal-attention hidden state $\widehat{\mathbf{h}}_t^{[0]}$. In parallel, the agent produces analogous embeddings for each option $\widehat{\mathbf{h}}_t^{[i]}$, and estimates the current value of each $\goatvaluehead_t^{[i]}$. This asks the agent to predict \emph{what would be its expected return if it was to focus on option $i$ until the end of the episode}, $\goatvaluehead_t^{[i]}$.
With the attention mechanism, the agent switches its hidden state to another option's hidden state if and only if the value of said option was higher than its current estimate of the value of the whole goal. This way the agent is internally encouraged to be consistent with the \emph{Value consistency} property of the game space.

More specifically, $\mathbf{h}_t$ is the hidden state of an LSTM~\citep{hochreiter1997long} that takes as input the processed pixel and proprioception observations. We attach an atomic predicate state predictor to the output of the LSTM: this predictor is a simple multi-task binary classifier, outputting $\mathbf{p}_t$ which predicts the dimensions of $\predicate(\state_t)$ relevant for $\goal$, and is trained as an auxiliary classification loss (\emph{i.e.} only shaping the internal representations, without explicitly affecting the policy~\citep{jaderberg2016reinforcement}). The \emph{goal attention (GOAT)} module then follows:
\begin{equation*}
\begin{aligned}
\mathbf{GOAT}(\mathbf{h}_t, \mathbf{p}_t, \goal) &:= \sum_i \stopgradient{\alpha(\goatvaluehead^{[i]}, \{ \goatvaluehead^{[j]} \}_{j=0}^{o} )} \widehat{\mathbf{h}}_t^{[i]}\\
\goatvaluehead^{[i]}_t &:= \goatvalue(\widehat{\mathbf{h}}_t^{[i]}) \;\;\;\;\;\; \forall_{i=0}^o\\
\widehat{\mathbf{h}}_t^{[i]} &:= \mathbf{GOAT}_\mathrm{unit}(\mathbf{h}_t, \mathbf{p}_t, \option_i) \;\;\;\;\;\; \forall_{i=0}^o\\
\mathbf{GOAT}_\mathrm{unit}(\mathbf{h}, \mathbf{p}, \goal) &:= \goatattention(\mathbf{h}, [\mathbf{p} ;f_\goalspace(\goal)]),
\end{aligned}
\end{equation*}
where $\goatattention(\cdot, \cdot)$ is an attention module~\citep{bahdanau2014neural}, $\stopgradient{ \cdot }$ denotes the stop gradient operation, meaning $\nabla_x \stopgradient{ x } = 0$, and $\alpha(a, A)$ is a weighting scheme, \emph{e.g.}: $\alpha_\mathrm{argmax}(a,A) = 1 \iff a=\max\{A\}$, or $\alpha_\mathrm{softmax, Z}(a,A) := \tfrac{\exp(a/Z)}{\sum_{b \in A} \exp(b/Z)}$.

Given this parameterisation, we add corresponding \emph{consistency losses}:
\begin{equation*}
\begin{aligned}
\ell^\mathbf{V}_t :=  \left ( \stopgradient{\goatvaluehead_t} - \goatvaluehead_t^{[0]} \right )^2 \;\;\;\;\;\;\;
\ell^\mathbf{h}_t := \left ( \stopgradient{\widehat{\mathbf{h}}_t} - \widehat{\mathbf{h}}_t^{[0]} \right )^2,
\end{aligned}
\end{equation*}
where $\widehat{\mathbf{h}}_t := \mathbf{GOAT}(\mathbf{h}_t, \mathbf{p}_t, \goal)$, $\goatvaluehead_t := \goatvalue(\widehat{\mathbf{h}}_t)$, and $f_\gamespace$ is the goal embedding function (see \figref{fig:goal_network}).
These losses encourage the value predicted for the full goal $\goal$ to not be smaller than the value of any of the options $\option_i$. 
A schematic view of this process is provided in \figref{fig:network} with the details of the goal embedding $f_\goalspace$ and atomic predicate predictions provided in \figref{fig:goal_network}.
Note, that these are all internal value functions predictions that do not use any privileged information. We observed that faster learning can be achieved if the value function $\valuehead$ used for RL itself does get access to extra information~\citep{vinyals2019grandmaster} in the form of the reward from the last step $\reward_{t-1}$. We add a simple L2 loss to align these two value heads in a co-distillation manner~\citep{codistillation_cited}:
$$
\ell^\mathrm{align} := \| \valuehead_t - \goatvaluehead_t \|^2.
$$
We do not stop gradients through $\valuehead$ meaning that the privileged information value head $\valuehead$ is penalised for expressing quantities that the internal $\goatvaluehead$ cannot model as well.

\subsection{Dynamic Task Generation}
Due to the vastness of task space, for any given agent, many tasks will either be too easy or too hard to generate good training signal. To tackle this, we allow the train task distribution to change throughout the learning process in response to the policy of the agent itself. The agent's neural network is trained with RL on the instantaneous training task distribution 
$\mathrm{P}_\policy(\xlandspace)$.

We operationalise 
$\mathrm{P}_\policy(\xlandspace)$
by using a filtering of a proposal distribution using a simple set of tests evaluating tasks usefulness for the current stage of learning.

Proposal train tasks are generated in a similar manner to the evaluation \evaltrain{} task set: worlds, games, and co-players are generated as described in \secref{sec:evalset} ensuring no collisions with the \evaltrain{} and \evalvalid{} sets (\secref{sec:holdout}).
We establish a task's usefulness by comparing the performance of the agent to the performance of a control policy $\policy_\mathrm{cont}$.

The intuition of using a control policy is that the agent will only train on a task if the agent's returns are significantly better than those of the control policy. This guarantees that the agent is performing meaningful actions in the task that affect the return. In practice, we set the control policy to be a uniform random action policy. However, an interesting alternative would be to set the policy to be the agent's past policy -- this would let us determine whether the agent's policy has recently improved or worsened on this task.

A proposal task is accepted (used for training) if and only if the following three criteria are met:
\begin{enumerate}
    \item The agent has a low probability of scoring high on a given task
    $$\mathrm{Pr}[\return_\policy(\xlandtask) > m_s] < m_\mathrm{solved}.$$
    \item The agent has a high probability of performing better than the control policy
    $$\mathrm{Pr}[\return_\policy(\xlandtask) > \return_{\policy_\mathrm{cont}}(\xlandtask) + m_>] > m_{>\mathrm{cont}}.$$
    \item The control policy is not performing well
    $$
    \mathbf{V}_{\policy_\mathrm{cont}}(\xlandtask) < m_\mathrm{cont}.
    $$
\end{enumerate}

At a high level, the filtering of proposal tasks gives a mechanism for removing tasks that are too-easy (criterion 1), tasks that are too-hard (criterion 2), and tasks in which the control policy is sufficient to achieve a satisfactory score (criterion 3), based on the agent's current behaviour at each point in training. 
All the above parameters $\mathbf{m}=\{m_>, m_\text{s}, m_\text{cont}, m_{>\text{cont}}, m_\text{solved}\}$ form agent-specific hyperparameters that define $\mathrm{P}_\pi(\xlandspace)$. We estimate the criteria using Monte Carlo with 10 episode samples for each policy involved.

For example, a control policy return threshold $m_\text{cont}=5$ would disallow any training tasks where a control policy is able to get a return of at least 5 (the reward is on a scale of 0 to 900). When using a uniform-random policy over actions as the control policy, this could be used to ensure the training task distribution doesn't contain tasks that are very easy to get reward. The combination, for example, of $m_>=2$ and $m_{>\text{cont}}=0.9$ would only allow training tasks where the agent achieves a return in all ten episode samples of at least 2 reward more than the return achieved by the control policy -- this could ensure that the agent only trains on tasks where its behaviour is already better than that of the control policy. As a final example, the combination of $m_\text{s}=450$ and $m_\text{solved}=0.1$ would disallow training on any task where the agent is able to achieve more than 450 reward on any of its episode samples -- this could filter out tasks where the agent is already performing well.

Whilst this filtering mechanism provides a way to supply the agent with a dynamic training task distribution, the filtering criterion itself may benefit from being dynamic. What is considered too-hard or too-easy at the beginning of training may encourage early learning, but cause saturation or stalling of learning later in training. Due to the vastness of the XLand task space we seek learning processes that do not saturate, but rather dynamically shift to ensure the agent never stops learning.

To address this, we incorporate population based training (PBT)~\citep{jaderberg2017population} which provides a mechanism to dynamically change hyperparameters of the learning process~\citep{jaderberg2019human}. Rather than training a single agent, we train a population of agents, each agent training on its own task distribution $\mathrm{P}_{\policy_k}(\xlandspace)$ that is controlled by its own hyperparameters  $\mathbf{m}_k$. Additionally, the learning rate and V-MPO hyperparameter $\epsilon_\alpha$ are added to the set of hyperparameters modified by PBT. 

PBT requires a fitness function to compare two agents and propagate the preferred agent. We use the normalised percentiles on the \evaltrain{} set. Periodically during training, agents are compared, and only if an agent Pareto dominates another agent in normalised score across percentiles it undergoes evolution -- the dominant agent's weights are copied, its instantaneous task distribution copied, and the hyperparameters copied and mutated, taking the place in training of the non-dominant agent. More details can be found in~\secref{app:pbt}.

This process constantly modifies the dynamic task generation process and agent population to drive iterative improvement in normalised percentiles.

\subsection{Generational training}
\label{sec:gentraining}

With this combination of deep RL and dynamic task distributions we hope to provide a training process to continually improve agents in terms of their normalised percentiles as measured on the \evaltrain{} task set. However, in practice, the limitations of RL and neural network training dynamics still pose a challenge in training agents on the XLand task space from scratch.

It has been observed that higher performance and faster training can be achieved in deep RL by first training an agent, then subsequently training a new agent on the identical task whilst performing policy distillation from the first agent~\citep{furlanello2018born,schmitt2018kickstarting,czarnecki2019distilling}. We employ this technique multiple times on populations of agents: a population of agents is trained, then a new generation of agents is trained distilling from the best agent of the previous generation's population, with this process repeated multiple times. Each generation bootstraps its behaviour from the previous generation. Furthermore, these previous generations also give us an opportunity to increase our pool of co-player policies and increase the diversity of our training experience, similarly to the AlphaStar league~\citep{vinyals2019grandmaster}. At each generation, our training procedure includes the best player from each previous generation in this pool of players. 

A final advantage of generational training of populations is that the learning objectives and agent architecture can vary generation-to-generation. We take advantage of this by using self reward-play: an RL objective which encourages exploration. In our training methodology, self reward-play is utilised for the first few generations followed by the regular RL objective in the later generations. 

\paragraph{Self reward-play}
One of the central desires of a generally capable agent is that the agent should catastrophically fail on as few tasks as possible. To target this objective we seek agents that minimise the smallest non-zero normalised percentile -- to obtain at least one timestep of reward in as many tasks as possible, the problem of exploration. We define \emph{participation} as the percentage of tasks the agent obtains a non-zero reward in. 

To aid learning participation, we present challenges to the agent that it is capable of satisfying by asking the agent to revert a changes in the environment that the agent itself previously created. Self reward-play rewards the agent for satisfying a goal $\goal$, and after succeeding the agent is rewarded for fulfilling $\texttt{not}(\goal)$ without resetting the environment, with this flip in goal repeating after each satisfaction. This can be seen as an agent playing in a self-play competitive manner against itself, where one player must satisfy $\goal$ and the other player must satisfy $\texttt{not}(\goal)$, however the players act sequentially, and are played by the same agent.

In practice, we implement this by using the reward $\reward^\mathrm{srp}_t := |\reward_t - \reward_{t-1}|$ and setting the discount $\gamma_t = 0$ if $\reward^\mathrm{srp}_t > 0$ (which rewards the agent for minimising the time until the next goal flip).

Empirically, we find that optimising for self reward-play drastically improves exploration. The agent is encouraged to interact with the world and to change its reward state, after which it must change the state back again, and so on. In comparison, when optimising the discounted sum of environment reward, changing the environment yields the risk of changing the (unobserved) environment reward from 1 to 0 which discourages the agent from interacting with the environment. As a result, agents that optimise with self reward-play achieve significantly higher participation in the same amount of training time (see \secref{sec:ablations}). However, by construction, self reward-play does not optimise agents to be competent (\emph{i.e.} whilst the smallest non-zero normalised score percentile is minimised, the normalised percentiles remain low). We discuss in detail how self reward-play is leveraged in \secref{sec:agent-training}.

\paragraph{Iterative normalised percentiles}
As discussed in \secref{sec:evalset}, the \evalvalid{} set contains a fixed set of co-player policies (used also to evaluate against). However, the \evaltrain{} set does not contain these, but only the trivially generated noop-action and random-action policies. For evaluation, co-player policies are required to play \evaltrain{} tasks with, and the normaliser score used by the normalised percentile metric also uses this fixed set of co-player policies. The generational training process allows us to start only with the trivially generated noop-action and random-action policies and to iteratively refine the \evaltrain{} normalised percentiles metric: each generation creates agents which are added to the \evaltrain{} set and used to update the normalised percentile metric, with the next generation incorporating the previous generation's policies in its training, with this process repeating, iteratively refining the normalised percentile metric and expanding the set of co-player policies. This means that the normalised percentiles metric on the \evaltrain{} set used to guide training changes each generation as more policies are added to the \evaltrain{} co-player set. Note that for all results reported in \secref{sec:results}, we report the normalised percentiles on the \evalvalid{} set which is fixed, with the same fixed set of co-player policies, for all generations.

\subsection{Combined Learning Process}
These three components of the training process -- deep reinforcement learning, dynamic task generation, and generational training -- are combined to create a single learning process. The three pieces are hierarchically related. On the smallest wall-clock timescale (seconds), deep RL provides weight updates for the agents' neural networks, iteratively improving their performance on their task distributions. On a larger timescale (hours), dynamic task generation and population based training modulate the agents' task distributions to iteratively improve the Pareto front of \evaltrain{} normalised percentile scores. Finally, on the largest timescale (days), generational training iteratively improves population performance by bootstrapping off previous generations, whilst also iteratively updating the \evaltrain{} normalised percentile metric itself.

\begin{figure*}
    \centering
    \includegraphics[width=0.7\linewidth]{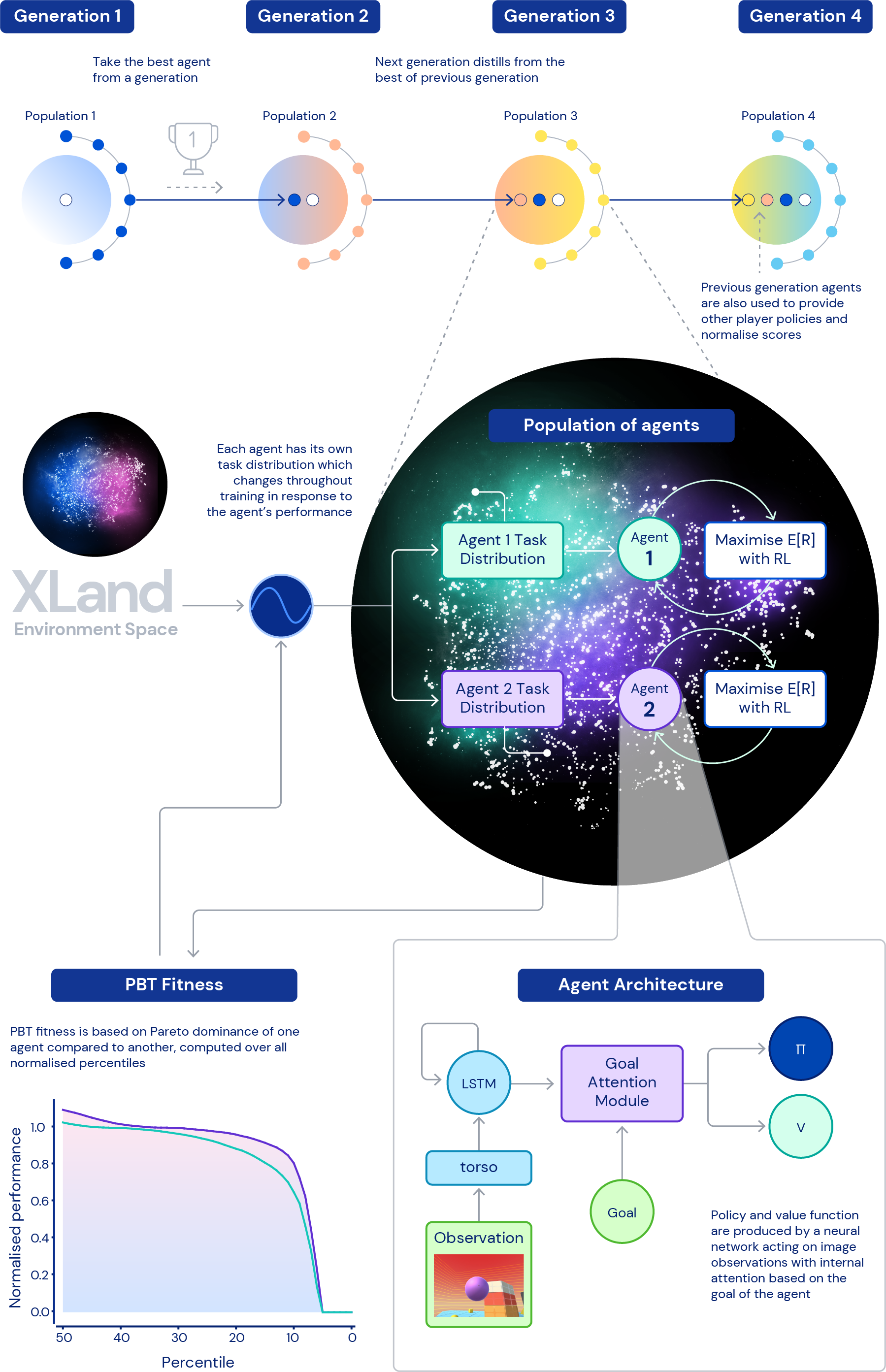}
    \caption{The combined learning process. \textbf{(Top)} Generations of agents are trained, composed of populations of agents where the best performing agents become distillation teachers of the next generation as well as co-players to train against.
    \textbf{(Middle)} Inside each population, agents are trained with dynamic task generation that continuously adapts the distribution of training tasks  $\mathrm{P}_{\policy_k}(\xlandspace)$ for each agent $\policy_k$, and population based training (PBT) modulates the generation process by trying to Pareto dominate other agents with respect to the normalised percentiles metric.
    \textbf{(Bottom)} Each agent trains with deep reinforcement learning and consists of a neural network producing the policy $\policy$ and value function $\valuehead$.}
    \label{fig:combined_learning}
\end{figure*}
\begin{figure*}
\includegraphics[width=\textwidth]{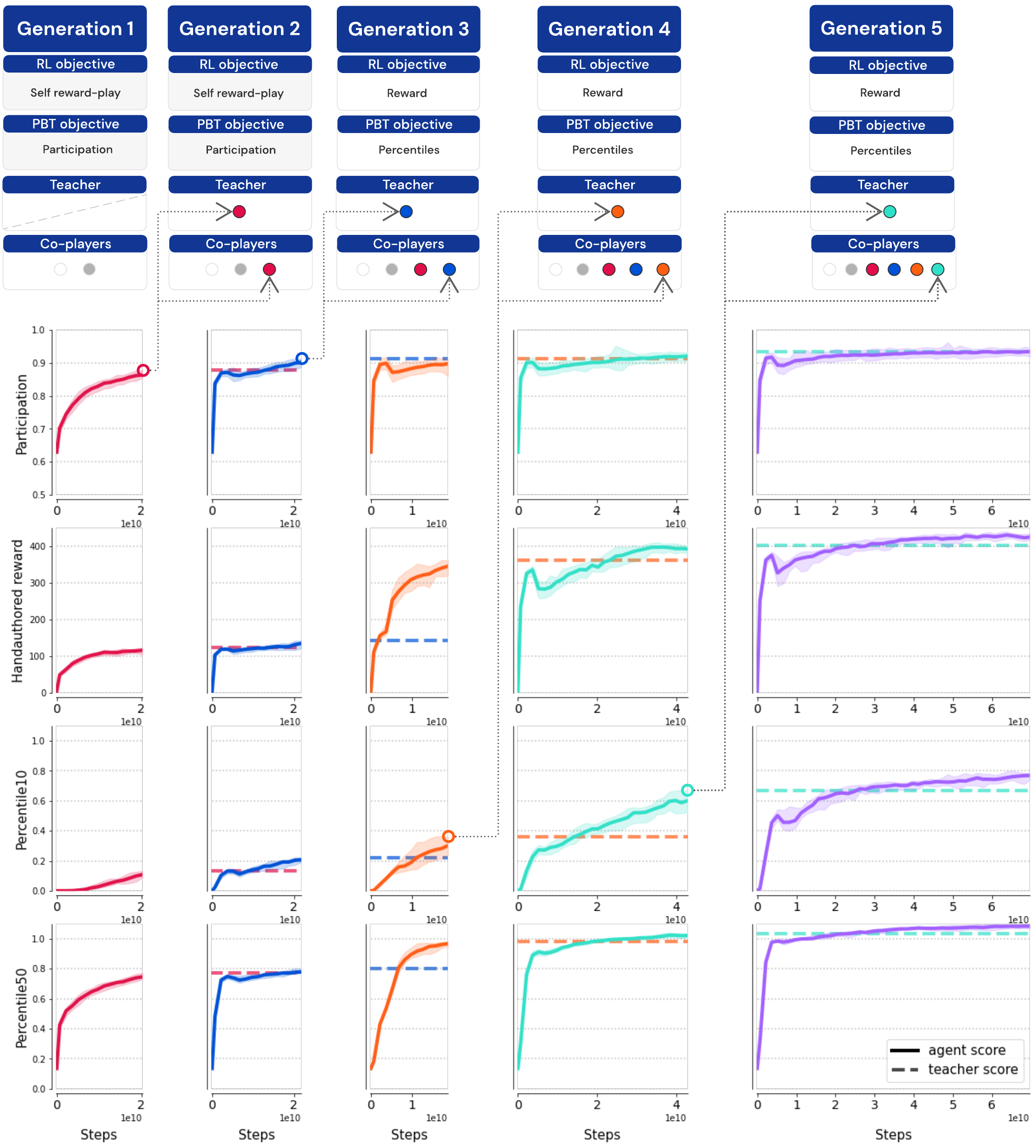}
\caption{Generations of performance as measured on the held out \evalvalid{} task set. 
The first two generations focus on the maximisation of participation using the self reward-play RL objective (\secref{sec:gentraining}). In between generations, the best agent wrt. the objective is selected and used as a teacher and additional co-player to play against in further generations. Generations 3-5 focus on the improvement of normalised percentiles, and use the raw reward for the RL algorithm. The dashed line in each plot corresponds to the performance of the teacher from the previous generation. The co-players are the set of policies that populate the co-players in these multiplayer tasks, with this set initialised to just the trivially created noop-action and random-action agents (white and grey circles).
}
\label{fig:generations}
\end{figure*}
\begin{figure*}[t]
\centering
\includegraphics[width=0.9\textwidth]{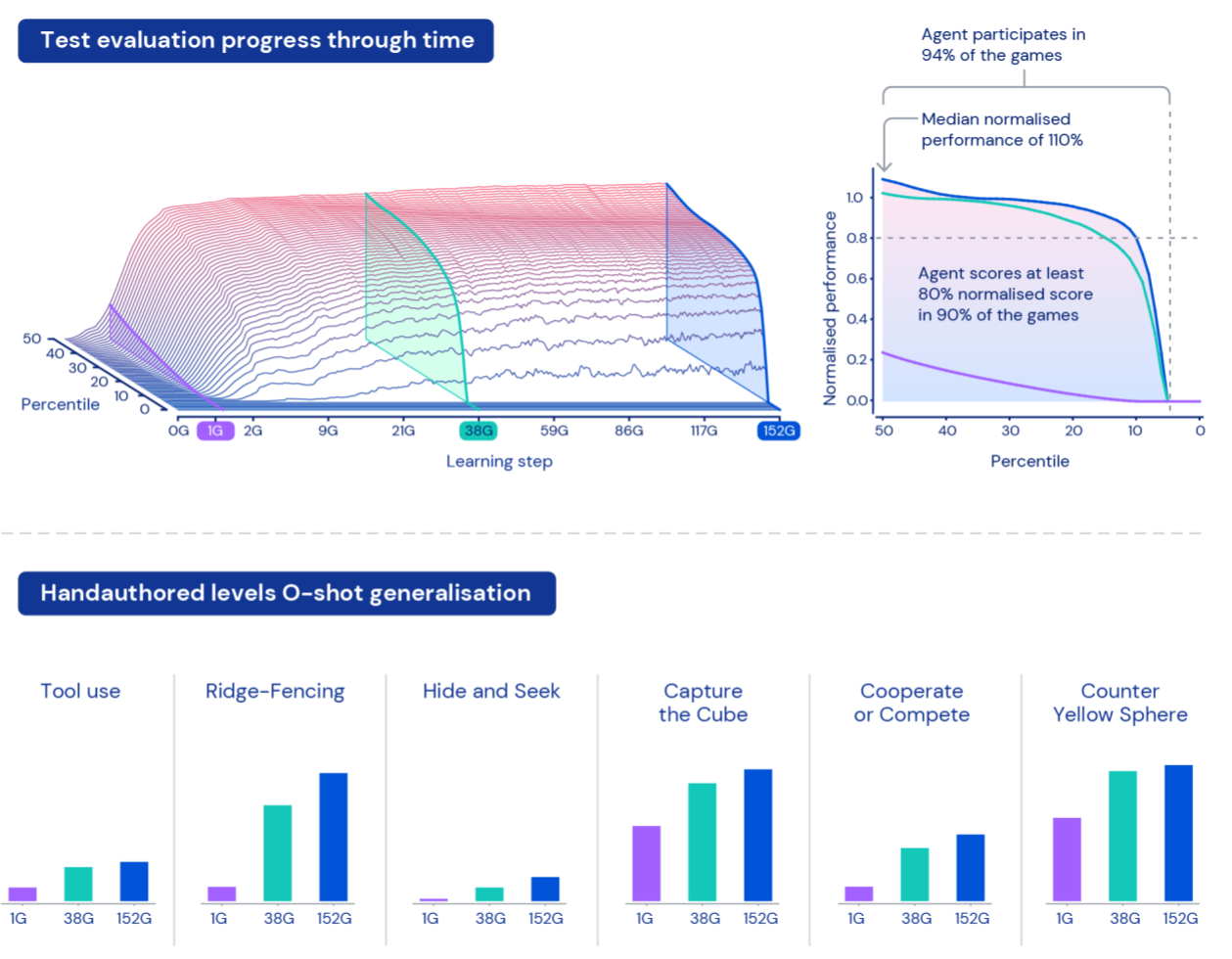}
\caption{\textbf{(Top)} On the left we see the \emph{learning surface}, showing the progress of a generation 5 agent through time with respect to each of the normalised percentiles. The surface shows the normalised score (height) for each percentile (depth) through training (x-axis). Therefore, the flat bottom of the surface (zero height) is the part of the space where the agent is not participating. On the right, we see an orthogonal projection onto the surface at the end of training. \textbf{(Bottom}) We highlight the performance on 6 hand-authored tasks at three points in training, showing how improvements in the normalised percentiles correspond to improvement in these hand-authored tasks.}
\label{fig:gen5-learning}
\end{figure*}

From the opposite perspective, the overall system continuously creates generations of agents seeking to improve the \evaltrain{} normalised percentile metric -- to gradually improve the coverage and competence on tasks. In order to do so, a generation's population is changing the distribution of training tasks for each agent such that the agents keep improving the Pareto front of \evaltrain{} normalised percentile scores. The agents themselves are updating their neural network weights with reinforcement learning based on the stream of experience they generate from their training task distributions, gradually improving their performance on this shifting distribution. The whole process is summarised in \figref{fig:combined_learning}.

The iterative nature of the combined learning system, with the absence of a bounded metric being optimised, leads to a potentially open-ended learning process for agents, limited only by the expressivity of the environment space and the agent's neural network.

\section{Results and Analysis}
\label{sec:results}

In this section, we show the results of training agents with the learning process introduced in \secref{sec:training}, with the specific experimental setup described in \secref{sec:resexpsetup}. The learning dynamics are explored in \secref{sec:restraining} with respect to the evaluation metric defined in \secref{sec:goal}. In \secref{sec:resperformance}, we analyse the zero-shot generalisation performance of the trained agent across the \evalvalid{} set. \secref{sec:res-capabilities} delves into some emergent agent behaviour that is observed on hand-authored probe tasks. Moving beyond zero-shot behaviour, in \secref{sec:finetuning} we show the results of finetuning the trained agents for wider transfer. Finally, in \secref{sec:representation} we analyse the representations formed by the agent's neural network.

All the results reported in this section are computed on tasks that were held-out of training.

\subsection{Experimental Setup}
\label{sec:resexpsetup}
More details on the architecture, hyperparameters, other elements of the experimental setup are provided in \secref{app:rl}, \ref{app:distillation}, \ref{app:network}, and \ref{app:pbt}.
Each agent is trained using 8 TPUv3s and consumes approximately 50,000 agent steps (observations) per second.

\subsection{Agent Training}
\label{sec:agent-training}
We trained five generations of agents, varying the learning setup with each generation. The results of this process is shown in \figref{fig:generations}. The learning process per generation is described below.

\begin{figure*}[t]
\centering
\includegraphics[width=\linewidth]{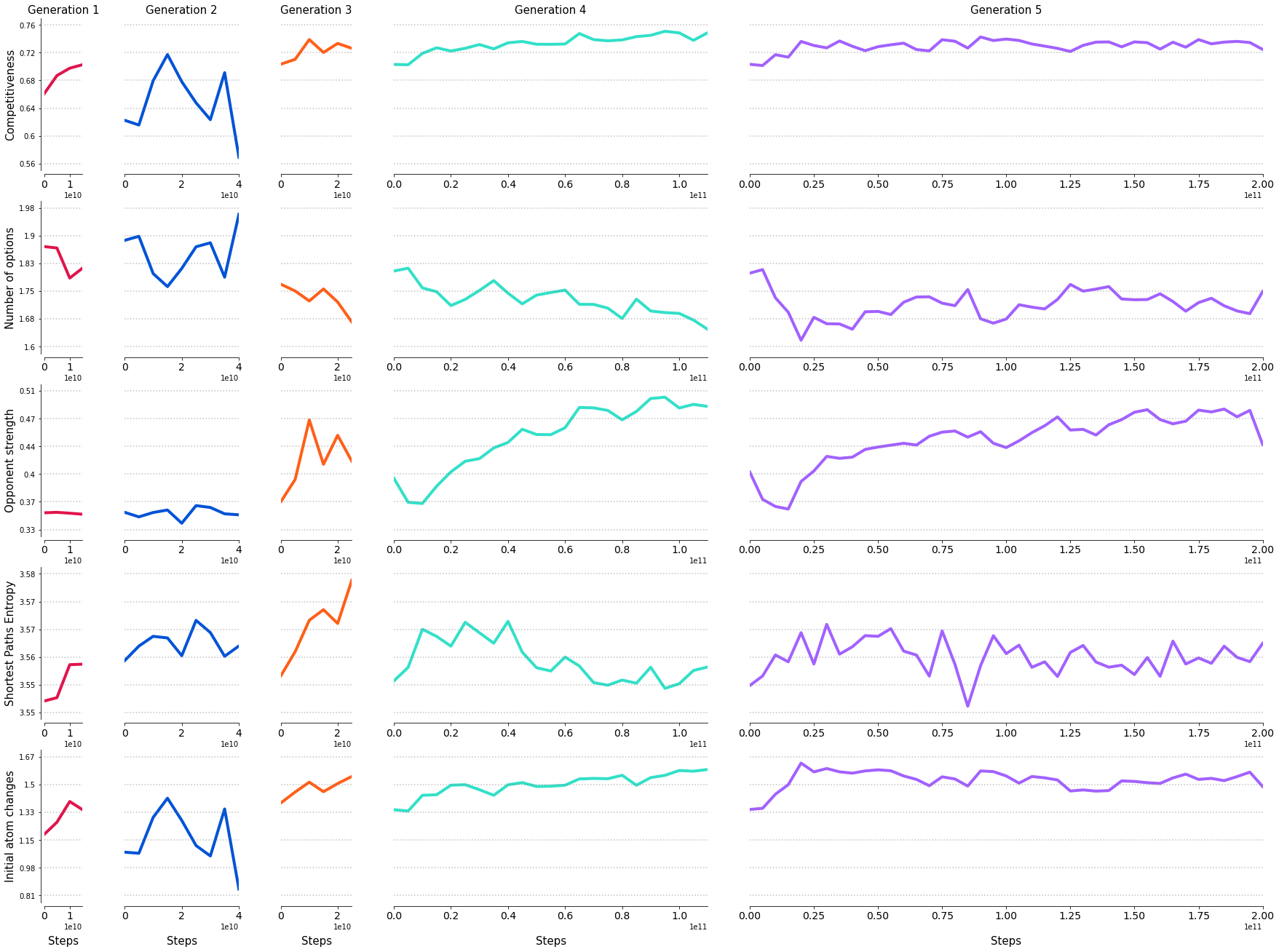}
\caption{Evolution of the training distribution of tasks due to dynamic task generation throughout 5 generations of agents (showing values from one agent per population only). We characterise the the training distribution by five measures (from the top): competitiveness (a property of games), number of options (a property of games), opponent strength (how performant the co-player in the task is), shortest paths entropy (a property of the worlds), initial atom changes (a property of the tasks, how many atomic predicates must be changed to satisfy an option). These change a lot throughout training, for example the strength of the opponents grows over time, generation 3 focuses more on worlds with larger shortest path entropy, and later generations focus on more competitive games.}
\label{fig:training_dist}
\end{figure*}
The co-player set of policies were initialised with a noop and a random policy. We used the generational mechanisms described in \secref{sec:gentraining}. At the end of each generation, we selected the best agent that was produced throughout the generation. This agent was then used in three ways by subsequent generations: 1) as a policy to use for distillation in the next generation, 2) as an additional policy in the co-player set of policies, and 3) as an additional player as part of the computation of the \evaltrain{} normalised percentile metric.

We varied the learning setup in the following way across generations. In the first two generations, the agent was trained with self reward-play to encourage exploration. In these generations, the fitness used for PBT was the average participation as measured on the \evaltrain{} task set. Subsequent generations were trained without self reward-play and used Pareto dominance over 10th, 20th and 50th percentiles of normalised score on the \evaltrain{} task set as PBT fitness. When selecting the best agent for the next generation, the agent with the highest participation was chosen in the first two generations, and the agent with the highest 10th percentile normalised score in subsequent generations.

After two generations of training, we obtained an agent trained with self reward-play with a high \evalvalid{} participation (91\%) but low \evalvalid{} 10th percentile and 50th percentile normalised scores -- 23\% and 79\% respectively. The generation 3 agents quickly outperformed these scores as they did not use self reward-play and instead maximised true reward. Our final agent in generation 5 reached 95\% participation (however it participates in 100\% of tasks that humans can, see details in \secref{sec:coverage}), 82\% 10th percentile, 112\% 50th percentile (median normalised score) on the \evalvalid{} set, and 585 average return on the hand-authored task set (which is provably at least 65\% of the optimal policy value), \figref{fig:generations}~(right).
The learning surface for the final 5th generation is shown in \figref{fig:gen5-learning}.

\label{sec:restraining}
\subsubsection{Dynamic Task Generation Evolution}
\figref{fig:training_dist} shows how various properties of our tasks change throughout training as a result of the dynamic task generation (DTG).

\begin{figure*}[t]
	\centering
     \begin{subfigure}[b]{0.32\textwidth}
         \centering
         \includegraphics[width=\textwidth]{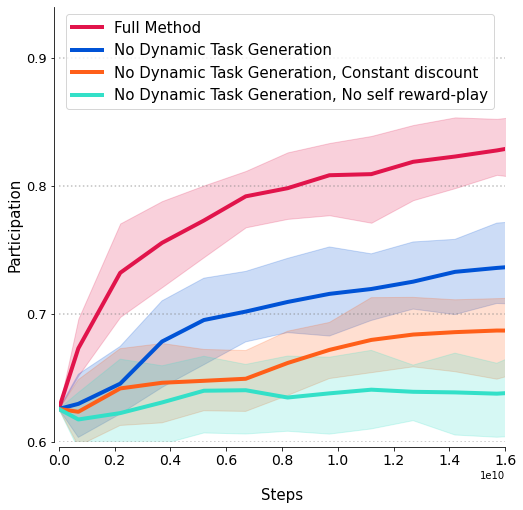}
         \caption{Participation as we ablate away dynamic task generation, the variable discount used in self reward-play, and self reward-play altogether.}
         \label{fig:gen1_ablations}
     \end{subfigure}
     \hfill
     \begin{subfigure}[b]{0.32\textwidth}
         \centering
         \includegraphics[width=\textwidth]{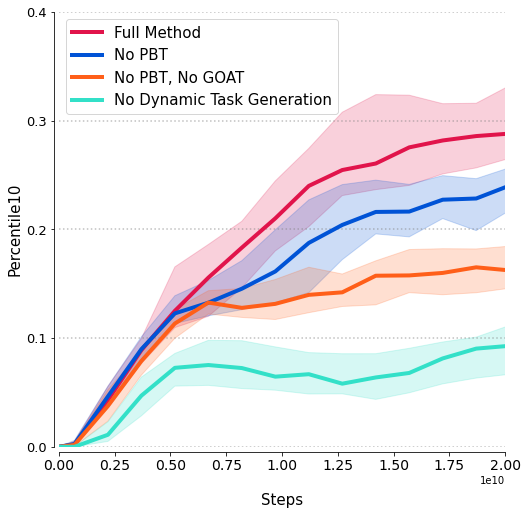}
         \caption{The 10th percentile normalised score as we ablate away PBT, the GOAT architecture, and dynamic task generation.}
         \label{fig:gen3_ablations}
     \end{subfigure}
     \hfill
     \begin{subfigure}[b]{0.32\textwidth}
         \centering
         \includegraphics[width=\textwidth]{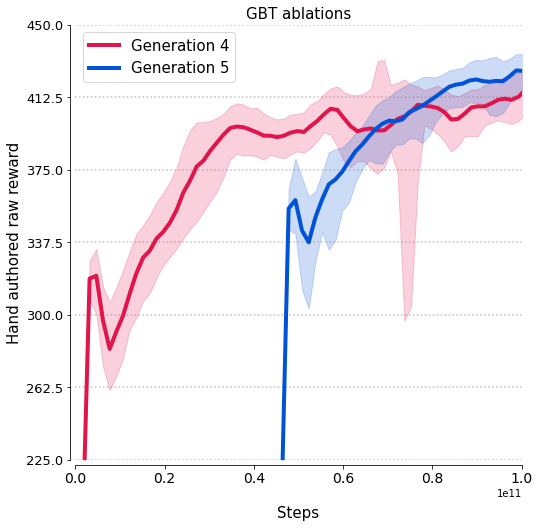}
         \caption{The 10th percentile normalised score of the fourth and fifth generation as we keep the fourth generation running.}
         \label{fig:gbt_ablations}
     \end{subfigure}
	\caption{Ablations of the training methods. In all plots, the curve designates the median agent performance in the population while the shaded area shows the spread between the best and the worst agent.}
\end{figure*}

We can see that for generation 3 and onward, DTG significantly increases the average strength of the co-players that the agent trains against. Similarly, there is an increase in the competitiveness as well as a decrease in the number of options of games presented to the agent for training. The composition of these 3 factors -- stronger opponents, more competitive scenarios, and less options, creates a training distribution of hard problems (since agents are forced to compete with capable opponents). Similarly, the number of initial atomic predicate changes needed gradually increases, meaning that agents are increasingly placed in scenarios where multiple predicate states must be changed from their initial state to obtain reward.

All these changes are driven by the agent's performance; there is no direct control given to the agent to focus on any of the above properties, and thus these dynamics are purely emergent.

\subsubsection{Ablation Studies}
\label{sec:ablations}
Our ablation studies evaluate the impact of different aspects of our training methodology.

\paragraph{Early generations: self reward-play and dynamic task generation.} As discussed in \secref{sec:gentraining}, early training in our environment is difficult. We use self reward-play to encourage the agent to explore changing the environment state, and dynamic task generation to avoid training on tasks that are initially too hard for the agent and would not provide any useful training signal. In this ablation, we trained multiple agents from scratch with a diverse pool of co-player policies. We show the participation of the different trained agents in \figref{fig:gen1_ablations}. Our full method, which used both dynamic task generation and self reward-play, reached a participation of 84\% after 16 billion steps. We see that removing in turn dynamic task generation, the use of zero discounts on step changes (part of our self reward-play procedure), and self reward-play resulted in significant reductions in performance. When none of these methods are used, the agent fails to learn any meaningful policy.

\paragraph{Later generations: Population based training, the GOAT architecture and dynamic task generation.} In our next ablation, we consider a setup similar to the third generation in our main experiments. The agents were not trained with self reward-play, but during the first 4 billion steps have a distillation loss towards the teacher policy of an agent that was trained with self reward-play. The agents were trained with a diverse pool of co-player policies. The results are shown in \figref{fig:gen3_ablations}. We trained each agent for 20 billion steps. Similarly to our main experiments, our full method uses PBT, the GOAT architecture and dynamic task generation. Our first ablation removes PBT from our method, replacing it by a simple sweep across 8 agents, which leads to a $\sim20\%$ reduction in performance of the best agent. Additionally removing the GOAT architecture from our method and replacing it with a simpler architecture similar to the one used in \cite{hessel2019multi} yields another $\sim30\%$ reduction in performance. Finally, removing dynamic task generation from our method whilst keeping other aspects constant leads to a $\sim65\%$ reduction in performance.

\paragraph{Generation based training.} In our final ablation, we consider the benefits of generation based training. We kept the fourth generation of main experiments from \secref{sec:restraining} running in order to compare its performance to the fifth generation. The results are shown in \figref{fig:gbt_ablations}. We offset the fifth generation's curve to the point the best agent from the fourth generation was selected. 
We can see that as training progresses the fifth generation outperforms the previous generation (both in terms of comparing best agents from corresponding populations, as well as comparing the averages), even when generation 4 was trained for the same amount of time.

\subsection{Performance Analysis}
\label{sec:resperformance}
Due to the vastness of task space, with unknown maximum scores, there is no single notion of performance to report. Consequently, we rely on relative performance analysis and other qualitative notions of progress described below.

\begin{figure*}[t]
\centering
\includegraphics[width=0.8\textwidth]{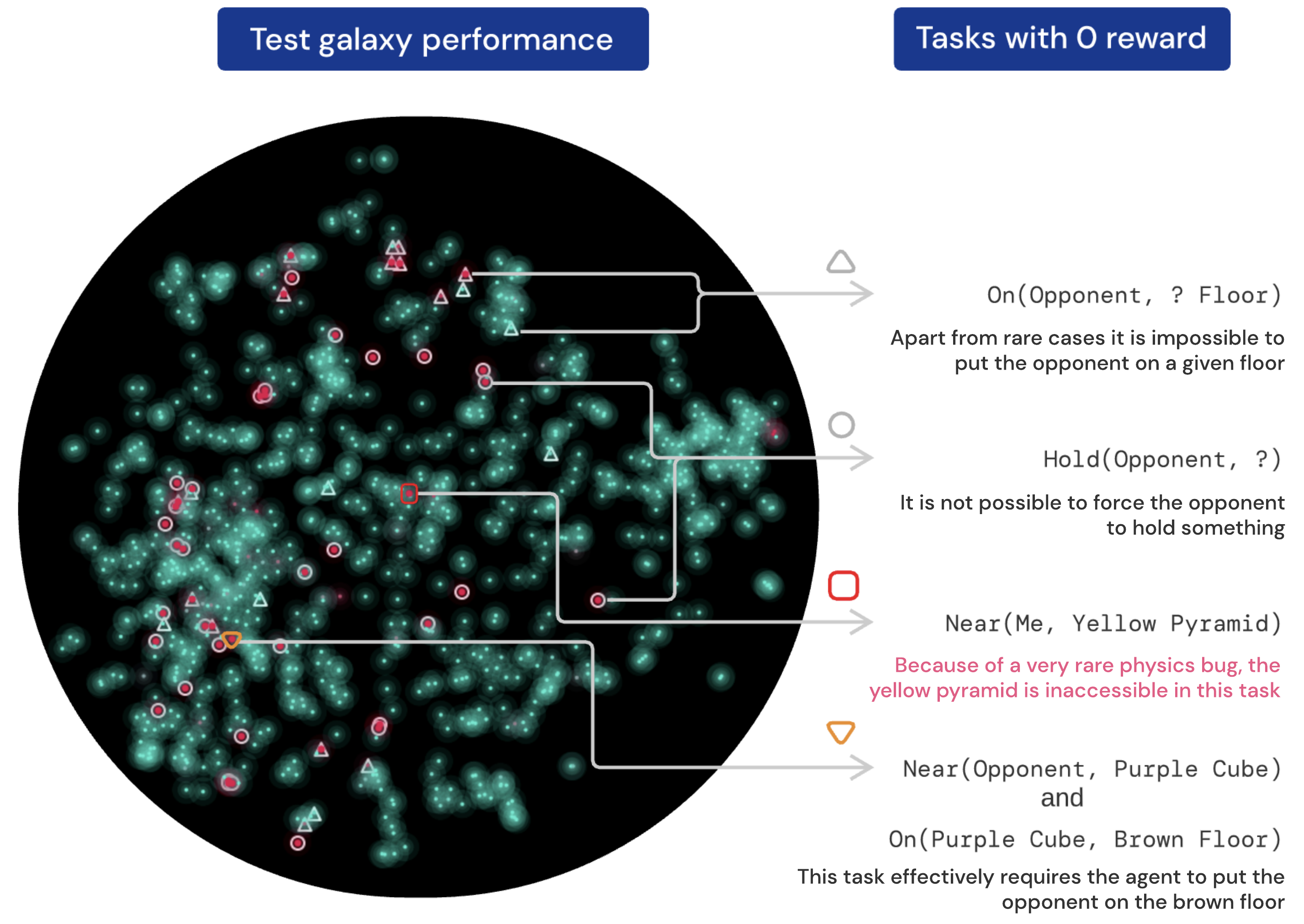}
\caption{A visualisation of the \evalvalid{} set of tasks, with the corresponding agent performance. The red colour corresponds to a low normalised score and green to a high one. We identify four sources of games the agent scores 0 reward on (listed on the right): 1) tasks that require the agent to put the opponent on a specific floor (marked as triangles in the galaxy); 2) tasks that require the agent to make the co-player hold an object (marked as circles in the galaxy); 3) a single task (in red in the galaxy) which is impossible due to a very rare physics simulation bug; 4) a single task (in orange in the galaxy) that requires the agent to put the co-player on a given floor by a composition of two predicates. After removing these four types of tasks, which cannot be solved even by a human, our agents participate in \emph{every \evalvalid{} task}.}
\label{fig:widelycapable}
\end{figure*}
\subsubsection{Coverage}
\label{sec:coverage}
First, we focus our attention on answering the question \emph{are there any \evalvalid{} tasks, where the agent never reaches a rewarding state?} We identify that there are indeed a few percent of this space where none of the agents ever score any points. Further investigation shows that all these failed tasks involve impossible challenges, requiring an agent to \emph{make the co-player hold something} (which, without the cooperation of the opponent is impossible) or to \emph{place the co-player on a specific floor} (which, can also be impossible to achieve given the physical simulation of the game). Furthermore, we identify a single task, which, due to a very rare physics bug is impossible to solve because the object of interest spawns outside the reachable region. \figref{fig:widelycapable} shows these games in the XLand galaxy. After removing the impossible tasks listed above our agent participates in every single \evalvalid{} task, thus suggesting they are indeed widely capable. 

\subsubsection{Relative Performance}
We investigate various properties of the games, and how they translate to the relative improvement of our agents (using \evalvalid{} normalised scores to measure this quantity).
\begin{figure}
    \centering
    \includegraphics[width=0.95\linewidth]{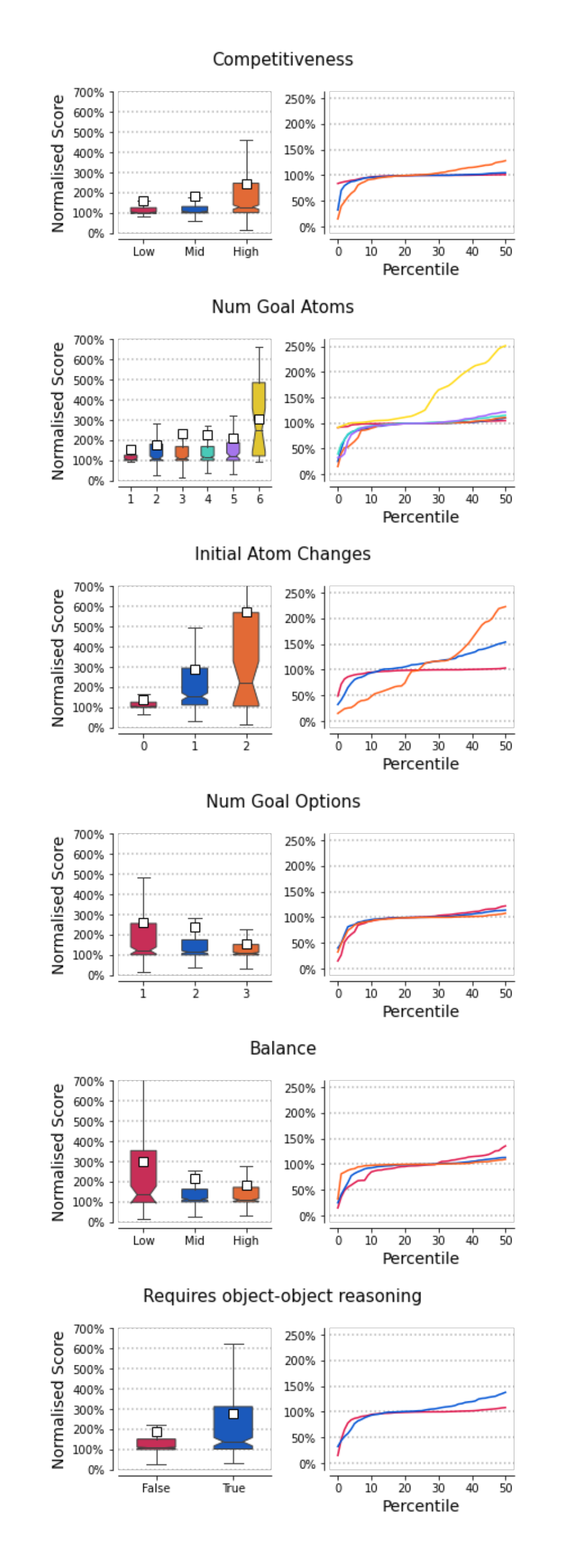}
    \caption{\textbf{(Left)} Box plots showing the distribution of normalised scores for the final agent across various types of validation tasks. Whiskers denote the minimum and maximum value, the notch denotes the median, and the box area is between the 25th and 75th percentiles. The white square denotes average performance. \textbf{(Right)} Normalised \evalvalid{} percentiles for the corresponding subsets of tasks.}
    \label{fig:performance}
\end{figure}
In \figref{fig:performance} we can see that the overall normalised score of our agent is higher on games which are more competitive, suggesting that it is in these challenging competitive scenarios our proposed learning process brings the biggest improvement relative to the pretrained evaluation policies in the \evalvalid{} set. Similarly, high normalised score is correlated with a large number of goal predicates (and thus a need to reason about many relations at the same time) as well as high initial atom changes (the number of relations that need to be changed, before an agent can get to a rewarding state). 
We also observe the biggest improvements with fewer options -- games where there is just one option are much harder on a purely navigational level, as an agent cannot \emph{choose} what to do, but rather is forced to satisfy a single option. Finally, we also see a big improvement relative to the evaluation policies when the agent is tasked with goals involving object-object interactions, such as \emph{make the yellow sphere be near the purple pyramid}, as opposed to tasks related to the players themselves, e.g. \emph{hold a purple sphere}. Overall, we see a general trend of agents showing the greatest improvements in the most challenging parts of our game space.

\subsection{General Capabilities}
\label{sec:res-capabilities}
We now provide an overview of some of the general capabilities of the agent observed, allowing them to participate in a variety of tasks, execute various behaviours, and show satisfactory handling of new, unexpected situations.

Whilst the current instantiation of XLand is extremely vast, one can easily hand-author tasks that could only extremely rarely, or cannot at all, be generated during training due to the constraints of our training task generation process. For example we can place agents in worlds that lack ramps to challenge their ability to navigate, we can make them face unseen co-players, and we can execute interventions mid-episode. These probe tasks allow us to better understand and clarify the limits of generality of our agents.

\subsubsection{Hand-authored tasks}
We now consider the qualitative behavioural properties of our agents at different points throughout training on hand-authored tasks (see \figref{fig:handauthored-tasks} for some some examples and Table~\ref{tab:handauthoredlist}~\&~\ref{tab:handauthoredlist2} for a list of all). We compare two agents on a selection of the hand-authored task set: the final generation 4 ($\policy_{\mathrm{G}_4}$) agent and the final generation 5 ($\policy_{\mathrm{G}_5}$) agent. A selection of videos of the generation 5 ($\policy_{\mathrm{G}_5}$) agent can be found in the \href{https://youtu.be/lTmL7jwFfdw}{supplementary results video here}.

\paragraph{Capture the flag.} In this two-player task, the agents' goal is to capture the cube in the opponent's base and bring it back to their own base. An agent gets a reward if the opponent's cube touches the floor of their own base while their own cube also touches the floor of their own base, with the opponent having an equivalent goal with respect to its base floor. Both agents are able to navigate to their opponent's base to capture their cube. However, $\policy_{\mathrm{G}_4}$ often finds it difficult to find the way back to its own base. Furthermore, it often gets tagged by the opponent, making it respawn at its initial spawn location. $\policy_{\mathrm{G}_5}$ on the other hand shows better navigational skills and usually finds its way back to its base after capturing the cube.
\paragraph{Hide and seek: hider.} $\policy_{\mathrm{G}_4}$ moves somewhat randomly with abrupt changes in direction. This can make it hard for the opponent to keep seeing it. $\policy_{\mathrm{G}_5}$ on the other hand moves very specifically away from the co-player and often up the ramp and onto the side of the platform opposite the co-player. This forces the co-player to go around the ramp.
\paragraph{Hide and seek: seeker.} $\policy_{\mathrm{G}_4}$ searches for the co-player throughout the world and then stands still once the co-player is in its vision. It does not anticipate the co-player's movement as it is about to come out of its vision. $\policy_{\mathrm{G}_5}$ prefers to continuously follow the co-player in order to be right next to it. In this way, it rarely lets the co-player out of its vision.
\paragraph{King of the hill.} In this two-player task, the agent gets a reward if it is the only player at the top of the hill (touching the white floor). Once they get to the top of the hill, both $\policy_{\mathrm{G}_4}$ and $\policy_{\mathrm{G}_5}$ stay there and are able to push away the co-player whenever it comes near. However, $\policy_{\mathrm{G}_4}$ sometimes fails to navigate to the top of the hill, getting stuck in a loop. $\policy_{\mathrm{G}_5}$ is more consistent in its navigational abilities to get to the top of the hill.
\begin{figure*}[t]
    \centering
    \includegraphics[trim=0 550 0 0,clip,width=0.9\linewidth]{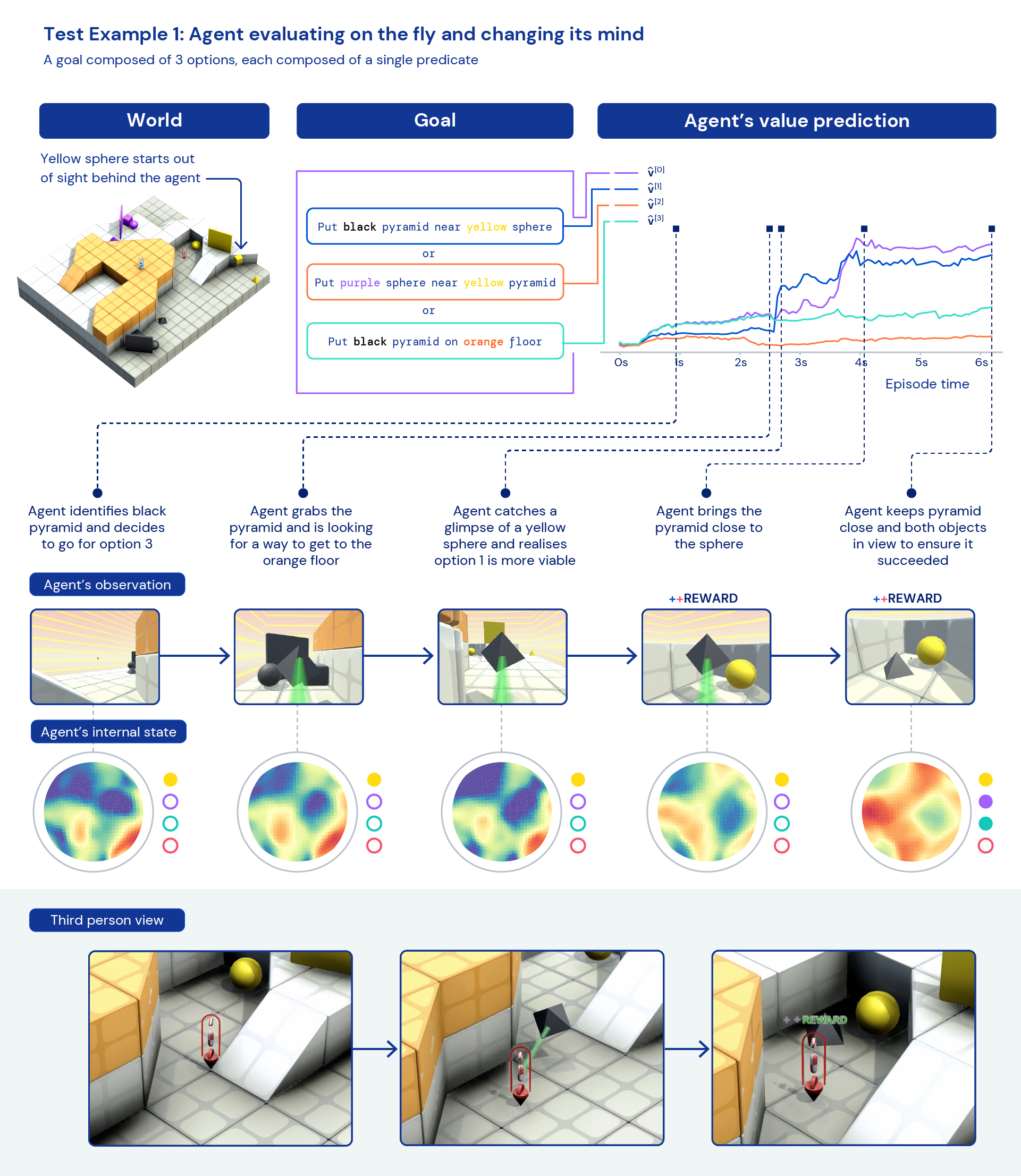}
    \caption{\textbf{(Top)} From the left: rendering of the world; a goal composed of 3 options, each represented as a single predicate; Plots of the internal value function predictions of the GOAT module, with the colours corresponding to specific options. \textbf{(Middle)} Call-outs of 5 situations, from the perspective of the agent. \textbf{(Bottom)} A Kohonen Network representing the activity of the GOAT module (\secref{sec:representation}). The four coloured circles represent the Kohonen Neurons activity (from top): whether the agent is early in the episode (yellow), if it is optimistic about future rewards (purple), if it thinks it is in a rewarding state (cyan), if it thinks multiple atoms are missing (orange). See \figref{fig:neurons} for more details.}
    \label{fig:multipleoptions} 
\end{figure*} 
\paragraph{XRPS Counter Yellow Sphere.} In XRPS (\secref{sec:game-diversity}), the agent can get points for holding any sphere, as long as its colour is not countered by the colour of the sphere the opponent is holding. However, the opponent player is goal-conditioned to hold the yellow sphere only. $\policy_{\mathrm{G}_4}$ tends to hold a sphere at random from the ones available. When this happens to be the black sphere, it gets no reward due to the co-player countering it with the yellow sphere. $\policy_{\mathrm{G}_5}$ on the other hand notices the co-player holding the yellow sphere and counters it by stealing the yellow sphere and holding it itself. It succeeds at holding it while the co-player tries to get it back. However, neither agent explicitly seeks to hold the purple sphere which would counter the opponent holding the yellow sphere.
\paragraph{Stop rolling.} In this task, the agents have to keep a sphere from rolling to the bottom of a slope. The agents only get a reward if the sphere is not touching the bottom floor and is not being held. $\policy_{\mathrm{G}_4}$ simply lifts the sphere up in the air and lets it drop, gaining rewards for the brief moments when the sphere is dropping. $\policy_{\mathrm{G}_5}$ throws the sphere up the slope and then tries to block it from rolling down with its body. Often, $\policy_{\mathrm{G}_5}$ manages to corner the sphere between its body and the wall as the sphere is on the slope and scores rewards for the remainder of the episode without moving.

\subsubsection{Behavioural case studies}
Let us now focus on 3 specific case studies showing interesting emergent behaviours.

\begin{figure*}[t]
    \centering
    \includegraphics[trim=0 550 0 0,clip,width=0.9\linewidth]{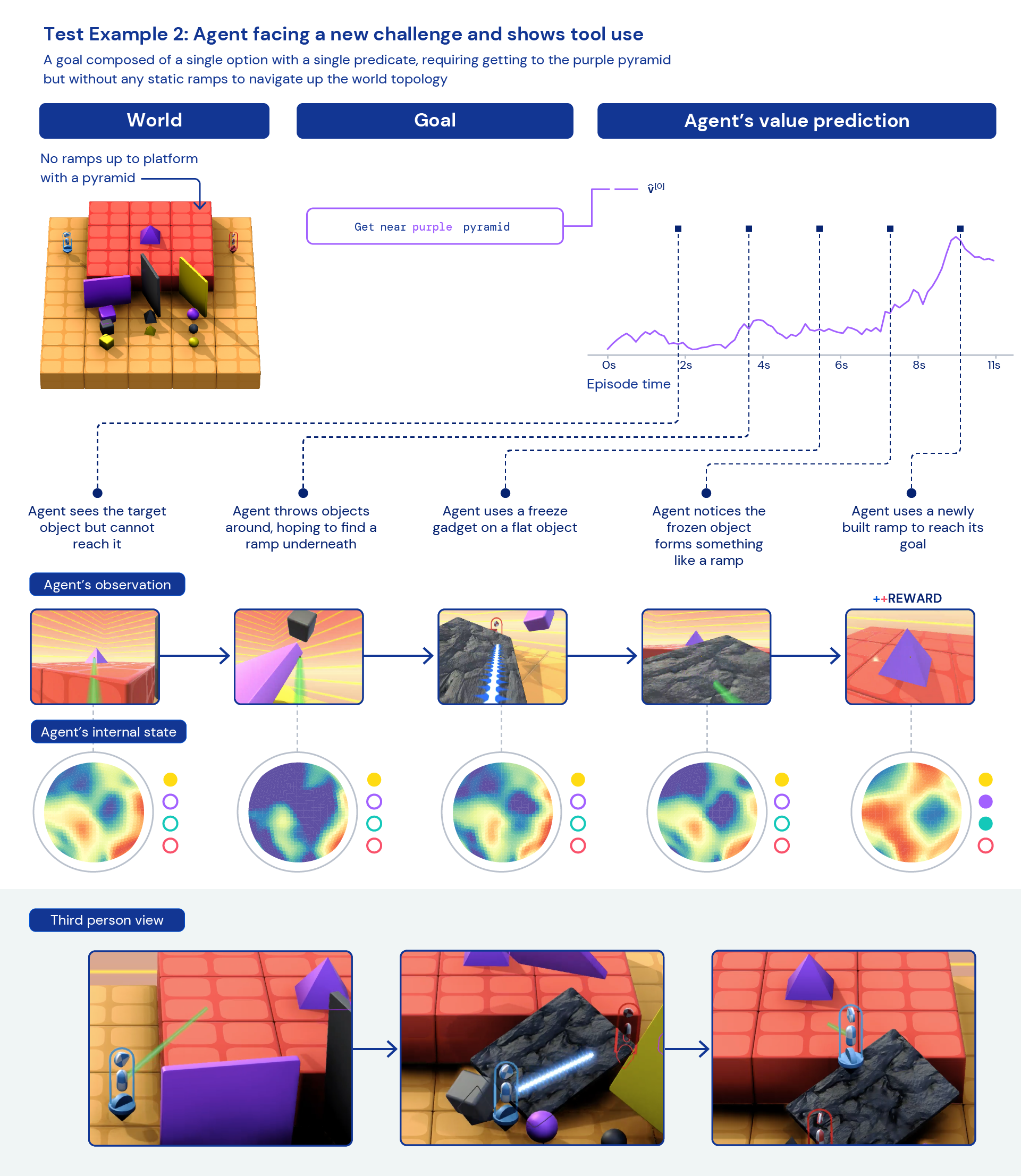}
    \caption{\textbf{(Top)} From the left: rendering of the world; a goal composed of one option; Plots of the internal value function prediction of the agent. \textbf{(Middle)} Call-outs of 5 situations, from the perspective of the agent. \textbf{(Bottom)} A Kohonen Network representing the activity of the GOAT module (\secref{sec:representation}). The four coloured circles represent the Kohonen Neurons activity (from top): whether the agent is early in the episode (yellow), if it is optimistic about future rewards (purple), if it thinks it is in a rewarding state (cyan), if it thinks multiple atoms are missing (orange). See \figref{fig:neurons} for more details.}
    \label{fig:tooluse} 
\end{figure*}    

\paragraph{On-the-fly option evaluation}
In \figref{fig:multipleoptions} we see an agent trying to solve a task with a goal consisting of 3 possible options. Initially, the agent does not see a yellow sphere, but it does see a black pyramid and the orange floor. Its third option rewards the agent for placing the black pyramid on the orange floor, and looking at the agent's internal option-values prediction, we see that indeed the value of the whole goal $\goatvaluehead_t^{[0]}$ (violet curve) is closest to the third option value $\goatvaluehead_t^{[3]}$ (green curve). Around 2.5s into the episode, the agent sees a yellow sphere, which leads to a dramatic increase in its internal prediction of what would happen if it was to satisfy option 1 instead ($\goatvaluehead_t^{[1]}$, blue curve), which rewards the agent for placing the black pyramid near the yellow sphere. As a result, the internal value function of the whole game switches to upper bound the first option, and rather than navigating to the orange floor, the agent brings the black pyramid next to the sphere. This case study exemplifies the internal reasoning of the GOAT module, hinting at intentional decisions about which options to satisfy based on the current state of the environment.

\paragraph{Tool use}
In \figref{fig:tooluse} we see an agent placed in a world, where it needs to get near to a purple pyramid placed on a higher floor. However, in this world there is no ramp leading to the upper floor -- this initial lack of accessibility is impossible to occur during training due to the procedural world generation process constraints. We observe the agent initially trying to move around the red block, looking for a ramp. It starts to throw various objects around, which can either be interpreted as looking for a ramp hidden underneath, or simply an emergent heuristic behaviour of trying to increase the entropy of the environment in a situation when the agent does not know what to do. Around 5 seconds into the episode a slab thrown by an agent lands in the position partially supported by the upper floor, and the agent uses a freezing gadget to keep it in place. A moment later the agent can see a target purple pyramid in front of it with a frozen object looking like a ramp leading to the purple pyramid's floor, and its internal value estimate rapidly increases, suggesting that the agent understands that it has found a solution to the task. The agent navigates onto the frozen object and reaches its goal. We can see that the internal representation activity (described in \secref{sec:representation}) at 10 seconds is very similar to the final internal activity from the previous case study -- we recognise this visual pattern as emerging when an agent is in a \emph{content} state.
\begin{figure*}[t]
    \centering
    \includegraphics[trim=0 550 0 0,clip,width=0.9\linewidth]{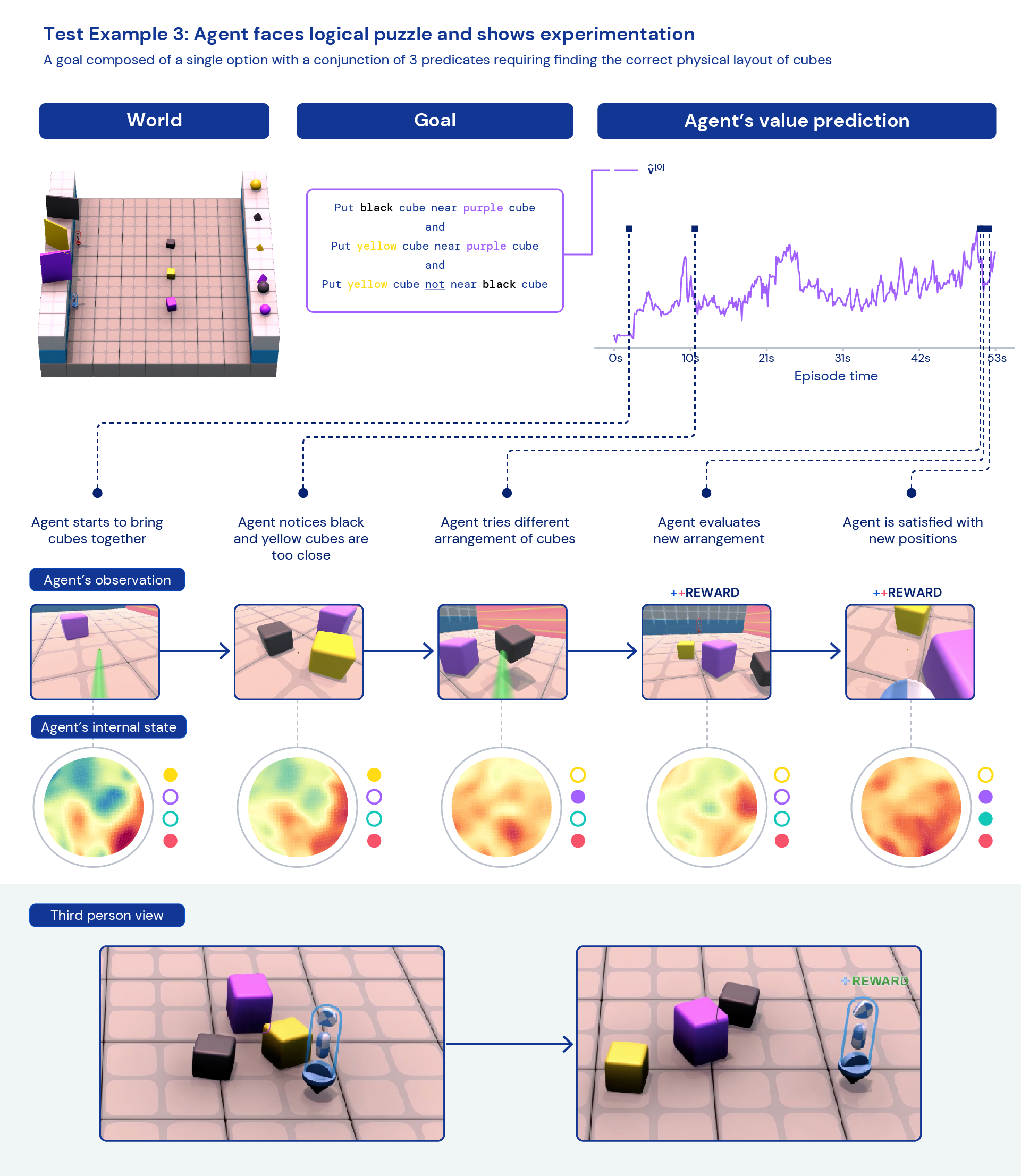}
    \caption{\textbf{(Top)} From the left: rendering of the world; a goal composed of one options; Plots of the internal value function prediction of the agent. \textbf{(Middle)} Call-outs of 5 situations, from the perspective of the agent. \textbf{(Bottom)} A Kohonen Network representing the activity of the GOAT module (\secref{sec:representation}). The four coloured circles represent the Kohonen Neurons activity (from top): whether the agent is early in the episode (yellow), if it is optimistic about future rewards (purple), if it thinks it is in a rewarding state (cyan), if it thinks multiple atoms are missing (orange). See \figref{fig:neurons} for more details.}
    \label{fig:experimenting} 
\end{figure*}    

\paragraph{Experimentation}
\figref{fig:experimenting} is a final case study, where an agent is placed in a big open room, with most of the objects removed from the reachable space, and only 3 cubes left. The task the agent is facing is to put the black cube near the purple cube, the yellow cube near the purple cube, without putting the black and yellow cubes near each other. This simple logical puzzle requires an agent to figure out that there is a spatial ordering that satisfies these principles: a line with the yellow cube, followed by the purple cube, followed by the black cube. Note, that whilst this does not look like a complex problem on a predicate level, it is a very hard exploration problem due to the physical instantiation of XLand tasks -- namely
$$
\tfrac{\#\{\state: \reward_\goal(\state) = 1\}}{\#\{\state: \state \in \statespace\}}
\ll 
\tfrac{\#\{\predicate(\state): \reward_\goal(\state) = 1\}}{N_\predicate}.
$$
From the agent's behaviour and internal value we can hypothesise that the agent is initially confused. It starts by bringing the cubes together. Then at around 10 seconds we can see it visually inspecting the scene with the yellow and black cubes too close, after which it tries to reshuffle them. This reshuffling process is repeated multiple times for the next few dozen seconds, until eventually around 50 seconds into the episode, the agent stumbles upon a spatial arrangement of the cubes that satisfies the goal, which the agent again inspects visually. Whilst still clearly not content when it comes looking at the agent's internal state/value prediction, the agent keeps the objects in the rewarding state and stops shuffling the cubes. This within-episode experimentation behaviour could be a general heuristic \emph{fallback} behaviour -- when it lacks the ability to 0-shot generalise through understanding, it plays with the objects, experiments, and visually verifies if it solved the task -- all of this as an emergent behaviour, a potential consequence of an open-ended learning process. Note, that agent does not perceive the reward, it has to infer it purely based on the observations.

\subsubsection{Multi-agent}
We now investigate some emergent multiplayer dynamics between agents playing in specific probe games. We take 13 agent checkpoints through training of the final (5th) generation of our agent (checkpoint 1 is the earliest in training through to checkpoint 13 which is the latest in training). For each of the probe scenarios described below, we play every single pair of checkpointed policies against each other. This way we obtain $13^2 = 169$ matchups, and evaluate each pair of players on 1000 different worlds (to marginalise over physical instantiation), allowing us to study the development of strategies, social behaviours and learning dynamics of the agents in these games. Note, that the agent was never trained against these checkpoints, the only co-players it ever experienced during training were from the previous generations. More details can be found in~\secref{app:ma}.

\begin{figure}[t]
    \centering
    \includegraphics[width=\linewidth]{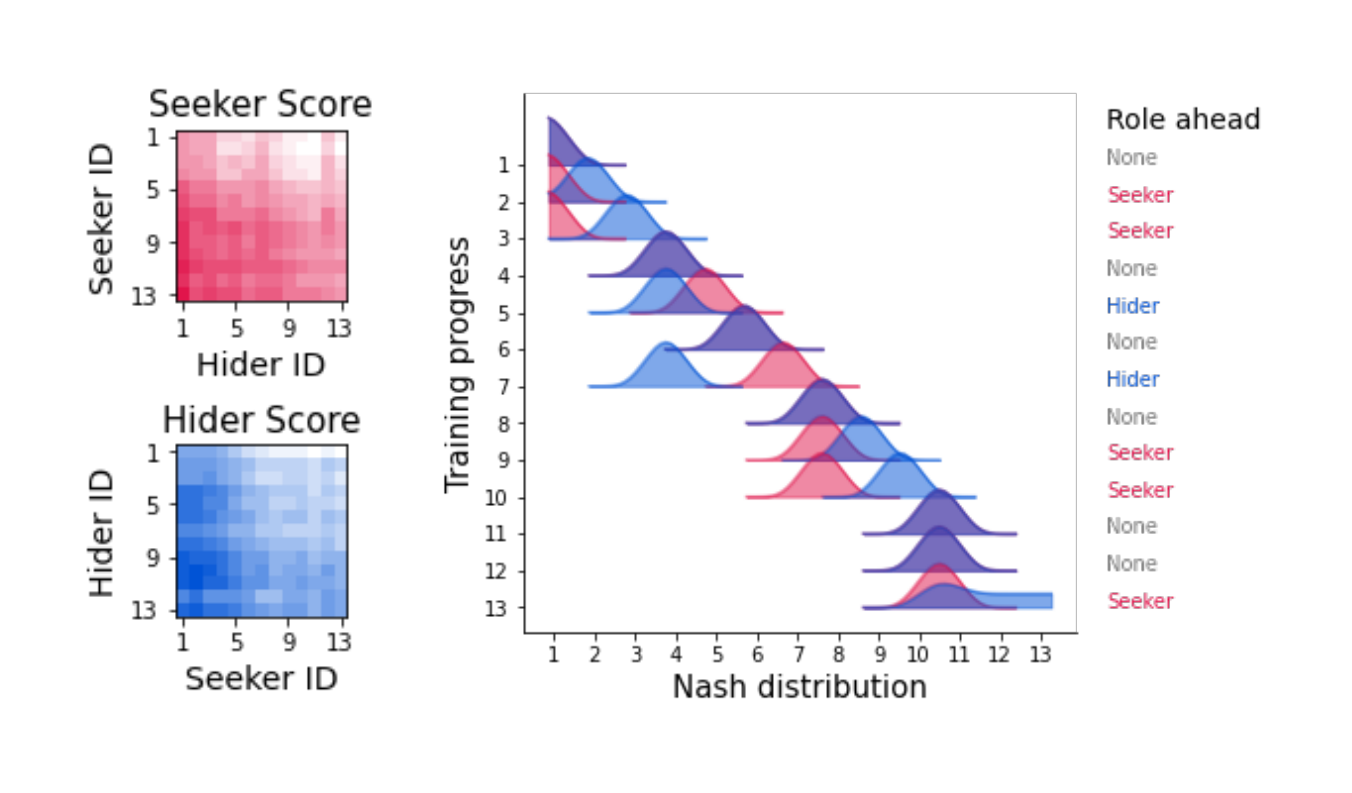}
    \caption{\textbf{(Left)} 
    Payoffs for the game of Hide and Seek played by checkpoints of the agent through training from start (1) to end (13), marginalised across 1000 different worlds, from the perspective of the seeker (top) and hider (bottom) player roles. (White is low, colour is high).
    \textbf{(Right)} The evolution of the Nash equilibrium distribution through training for each role of the player. One can note the back-and-forth dynamic of the hider and seeker improving over time. The agent never trained with these checkpoints.}
    \label{fig:hide-and-seek} 
\end{figure}   
\paragraph{Hide and Seek} We revisit the game of hide and seek in this new evaluation context. 
It is an asymmetric, imbalanced, fully competitive game.
\figref{fig:hide-and-seek} shows the results of this experiment. With more training (later checkpoints), the results show the agents keep improving in the execution of both hider and seeker roles, showing a somewhat transitive strength progression, without exhibiting forgetting that can occur in multi-agent training~\citep{vinyals2019grandmaster,czarnecki2020}, and the Nash equilibrium is mostly centered around the newest agents. We can also see that there is a noticeable back-and-forth dynamic between the seeker and hider strategy, as initially the Nash equilibrium for the seeker stays at checkpoint 1 (early in training), whilst the hider Nash equilibrium keeps moving to the newest checkpoint. This suggests that the agent is gradually improving its hiding strategy. Later, we see the opposite -- the hider Nash equilibrium stops moving, while the seeker Nash equilibrium keeps improving, with this switch happening multiple times. Note that this is being observed without the agents ever training against each other, thus we hypothesise that these developments have to be coming from agents acquiring new behaviours and strategies in other parts of XLand task space. In particular, it is worth noting that during training, whilst the agent does not play against its checkpoints, and even though the game of hide and seek itself is not a part of the training space, agents are facing games where their goal is to ``see the other player'', but the other player's goal will be something else. Consequently, even though they only train against a few instances of other agents, the space of behaviours they can encounter is enormous because these agents are themselves conditioned on a variety of goals.

\begin{figure}[t]
    \centering
    \includegraphics[width=0.8\linewidth]{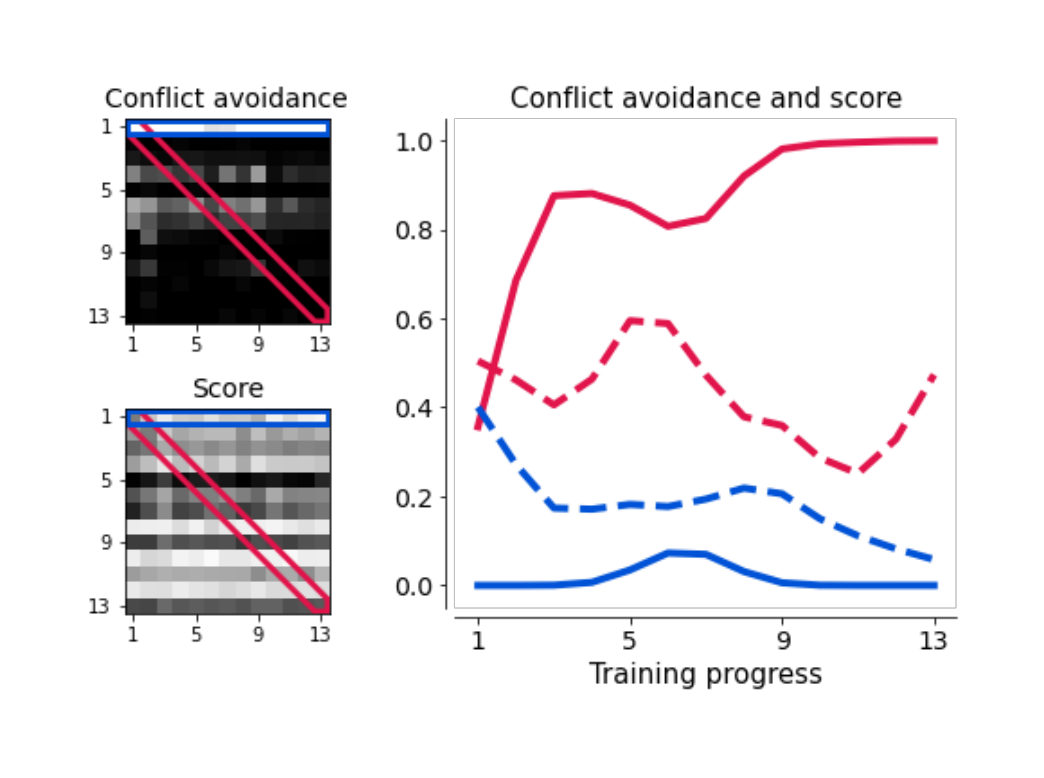}
    \caption{\textbf{(Left)} 
    Probabilities of each agent choosing to avoid conflict and the corresponding normalised scores, as a function of their total training time (1 meaning least trained, and 13 meaning the most trained). (White is low probability, black is high probability).
    \textbf{(Right)} Fraction of conflict avoiding behaviours (solid curve) and the corresponding normalised score (dashed curve). The agent becomes more conflict avoiding over time whilst preserving its performance when matched with a copy of itself (red curve), while the earlier agent playing against later agents is not avoiding conflict and its performance also keeps decreasing (blue curve).}
    \label{fig:non_conflict} 
\end{figure}

\paragraph{Conflict Avoidance} We hypothesise that as training progresses agents might develop the behaviour of avoiding conflict with other agents in the situations where there is an alternative non-conflicting option to be satisfied. We create a simple game, where an agent can choose to place one of two spheres on a specific floor, while the other agent wants to put one of these spheres on a different floor. With both spheres being equidistant from a target floor, the only reason to pick the non-conflicting sphere is in order to avoid conflict with the other agent. In \figref{fig:non_conflict} we can see that as the agent trains, it exhibits more and more conflict-avoiding behaviour, even though on average this does not necessarily lead to an increase in return on this particular task. However, empirically when early not-conflict-avoiding checkpoints play with increasingly trained checkpoints, they achieve a decreasing amount of reward. Note, that the agents are not training against each other, meaning that this development in behavioural response is purely an effect of the dynamic training distribution encountered during the open-ended learning process.

\begin{figure}[t]
    \centering
    \includegraphics[width=\linewidth]{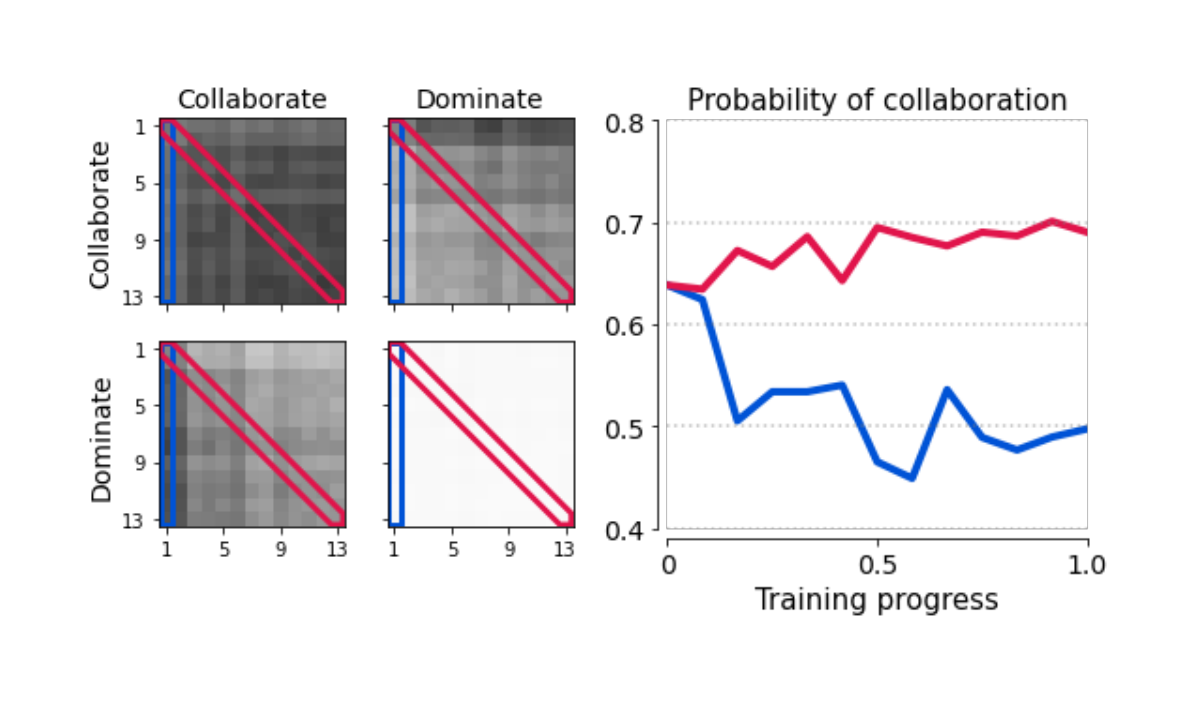}
    \caption{\textbf{(Left)} 
    Probabilities of each agent choosing to collaborate or dominate, as a function of their total training time (1 meaning least trained, and 13 meaning the most trained). (White is low probability, black is high probability).
    \textbf{(Right)} Fraction of collaborative behaviours in a Chicken-like game through agent training. The agent becomes more collaborative over time when matched with a copy of itself (red curve), and dominates more with earlier versions of itself (blue curve).}
    \label{fig:chicken_colab} 
\end{figure}

\paragraph{Chicken Game} 
In this experiment, we create an XLand version of a game-theoretic social dilemma called \emph{Chicken}. In this setup, each agent can choose to either cooperate with its co-player or to try to dominate it. We observe two interesting trends with respect to the tendency to seek cooperative solutions in \figref{fig:chicken_colab}. First, if an agent is playing with a checkpoint from very early in training, it tends to dominate it more. On the other hand, when playing with a copy of itself (self-play) its tendency to collaborate increases over training time. One simple explanation of this phenomenon is that for cooperation to work, both sides need to be capable of doing so. Consequently, it is perhaps harder to cooperate with a less capable agent. However, once facing someone of exactly same strength (self-play) the collaborative solution becomes preferred.

\subsubsection{Goal Interventions}
During training our agents always received a single goal throughout an episode, the same goal at every timestep. We study whether the agent is able to adapt on-the-fly if this property is broken, and the goal changes mid-way through a single episode.

\begin{figure}[t]
    \centering
    \includegraphics[width=0.6\linewidth]{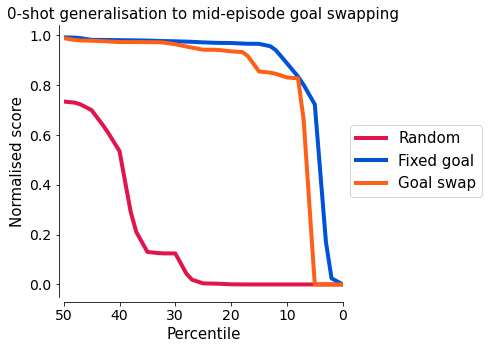}
    \caption{Performance of the agent in 0-shot generalisation experiments where the goal of the agent is changed in the middle of an episode. Note that agents never perceived dynamically set goals during training.}
    \label{fig:goal_swap}
\end{figure}

We sample $500$ tasks all consisting of single option, one predicate games from the \evalvalid{} set. We run the agent for an extended episode of 3/2 length of a regular episode, where in the first 1/3 the agent is given one goal (where we ignore its performance), and then we change the goal supplied to the agent to a different one. To simplify the setting, the co-players use the noop-policy, simulating a single-player game.

In \figref{fig:goal_swap} we compare the normalised score of the agent evaluated in this scenario with the agent playing the same game but whose internal state is reset when the goal changes to simulate starting the episode from scratch with a fixed goal. 
We also show the performance of the agent taking random actions for reference. We notice that the performance of the agent with the changed goal is almost exactly the same as with a fixed goal, showing robustness to goal changes.

\subsubsection{Failed Hand-authored Tasks}
Whilst there are many tasks the agent participates in, there are also some hand-authored tasks the agent does not, never achieving a single reward. Some examples are: 

\paragraph{Gap tasks} Similar to the task in \figref{fig:tooluse}, in this task there is an unreachable object which the agent is tasked with being near. The object is unreachable due to the existence of a chasm between the agent and object, with no escape route (once agent falls in the chasm, it is stuck). This task requires the agent to build a ramp to navigate over to reach the object. It is worth noting that during training no such inescapable regions exist. Our agents fall into the chasm, and as a result get trapped. It suggests that agents assume that they cannot get trapped.

\paragraph{Multiple ramp-building tasks} Whilst some tasks do show successful ramp building (\figref{fig:tooluse}), some hand-authored tasks require multiple ramps to be built to navigate up multiple floors which are inaccessible. In these tasks the agent fails.

\paragraph{Following task} One hand-authored task is designed such that the co-player's goal is to be near the agent, whilst the agent's goal is to place the opponent on a specific floor. This is very similar to the \evalvalid{} tasks that are impossible even for a human, however in this task the co-player policy acts in a way which follows the agent's player. The agent fails to lead the co-player to the target floor, lacking the theory-of-mind to manipulate the co-player's movements. Since an agent does not perceive the goal of the co-player, the only way to succeed in this task would be to experiment with the co-player's behaviour, which our agent does not do.

\begin{figure*}
    \centering
    \includegraphics[width=\linewidth]{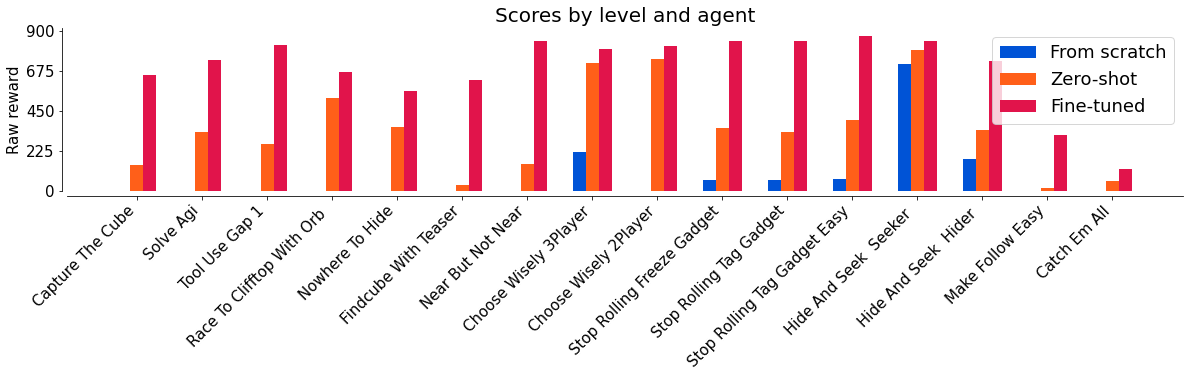}
    \caption{Comparison of three agents from different training regimes on a range of hand-authored levels. Scratch: An agent trained from scratch for 100 million steps. Zero-shot: the agent trained using our methodology and evaluated on these held out levels zero-shot. Fine-tuned: the same agent but trained for an additional 100 million steps on the level. 100 million steps is equivalent to 30 minutes of wall-clock time in our setup. This rapid finetuning improves the agent score significantly compared to zero-shot, and in the majority of cases training from scratch does not achieve any reward.}
    \label{fig:finetuning} 
\end{figure*}

\subsection{Finetuning for Transfer}
\label{sec:finetuning}
Throughout this section we have so far demonstrated zero-shot generalisation to new tasks. The breadth of coverage of the agent's behaviour suggests that whilst zero-shot performance can be achieved on many out-of-distribution test tasks, there is the potential for very quick adaptation with finetuning.

Using a simple training setup -- without PBT, dynamic task generation, or any other hyperparameter tuning -- we finetune the weights of the generally capable agent previously analysed for 100 million steps (approximately 30 minutes of training) on a number of tasks from the hand-authored set. The results are shown in \figref{fig:finetuning}. 

The results show in all cases an increase in reward achieved by the finetuned agent compared to the zero-shot performance, with the finetuned agent showing a drastic improvement of 340\% on average. By construction, the maximum reward that could ever be achieved on an XLand task of 900 timesteps is $\mathbf{V}^*(\xlandtask) \leq 900$. Using 900 as an upper bound of optimal reward per task (which is a very loose one, since even an optimal policy needs some time to reach objects of interest etc.), learning from scratch scores at least 9\% of the performance of the optimal policy, zero-shot performance is at the level of 39\% and the finetuned agent achieves 77\%. With the same computational budget and 30 minutes of training, learning from scratch on these tasks fails in the majority of tasks.

The task \emph{Make Follow Easy} is described in the previous section as one of the tasks the agent fails to zero-shot generalise to. With 30 minutes of finetuning, the agent is able to achieve reward consistently in this task, learning successfully to coax the co-player to the target floor.

These experiments show the potential of massively multi-task RL pre-training, as is performed in this work, for the subsequent transfer with finetuning to many different downstream target tasks.

\subsection{Representation analysis}
\label{sec:representation}
We now move our attention towards understanding how agents operate and the way they represent the simulated environment.

\paragraph{Kohonen Network} There are a multitude of methods to analyse the internal representations of agents and understand what knowledge is encoded in neuron activations~\citep{goh2021multimodal} applicable in various situations. We utilise Kohonen Networks (also known as Self-Organising Maps)~\citep{kohonen1982self} to investigate the high dimensional representations learned by our agents. This technique unifies three types of analysis, allowing us to:
\begin{itemize}
    \item visualise the space of internal representations wrt. some labelling (often done with T-SNE~\citep{van2008visualizing}),
    \item visualise the current state of the agent (\emph{i.e.} a single $\mathbf{h}_t$) (previously done, for example, with a Neural Response Map~\citep{jaderberg2019human}),
    \item conduct simple concept decoding tests (often done with linear classifier probes~\citep{alain2016understanding} or single neuron activity analysis~\citep{quiroga2005invariant}). 
\end{itemize}

A Kohonen Network is a collection of $K$ neurons $\mathfrak{h}_i \in \mathfrak{H} := \mathbb{R}^n$ trained to represent a dataset composed of points $x_j \in \mathbb{R}^n$ under some notion of distance (here we use standard Euclidean distance), using a pre-determined structure between the neurons that prescribe the geometry one is looking for in the dataset. In our work we use neurons arranged as a grid filling a 2-dimensional circle, giving each neuron a fixed position $\mathfrak{k}_i \in \mathfrak{K} := \mathbb{R}^2$. 
To train the network, we iteratively minimise the following per iteration loss using gradient descent
\begin{equation*}
\begin{aligned}
\ell^\mathfrak{H}(\mathfrak{h}) &:= \sum_{i,j} \max \left \{0,\tfrac{\mathrm{d}_\mathrm{max}}{\mathrm{d}_\mathrm{max} -\| \mathfrak{k}_i - \mathfrak{k}_{\iota(x_j)} \|} \right \} \| x_j - \mathfrak{h}_{i} \|^2\\
\iota(x) &:= \arg\min_j  \| x - \mathfrak{h}_{j} \|^2.
\end{aligned}    
\end{equation*}
Intuitively, for each point in the dataset, the closest Kohonen Neuron is selected (the \emph{winning} neuron) and moves the neuron a bit closer to this data point, together with other neurons that are nearby in grid $\mathfrak{K}$ space, with their adaptation downscaled proportionally to how far away from the winning neuron they are.
By fitting the Kohonen Network to the data in this manner, we are asking \emph{what 2d circle-like shape can fit into the n-dimensional dataset in such a way that its position corresponds to the density of the data?} More details can be found in~\secref{app:internals}.
\begin{figure*}[t]
\centering
\includegraphics[width=\linewidth]{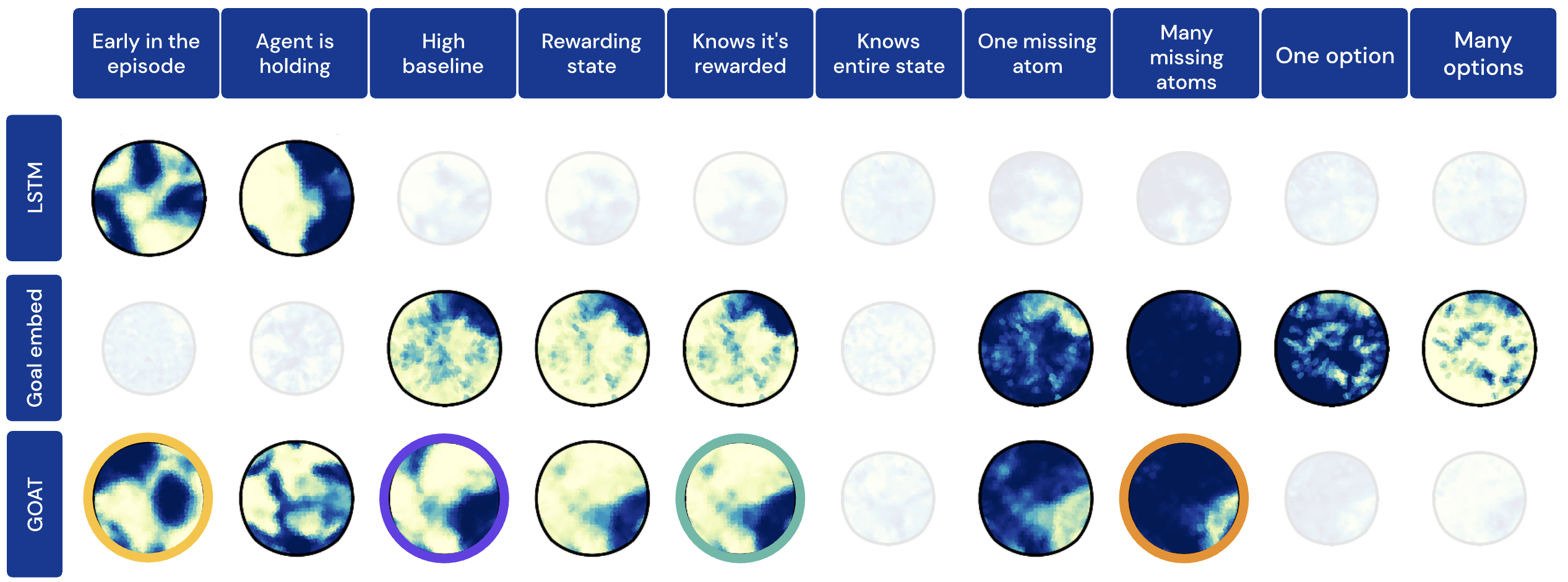}
\caption{Internal representation analysis of the agent. We use Kohonen Network representations of various properties for three different modules of the agent (LSTM, goal embedding, GOAT). Within a Kohonen Network, the bright yellow colour denotes states where the property is true, and blue where it is false. We shade out plots which represent combinations of properties and modules where the given property is not represented in a statistically significant manner by the output of the module (see \secref{sec:representation}).}
\label{fig:internals}
\end{figure*}

We gather 30k episodes of our trained agent across tasks sampled uniformly from the \evalvalid{} set, and use activations $x_j$ of the outputs of the LSTM, goal embedding module, and GOAT module to train three Kohonen Networks respectively. Next, we identified a collection of binary properties corresponding to state $\state_j$ represented in these episodes, \emph{e.g.} whether it is early in the episode, whether the agent is holding an object, whether the agent is in a rewarding state, \emph{etc.} For each probe property $p$ we assign a colour to a specific Kohonen Neuron $\mathfrak{h}_i$ given by the fraction of data points containing the property relative to all the states that were mapped to this neuron:
$$
c_{pi} := \tfrac{\#\{ x_j: i = \iota(x_j)   \wedge p(\state_j)\}}{\#\{ x_j: i = \iota(x_j) \}}.
$$
In \figref{fig:internals} one can see qualitatively that different properties are clearly represented in different parts of the network. To quantify this, we compute the \emph{Balanced Accuracy} (BAC) of a classifier which assigns a label to each state by a majority vote of labels inside each cluster (\emph{i.e.} set of points mapped onto a given neuron), formally:
$$
\hat{p}_{\mathrm{T}_\mathrm{module}}(x) := c_{p \iota(x)} \geq \mathrm{T}_\mathrm{module}
$$
for some threshold $\mathrm{T}_\mathrm{module} \in [0, 1]$, and we compute 
$$
\mathrm{BAC}(\hat{p}, p) := \max_\mathrm{T} \tfrac{1}{2}\left [ \tfrac{\mathrm{TP}[\hat{p}_\mathrm{T}, p]}{\mathrm{TP}[\hat{p}_\mathrm{T}, p] + \mathrm{FN}[\hat{p}_\mathrm{T}, p]} + \tfrac{\mathrm{TN}[\hat{p}_\mathrm{T}, p]}{\mathrm{TN}[\hat{p}_\mathrm{T}, p] + \mathrm{FP}[\hat{p}_\mathrm{T}, p]} \right ],
$$
where $\mathrm{TP, TN, FP, FN}$ is the fraction of true positives, true negatives, false positives and false negatives from a predictor $\hat{p}$ and the ground truth $p$.
We decide that the information is present in a specific representation if and only if $\mathrm{BAC}(\hat{p},p) \geq 0.8$, meaning that if we were to randomly select a state where the property is true or false, we could with at least $80\%$ probability correctly guess this label based purely on the colour of the corresponding Kohonen Neuron.

Using this quantitative measure of information present in \figref{fig:internals}, we can first see that the notion of the flow of time, and whether an agent is holding an object is clearly visible in the LSTM cell output, but is completely missing from the goal embedding module. It is however preserved at the output of the GOAT module, meaning that this information is probably useful for further policy/value predictions.

We can also see that the agent clearly internally represents that it is in a rewarding state. This is significant given that the agent does not receive its reward, nor the past rewards, as an input. The reward signal is used purely as part of RL training, so during inference the agent needs to be able to infer this information from its observations. Consequently, this implies that the agent is capable of using its RGB input to reason about the relations between objects, and their correspondence to the logical structure of the goal at hand. We further investigate whether this representation of a rewarding state is consistent with the agent's internal atomic predicate prediction (denoted in \figref{fig:internals} as \emph{rewarding state and knows it}), where we further require all the atomic predicate predictions that are relevant to the rewarding state (\emph{i.e.} selected option) to be correct. We can see that this information is also very well represented. On the other hand, if we ask whether the agent represents the atomic predicates states of all relations involved in the goal (\emph{i.e.} the atomic predicate states contributing to other options, that agent might not be pursuing right now) we see this information is not present in any of the modules we investigated. This suggests that agent has a very good, but focused, understanding of the state of the world, and attends mainly to the aspects of state that are relevant to the option it is currently following.

We can ask an analogous question of whether the agent is aware of how many atomic predicates states it needs to change before it can obtain a reward. The distinction between having to flip one atomic predicate or more is clearly encoded in the goal embedding module -- with a small island of activations in the upper right corner corresponding to multiple missing atomic predicates, with the smooth big region around it corresponds to needing to flip exactly one. While this information is clearly preserved in the GOAT module output, we can see that they are mapped onto similar regions, suggesting that as the information is processed through the network and reaches the point where only policy/value needs to be produced, this distinction is potentially less relevant.

Finally, details regarding the exact game that an agent is playing (\emph{e.g.} number of options involved) is clearly represented in its goal embedding module, but is then not propagated to the GOAT module, suggesting that whatever decision needs to be made that affects the policy/value can be done solely at the goal embedding level, and does not need to be integrated with the LSTM output.
\begin{figure}[t]
\centering
\includegraphics[width=0.9\linewidth]{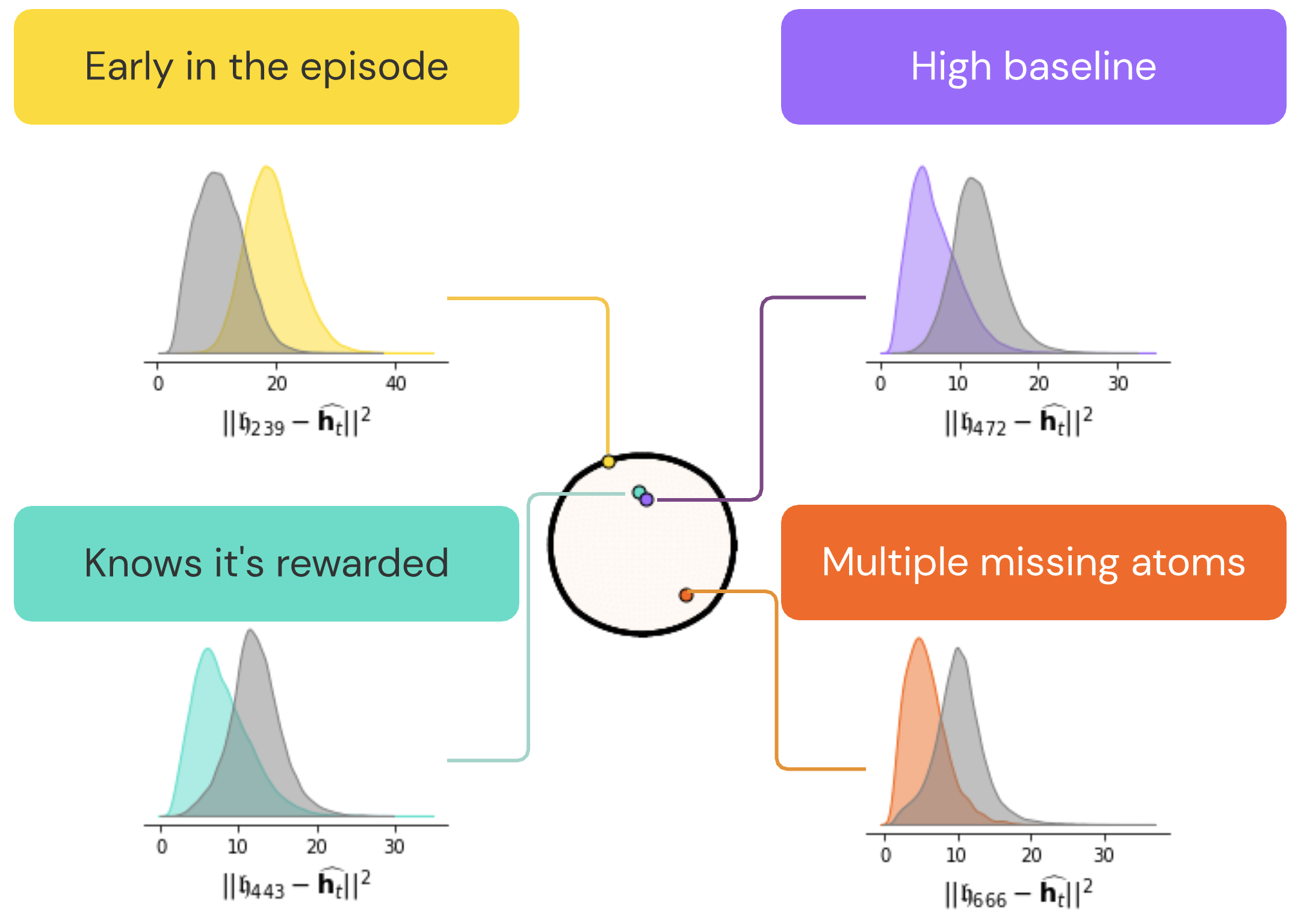}
\caption{Internal representation analysis of the agent. The Kohonen Neurons encode four well represented concepts from \figref{fig:internals}. The kernel density estimation plots represent the density of the activity of the neuron when the concept is true (in colour) or false (in gray).}
\label{fig:neurons}
\end{figure}

\paragraph{Kohonen Neurons} An associated question that one could ask is whether there exists a single Kohonen Neuron coding for a specific property. Note that a Kohonen Neuron does not correspond to a single neuron in a neural network of the agent, but rather a collection of them, found using unsupervised training (and thus more related to general notions of distributed sparse representations than so called grandmother cells~\cite{connor2005friends}). This can be seen more as a distributed concept, though not simply a linear classifier probe~\citep{alain2016understanding}, as the Kohonen Neuron is found without access to the corresponding labels.
$$
\bar{p}_{i\mathrm{T}_\mathrm{neuron}}(x) := c_{p i} \geq \mathrm{T}_\mathrm{neuron}, 
$$
and
$$
\mathrm{BAC}(\bar{p}, p) := \max_{i,\mathrm{T}} \tfrac{1}{2}\left [ \tfrac{\mathrm{TP}[\bar{p}_{i\mathrm{T}}, p]}{\mathrm{TP}[\bar{p}_{i\mathrm{T}}, p] + \mathrm{FN}[\bar{p}_{i\mathrm{T}}, p]} + \tfrac{\mathrm{TN}[\bar{p}_{i\mathrm{T}}, p]}{\mathrm{TN}[\bar{p}_{i\mathrm{T}}, p] + \mathrm{FP}[\bar{p}_{i\mathrm{T}}, p]} \right ].
$$
We note that for being early in the episode, having a high baseline, being in a rewarding state, and for multiple missing atomic predicates, we can identify corresponding Kohonen Neurons achieving $\mathrm{BAC}$ of over 75\%, \figref{fig:neurons}.

\begin{figure}[t]
    \centering
    \includegraphics[width=0.5\linewidth]{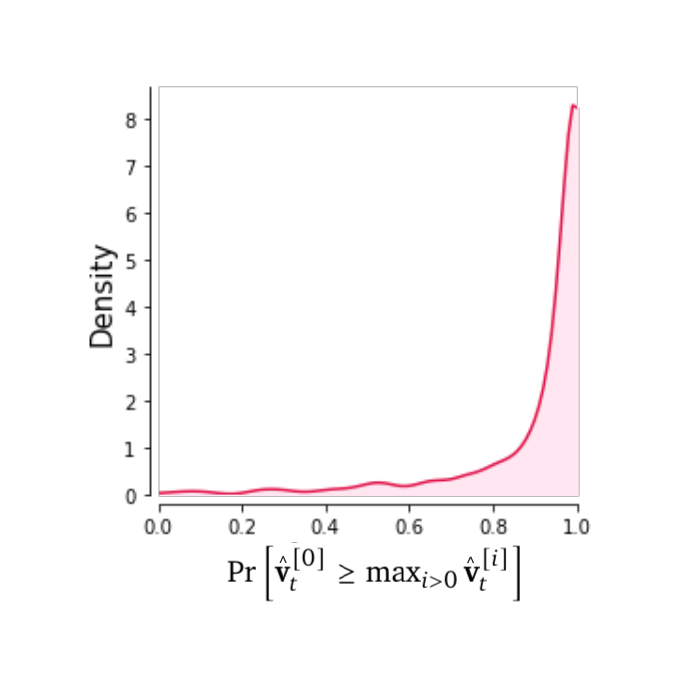}
    \caption{The kernel density estimation of the fraction of frames inside a single episode where the agent's internal value estimation of the whole goal is lower bounded by the maximum value over options (\emph{value consistency}, Theorem~\ref{thm:value_consistency}). We only consider goals with two and three options, as this property is trivially true for one option goals.}
    \label{fig:value_cons}
\end{figure}

\paragraph{Value consistency} In \secref{sec:agent} we discussed \emph{value consistency}, the fact that an optimal policy value of the game composed of multiple alternatives is always lower bounded by the maximum value of the optimal policy for each separate option. Whilst the agent is encouraged to preserve a similar property over its current policy, it is not fully enforced. We investigate how consistent the trained agent is in this respect by looking at its internal values for each option and computing
$
\mathrm{Pr}\left [\goatvaluehead_t^{[0]} \geq \max_{i>0} \goatvaluehead_t^{[i]} \right].
$
In \figref{fig:value_cons} we show the density estimation of episodes where a specific probability of value consistency occurs. In expectation, our agent is value consistent around 90\% of the time (for the goals with more than one option, since by definition an agent is always value consistent with one option goals). Value consistency is clearly shown in a previously discussed example, \figref{fig:multipleoptions}, with the value of the full game upper bounding values of the individual options, even as the individual option values fluctuate.

\section{Related Work}
\label{sec:related}
This work builds heavily upon the ideas of many related works. We now review some of these in the areas of multi-agent learning and progressive learning, iterative improvement and percentiles, procedural environment generation, curriculum over tasks, curriculum over goals, and world-agent co-evolution.

\paragraph{Multi-agent and progressive learning.} 
Our environment is multi-agent, and as such we face challenges of multi-agent learning systems, characterised previously as non-stationarity, exploration, and interaction modelling~\citep{lowe2017multi,mahajan2019maven,bowling2000convergence}. Like others, we also see multi-agent reinforcement learning as a potential solution to other challenges, such as the design of autocurricula~\citep{autocurricula_manifesto_cited} or even end-to-end learning of pixel-perception based agents~\citep{jaderberg2019human}.
The notion of generations of agents, forming a growing set (or \emph{league} in~\cite{vinyals2019grandmaster}) of agents lies at the core of many multi-agent learning algorithms~\citep{psro_cited,balduzzi2019open}. The difference in this work is the utilisation of the generational split to encapsulate self-contained units of reinforcement learning such that the objective can change between generations, as well as the utilisation of a massive space of games being solved in parallel.
This progressive growing of the set of agents on multi-task spaces is also related to progressive learning~\citep{furlanello2018born,schwarz2018progress,rusu2016progressive}, as well as multi-task learning with knowledge sharing~\citep{teh2017distral}.
\cite{contextual_games_cited} proposes a mathematical framework of \emph{contextual games}, which could be used to view XLand goal conditioned agents. They show an effective sampling strategy of scheduling games under an assumption of smoothness of mapping from contexts to optimal policies.
From a formal standpoint the goal space of XLand forms a Boolean algebra and thus can benefit from exploitation of this structure \citep{tasse2020boolean,todorov2009compositionality,van2019composing}.
Currently, we exploit these properties in our GOAT module, as well as how we navigate game space to create games with specific properties.
\cite{vezhnevets2020options} studies architectures and auxiliary losses~\citep{jaderberg2016reinforcement} in a multi-agent setting with hindsight knowledge of agents' goals, which could be applied to our setting. \cite{leibo2017multi} studies sequential social dilemma, in particular trying to identify well known social dilemma classes~\citep{robinson2005topology} in empirical payoffs emerging from RL simulations, which our multi-agent analysis draws upon. Automated identification of varied social dilemma in our setup is an interesting open research question.

\paragraph{Iterative improvement and percentiles.}
Iterative notions of improvements have been used, particularly in multi-agent systems, either explicitly by targeting the goal with respect to known opponents~\citep{mcmahan2003planning,balduzzi2019open}, implicitly by using internal ranking between agents~\citep{jaderberg2019human}, or simply because of the reliance on self-play~\citep{silver2016mastering}. In this work we use similar ideas but applied to worlds and games in addition to other agents (co-players), and propose a way to deal with non-comparable reward scales of the resulting tasks.
When thinking about general capability and catastrophic failure of policies, the field of robust and risk sensitive reinforcement learning~\citep{Borkar2010RiskconstrainedMD,prashanth2013actor,di2012policy} has been analysing variability in obtained rewards to find safer solutions. In particular, percentile-based measures have been utilised~\citep{filar1995percentile,delage2010percentile} to ensure/target specific guarantees of a probability of obtaining a reward in a given task.
In this work we use similar ideas on the level of distribution over tasks, rather than on the level of individual policy outcomes.
The use of curves of normalised score with respect to percentiles to visualise and characterise performance is inspired by ROC curves~\citep{hanley1982meaning}.

\paragraph{Procedural environment generation.} Many previous works have used procedural generation and evolution to create interesting environments for players (both agents and humans). \cite{togelius2008experiment} propose an evolving system to generate interesting rules for a game by selecting games in which random agents score poorly and trained agents score highly. \cite{evolvingmario_cited} use a Generative Adversarial Network (GAN, \cite{gan_cited}) to generate Super Mario Bros levels. They further search the latent space of the GAN using evolutionary methods to discover levels that are difficult but achievable for a previously trained agent. \cite{justesen2018illuminating} train an agent in a procedurally generated environment and update a difficulty parameter based on the agent's recent success rate -- we make use of similar measures to influence task generation in our work. \cite{evocraft_cited} evolve Minecraft levels, both via interactive and automated evolution. CPPN2GAN \citep{cppn2gan_cited} generates large diverse game levels by combining GANs, Content producing Compositional Pattern
Producing Networks (CPPNs, \citep{cppn_cited}) and the NEAT evolutionary algorithm \citep{neat_cited}. The GAN is first trained on a dataset of existing levels to reproduce individual rooms. A CPPN is then evolved to transform grid's coordinate locations into a latent representation that can be input to the GAN. The CPPN is evolved to maximise metrics such as the length of the shortest path to solve a level. In PCGRL \citep{pcgrl_cited}, a deep RL agent is made to edit worlds in order to maximise a bespoke reward function, such as generating long paths for a maze. 

\paragraph{Curriculum over tasks.} Both our procedures for world-agent co-evolution~(\secref{sec:app-procgen}) and dynamic task generation are examples of automated curriculum learning (ACL, \cite{ijcai2020-671}). In ACL, the training distribution of the agent is automatically adapted throughout training. A number of methods attempt to use learning progress \citep{kaplan2007search,schmidhuber2010formal} on a task as a way to decide whether the task should be trained on or not \citep{graves2017automated}. In the context of reinforcement learning, this has been used to select tasks or task parameters \citep{matiisen2020teacher, portelas2020teacher, kanitscheider2021multitask}. \cite{rubiks_cited} automatically adapt the parameters of their environment for solving a Rubik's cube with a robot hand. They start with an narrow domain distribution and continuously expand this distribution when the agent is seen to have good performance at its boundaries. Prioritised Experience Replay \citep{schaul2015prioritized} changes the distribution with which experience is replayed by prioritising those with high Temporal Difference (TD) error. Similarly, \cite{jiang2021prioritized} propose Prioritised Level Replay which samples new levels to play on based on the observed TD error in recent experience on those levels. In CARML, \cite{carml_cited} adapt the task distribution to form a curriculum for meta-RL by maximising the information between a latent task variable and their corresponding trajectories. In PowerPlay, \cite{schmidhuber2013powerplay} propose a framework to continuously seek the simplest unsolved challenge to train on. The adaptation of curricula for many of these works use hand-crafted heuristics, as we do with dynamic task generation, however in our case the parameters of the heuristic itself are adapted with PBT.

\paragraph{Curriculum over goals.} A large body of work is concerned with the training of goal-conditioned agents \citep{uvfa_cited} in a single environment. In these past works, the goal usually consists of the position of the agent or a target observation to reach, however some previous work uses text goals~\citep{colas2020language} for the agent similarly to this work. When the goal is a target observation, most methods acquire new goals by sampling observations previously generated in the environment: \cite{NEURIPS2018_7ec69dd4} generate visual goals by training a Variational Auto-Encoder \citep{vae_cited} over the generated experience. Hindsight Experience Replay (HER, \cite{her_cited}) trains a goal-conditioned agent by replaying trajectories with the agent conditioned on the goal that was achieved in the trajectory. \cite{NEURIPS2019_83715fd4} add a curriculum to Hindsight Experience Replay by dynamically changing the selection of trajectories for replay. \cite{skewfit_cited} propose a method to increase the importance of rarely sampled observation as goals. \cite{discern_cited} propose a variety of goal achievement reward functions which measure how similar a state is to the goal state. \cite{settersolver_cited} perform a curriculum over environment goals in randomly initialised 2D and 3D worlds. A setter generates goals for a solver agent. The setter minimises a few different losses which aim to yield a wide variety of tasks of various difficulties for the current agent policy. CURIOUS \citep{curious_cited} sets a curriculum over environment goals by prioritising goal spaces which have shown recent learning progress and then sampling goals uniformly over goal spaces.

\cite{pmlr-v80-florensa18a} propose an adversarial goal generation procedure in which a goal-GAN generates goals for locomotion tasks that the agent must solve. The objective of the goal setter is similar to that used in our world-agent co-evolution procedure: guarantee that the success probability is within a preset range. \cite{DBLP:conf/nips/ZhangAP20} choose goals where there is high epistemic uncertainty on the Q-function. AMIGo \citep{amigo_cited} also generates a curriculum of goals for the agent but does so by looking at the current number of steps needed by the agent to reach the goal. 

In Asymmetric self-play \citep{asym_sp_1_cited, asym_sp_2_cited}, two agents interact in turn in the environment: Alice and Bob. Alice first plays in the environment and generates a trajectory. From there, Bob can either be tasked with returning the player to its original location, or, in a new episode, reaching the same state that Alice achieved. The self reward-play modification (\secref{sec:gentraining}) can be seen as a sequential version of this within a single episode and the same agent playing both Alice and Bob. 

\paragraph{World-agent co-evolution.} Our procedure for world-agent co-evolution~(\secref{sec:app-procgen}) shares similarity with POET~\citep{wang2019poet, wang2020poet} and PAIRED~\citep{dennis2020emergent}. In all cases, the procedure generates a dynamic high-dimensional world distribution for agents. In POET, a population of environment-agent pairs is evolved through time. Agents are continuously trained on their paired environment. Occasionally, agents are transferred to other environments in the population. In PAIRED, two agents are coevolved: a protagonist agent and an antagonist agent. The protagonist agent attempts to solve tasks generated by the antagonist agent. The antagonist also plays in the generated environments. The difference between the average score of the protagonist and the best score of the antagonist across multiple trials is defined as the regret. The protagonist is trained to minimise this regret while the antagonist is trained to maximise it. Compared with both these methods, our proposed procedure is simpler: it only requires a single agent to be trained to solve tasks. We filter levels only based on the agent's estimated probability of success. Finally, the use of the world-agent co-evolution process to create the base distribution for training and evaluation for the remainder of our learning process is an example of AI-generating algorithms~\citep{clune2019ai}.

\section{Conclusions}
\label{sec:conclusions}
In this work, we introduced an open-ended 3D simulated environment space for training and evaluating artificial agents. We showed that this environment space, XLand, spans a vast, diverse, and smooth task space, being composed of procedurally generated worlds and multiplayer games. We looked to create agents that are generally capable in this environment space -- agents which do not catastrophically fail, are competent on many tasks, and exhibit broad ability rather than narrow expertise. An iteratively revised metric of normalised score percentiles on an evaluation set of tasks was used to characterise general capability, and a learning process to drive iterative improvement created. This learning process is composed of agents training with deep RL, on training task distributions that are dynamically generated in response to the agent's behaviour. Populations of agents are trained sequentially, with each generation of agents distilling from the best agent in the previous generation, iteratively improving the frontier of normalised score percentiles, whilst redefining the metric itself -- an open-ended learning process.

Combining this environment space with such a learning process resulted in agents that appear to have broad ability across our held-out evaluation space, catastrophically failing on only a small percentage of tasks that are humanly impossible. We qualitatively and quantitatively characterised some of the emergent behaviours of this agent and saw general behavioural heuristics such as experimentation and success recognition, and the tendency to cooperate more with other competent agents, behaviours which appear to generalise to many out-of-distribution probe tasks. These behaviours are driven by rich internal representations that we analysed, showing clear representations of the structure and state of the goals they are tasked to follow.

These results hint at the ability to train agents, without human demonstrations, which exhibit general capabilities across vast task spaces. Beyond zero-shot generalisation, the ability to quickly finetune these pretrained agents on complex out-of-distribution tasks was demonstrated clearly. We hope the presented methods and results pave the way for future work on creating ever more adaptive agents that are able to transfer to ever more complex tasks.

\section*{Author Contributions}
\label{sec:author}
The following lists the main contributions of the authors to the work presented.
\\\\\textbf{Adam Stooke:} Learning process development and research investigations.
\\\textbf{Anuj Mahajan:} Agent analysis.
\\\textbf{Catarina Barros:} Environment development and visuals.
\\\textbf{Charlie Deck:} Environment development.
\\\textbf{Jakob Bauer:} Infrastructure development, learning process development, research investigations, and technical management.
\\\textbf{Jakub Sygnowski:} Infrastructure development, agent analysis, and research investigations.
\\\textbf{Maja Trebacz:} Research investigations.
\\\textbf{Max Jaderberg:} Learning process development, research investigations, manuscript, visuals, XLand concept, and team management.
\\\textbf{Michael Mathieu:} Learning process development and research investigations.
\\\textbf{Nat McAleese:}  Infrastructure development and research investigations.
\\\textbf{Nathalie Bradley-Schmieg:} Program management.
\\\textbf{Nathaniel Wong:} Environment development and visuals.
\\\textbf{Nicolas Porcel:} Environment development.
\\\textbf{Roberta Raileanu:} Research investigations.
\\\textbf{Steph Hughes-Fitt:} Program management.
\\\textbf{Valentin Dalibard:} Learning process development, infrastructure development, research investigations, agent analysis, and manuscript.
\\\textbf{Wojciech Marian Czarnecki:} Learning process development, research investigations, agent analysis, manuscript, visuals, and XLand concept.
\\\\All authors shaped the final manuscript.

\section*{Acknowledgements}
We would like to thank Simon Osindero, Guy Lever, and Oriol Vinyals for reviewing the manuscript, Satinder Singh and Koray Kavukcuoglu for support of the project, and Marcus Wainwright and Tom Hudson for additional environment art and support. We also thank the wider DeepMind research, engineering, and environment teams for the technical and intellectual infrastructure upon which this work is built.

\bibliographystyle{abbrvnat}
\nobibliography*
\bibliography{template_refs}

\appendix
\renewcommand{\appendix}{}

\section{Appendix}
\begin{figure}[t]
    \centering
    \begin{tabular}{cc}
    \includegraphics[width=\linewidth]{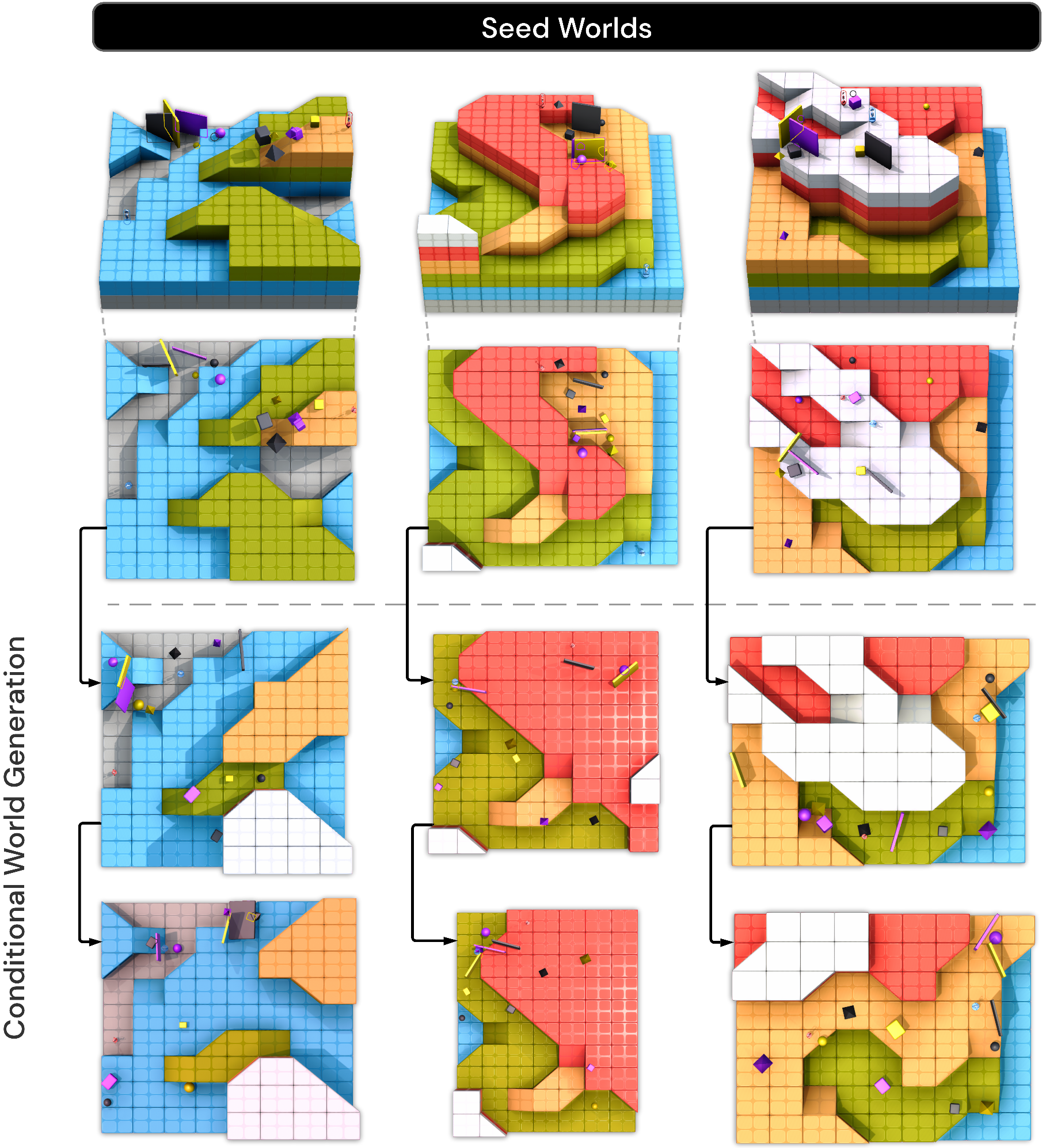}
    \end{tabular}
    \caption{Worlds can be generated conditioned on an existing world. This allows smooth variation of worlds. This figures shows three examples of this process, with each column showing an initial seed world and the rows showing two steps of conditional world generation.}
    \label{fig:worldmutation}
\end{figure}
\begin{figure*}[t]
    \centering
    \begin{tabular}{cc}
    \includegraphics[width=\textwidth]{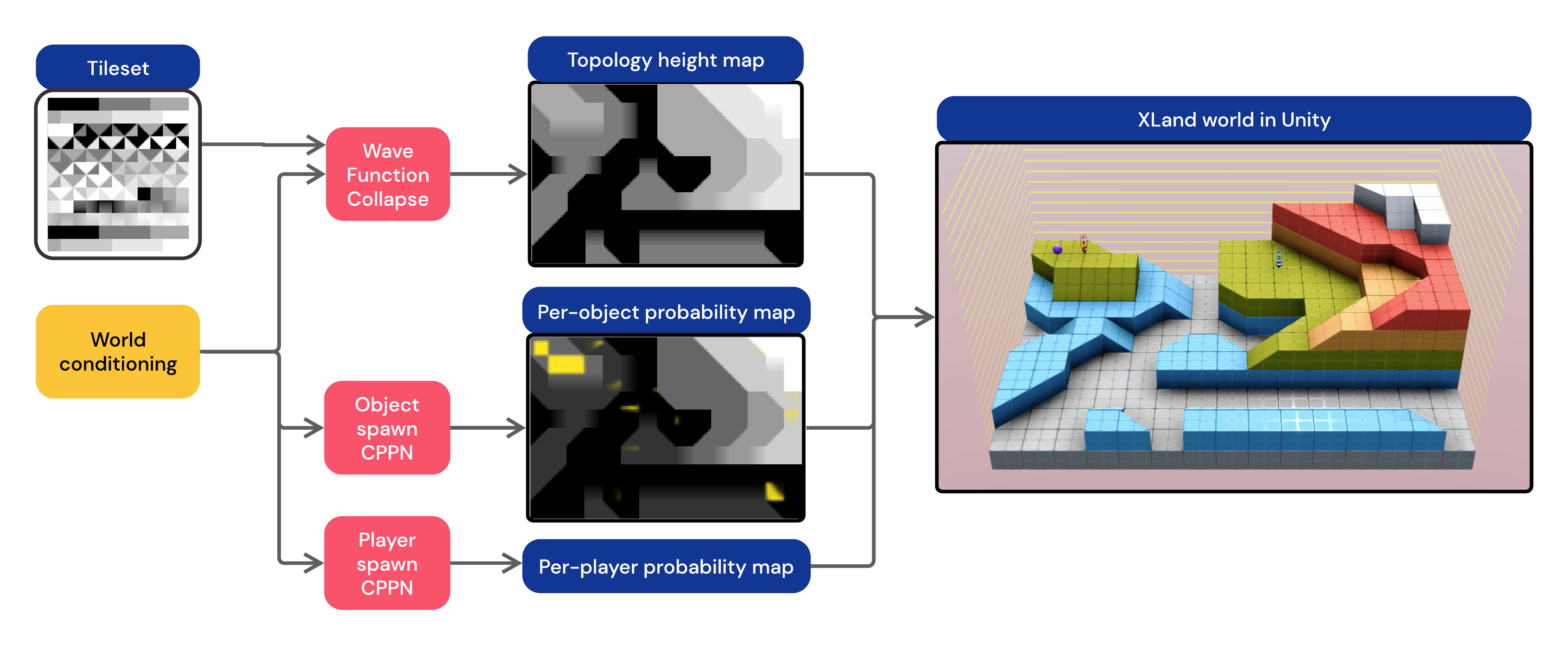}
    \end{tabular}
    \caption{The steps of procedural world generation. The process is conditioned on a seed and optionally an existing world to be similar to. Wave Function Collapse~\citep{wavefunctioncollapse} acting on a tileset of primitive building blocks creates a height map of the topology. Additionally, for each object and player, a CPPN~\citep{stanley2007compositional,ha2016abstract} creates a probability map for the entity's initial spawn location. These elements are combined in the Unity game engine to produce the playable world.}
    \label{fig:worldgen}
\end{figure*}

\subsection{Worlds}

\subsubsection{Procedural World Generation}
\label{sec:app-procgen}

\figref{fig:worldgen} gives an overview of the generation process which we now describe in more detail for each of the components.

\paragraph{Topology}
Tile assignments are procedurally generated using the Wave Function Collapse algorithm (WFC)~\citep{wavefunctioncollapse}. WFC acts as a constraint satisfaction algorithm, acting on a set of tiles which correspond to 3D geometry such as floor elements of different heights, ramps, and diagonal floor pieces with associated connectivity rules. WFC iteratively samples a grid location, samples a tile to be assigned to that location, and then updates each remaining grid locations' probability distribution over what tiles can be sampled given the constraints defined by tile connectivity rules. This process repeats, and ends when all grid locations have an assigned tile. The largest connected component of exposed floor is defined as the playable area of the world. Finally, a random scaling of the three dimensions of the tile elements is performed to create non-cuboidal tile elements, and random lighting applied (location, direction, intensity, hue). We additionally randomly apply reflections of topology to sometimes create symmetric worlds. The result is the ability to procedurally generate a wide variety of convex topologies composed of varied yet coherent structures.

\paragraph{Objects}
An object's initial position in the world is determined by sampling from a 2D probability map corresponding to the top-down map of the world topology, with non-zero values in the playable area of the world, and subsequently positioning the object in 3D at the floor level at the sampled 2D point. The probability map is given by a fixed parameter Compositional Pattern-Producing Network (CPPN)~\citep{stanley2007compositional,ha2016abstract} which takes in the 2D position, height, and tile identity at each position in the map, and a position-independent latent variable associated with the instance of the object. This allows the probability of object placement to be associated with certain floors, elements of the topology, absolute or relative locations, in a highly non-linear manner determined by the latent variable. Object instances have a randomly sampled size, a colour, and a shape. There are three colours -- black, purple, yellow -- and four shapes -- cube, sphere, pyramid, slab. Object locations can be sampled independently as per the process previously described, or in spatial groups clustered by shape or colour.

\paragraph{Conditional world generation}
The mechanisms described so far allow us to create a world generating function, where each world sample is drawn independently $\world \sim P_\worldspace(\cdot)$. However, it is also possible to condition the world sampling such that a new world $\hat{\world}\sim P_\worldspace(\world)$ is similar to a given world $\world$.
To achieve this for the topology, we can bias the initial probability over each grid location by the delta function of the conditioned worlds topology. For the object and player locations we add Gaussian noise to the latent variable associated with each object and player, and for all other categorically sampled quantities we resample. Some examples of this process is shown in \figref{fig:worldmutation}. We show in \secref{sec:worldproperties} that this results in the ability to smoothly vary worlds and can be used to generate worlds via an evolutionary process.

\paragraph{Game conditioned worlds}
We can also condition the generation of worlds such that a particular game $\game$ is achievable given the topology and layout of objects. We define a function $\world_\game = f(\world, \game)$ which takes an existing world $\world$ and a game $\game$ and returns a new world $\world_\game$ such that all players, objects, and topological elements (\emph{e.g.} floors) referenced in the game will be present and exposed in the playable area.

\paragraph{World-agent co-evolution}
Whilst our procedural world generation function $P_\worldspace(\cdot)$ has a vast and diverse support, it is interesting to consider how to shift the distribution towards more interesting worlds, particularly those that may pose navigational challenges to players. To explore this, we created an RL training process to train an agent to maximise reward on a dynamic set of worlds, but with a static game which always consist of a single player (the agent) with the goal \emph{``Be near a yellow cube''}. Our procedure maintains two sets of worlds which dynamically change as the agent trains: the train set and the evaluate set.
When the agent requests a world to train on, we sample one uniformly from the train set.

\begin{figure*}[t]
    \centering
    \includegraphics[width=\linewidth]{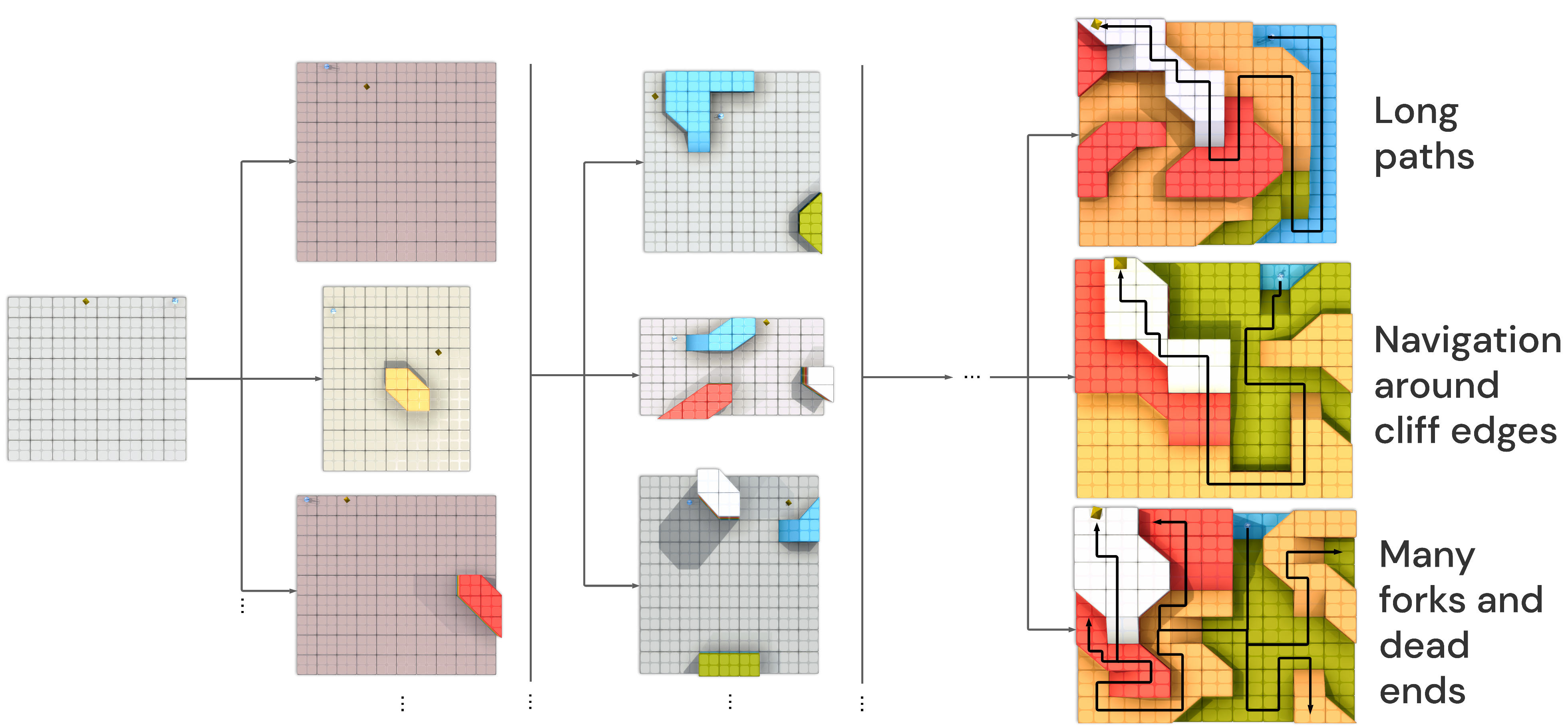}
    \caption{The process of world-agent co-evolution resulting in complex worlds. \textbf{(Left)} The initial seed world to the train world set. The agent is trained to maximise the reward of its goal \emph{``Be near a yellow pyramid''}. \textbf{(Middle)} The progression of the worlds in the train set as training progresses. Worlds undergo evolution with a minimum criterion fitness such that a world must be solved sometimes but not too often by the agent. \textbf{(Right)} The resulting world set is more diverse in terms of the navigational feature space and exhibit interesting topological elements.}
    \label{fig:worldevo}
\end{figure*}

The train set initially contains only a world consisting of an open room (\figref{fig:worldevo} (left)) and the evaluate set is set to be empty.  An evolutionary algorithm governs the progression of the train worlds similarly to criterion co-evolution~\citep{brant2017minimal}. At regular intervals, our procedure attempts to generate a new world to add to the train set. It selects a random world from the existing train set. It then mutates this world using the conditional world generation process described previously, $\world_\text{child} \sim P_\worldspace(\world_\text{parent})$. This new world is added into the evaluate set. To evaluate a world in the evaluate set, we make the agent play in the world for 100 episodes. None of these episodes are used for RL training. 
If the agent scores a reward in at least one episode but less than half of the 100 episodes, the corresponding world is added to the train set.
For each world in the train set, we also monitor the agent scores across the last 100 episodes and discard the world if the agent does not still meet this criterion.

The agent is therefore continually training on worlds which pose some navigational challenge, though not too challenging. Since the fitness of worlds is related to the agent's behaviour, as the agent trains and improves in its navigational behaviour, we observe the complexity of the worlds in the train set continually increases. \figref{fig:worldevo}~(middle) illustrates the train set of this process as both the agent and the train world distribution co-evolve. We can see in \figref{fig:worldevo}~(right) how these worlds exhibit some interesting features such as long navigational paths, forks in paths that can be taken, thin paths such that the agent easily fall off, and hidden goal objects.

\subsubsection{Counting worlds}
\label{app:worldcount}
\begin{figure}[t]
    \centering
    \includegraphics[width=\linewidth]{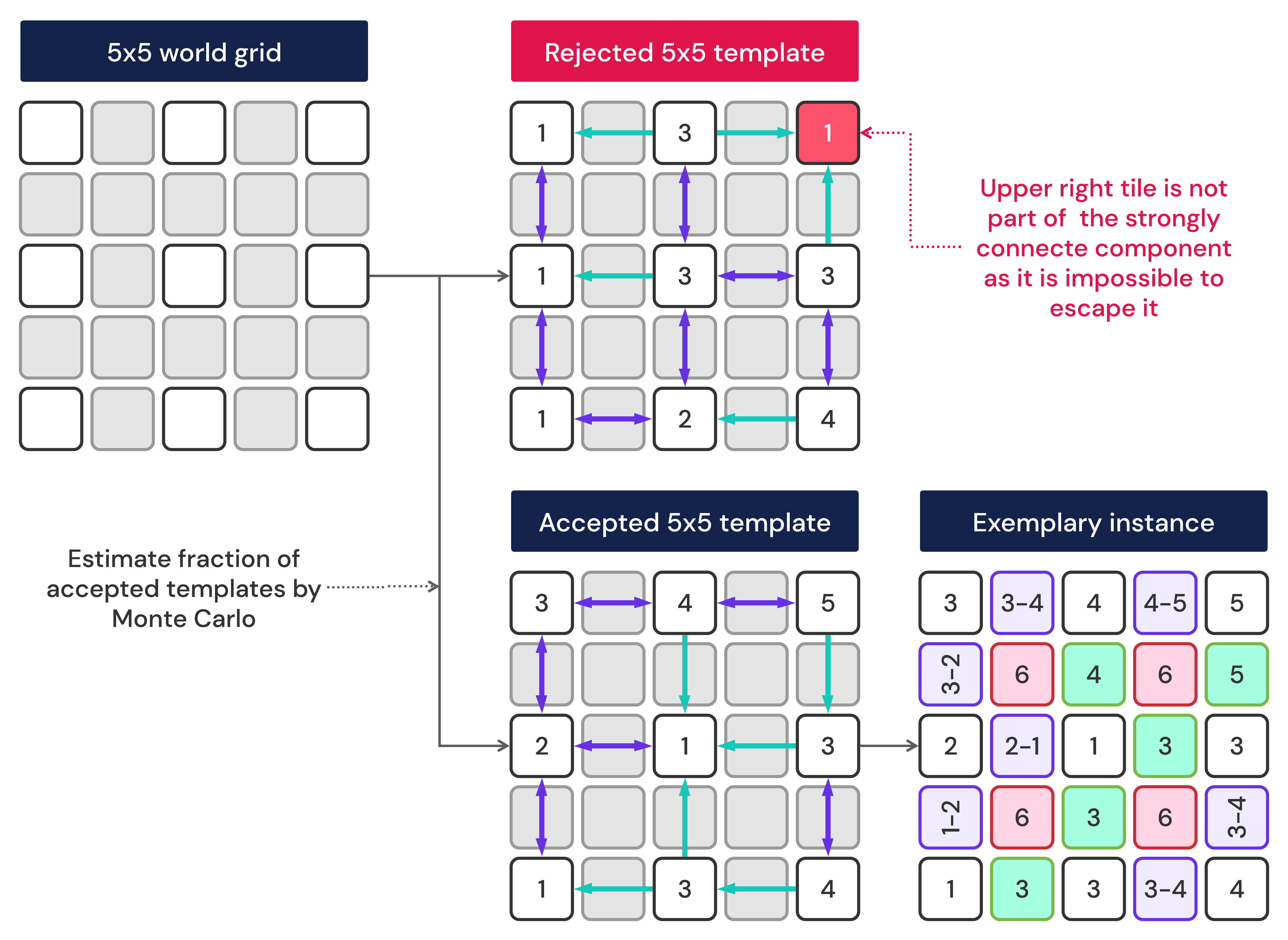}
    \caption{Visualisation of the process used to estimate the number of correct worlds of a given size. Cyan lines represent one directional edges, while purple ones bidirectional. In the exemplary instance, filled tiles have colours corresponding to the edges they replaced, and in red we we see "pillars" used to fill in missing pieces.}
    \label{fig:worldcountalg}
\end{figure}

We would like to count how many $w \times h$ worlds are there, such that there exists a region $A$ with the following properties:
\begin{itemize}
    \item its size is at least $\tfrac{wh}{2}$.
    \item for every two points in $A$ there exists a path between them.
    \item there is no path that leads from $A$ outside (and thus there are no irreversible decisions of leaving the region).
\end{itemize}
Due to the complexity of this task, we provide a lower and upper bound rather than the exact number of worlds.
The upper bound is trivial: we have 6 possible flat tiles (one per level), 4 possible orientations of ramps and 4 possible orientations of "diagonal tile". Consequently, we have at most $6^{w \cdot h} \cdot (1 + 4 + 4)^{w \cdot h}$ such worlds.
We now turn our attention to a lower bound. Let us take a world and focus on every other tile (and thus operate on the $\lceil \tfrac{w}{2} \rceil \times \lceil \tfrac{h}{2} \rceil$ subgrid, $\world'$, see \figref{fig:worldcountalg}). 
We argue that if after assigning floor levels to each point in this subgrid the resulting graph $G_{\world'}$ has a single strongly connected component (SCC), then there exists at least one world in the full grid that satisfies the desiderata.
Because the graph is strongly connected (has one SCC) this means that there is a path between every two points, and naturally there is no path leaving this world.
However this world is at most 1/4th of the size of the $\world$, thus we embed it in our bigger world, by filling in the missing tiles.
For every edge of $G_{\world'}$ that is bidirectional we put a corresponding ramp (purple in \figref{fig:worldcountalg}), if the edge is one directional we fill it with a flat tile at the height of maximum of neighbouring heights (teal in \figref{fig:worldcountalg}). We fill the remaining 1/4th of tiles with the highest floor (red in \figref{fig:worldcountalg}). 
We treat $\world'$, together with tiles that we added in place of edges (as well as potentially some of the highest floors if they are accessible) as our region $A$. This $A$ is at least of size 75\% of $\world$. Every pair of points has a path between them, since by construction there is a path in $\world'$, and the tiles we added do not form any dead ends. In particular if any of the highest floors become accessible, the player can always jump down from it to a neighbouring tile of lower height. Consequently there are no paths leaving $A$.

To compute the exact number we take number of all possible $\world'$ which is $6^{\tfrac{wh}{4}}$ and then estimate the probability of such a world forming exactly one SCC by Monte Carlo sampling with 200,000 worlds (Table~\ref{tab:mcworldsize}). For simplicity we also put $n=w=h$.
Additionally, if $w$ ($h$) is even, then there is a final column (row) at the edge of the world that is less constrained with respect to our desiderata: for each of $\tfrac{h}{2}$ ($\tfrac{w}{2}$) tiles we can fill in the neighbouring tiles by either the neighbour value, or a 1-off version. This leads to an additional factor of $2^{w+h}$.
\begin{table}[t]
\centering
\begin{tabular}{lll}
\toprule
$n$&
$\lceil \tfrac{n}{2} \rceil$ & $\mathrm{Prob}[\#\text{SCC}(\world')=1]$\\
\midrule
1 & 1 & $= 100\%$\\
2 & 1 & $= 100\%$\\
3 & 2 & $\approx 16\%$ \\
4 & 2 & $\approx 16\%$ \\
5 & 3 & $\approx 3\%$ \\
6 & 3 & $\approx 3\%$ \\
7 & 4 & $\approx 0.5\%$ \\
8 & 4 & $\approx 0.5\%$ \\
9 & 5 & $\approx 0.07\%$ \\
10& 5 & $\approx 0.07\%$ \\
11& 6 & $\approx 0.005\%$ \\
12& 6 & $\approx 0.005\%$ \\
13& 7 & $\approx 0.0006\%$ \\
14& 7 & $\approx 0.0006\%$ \\
\bottomrule
\end{tabular}
\caption{Monte Carlo estimations of the fraction of worlds with a single connected component as a function of the world size $n$ (so the world is $n \times n$ tiles). We use 200,000 samples.}
\label{tab:mcworldsize}
\end{table}

\subsubsection{Worlds linear projection}
\label{app:worldprojection}
In order to find a linear projection of the world space that visualises some property of interest $h_{\cdot}(\world) : \worldspace \rightarrow \mathbb{R}_+$ we
define a simple objective:
$$
\ell_\mathrm{smooth}(\theta) := 
\sum_\world \big \|\| \underset{\mathrm{projection}}{\langle \underbrace{\widehat{\tau}(\world), \theta_\mathbf{W} \rangle}} - \theta_\mathbf{b} \|^2 - (\underset{\mathrm{target}}{\underbrace{h_\cdot(\world)}} - \theta_c)  \big \|^2 .
$$
and find a projection $\theta = (\theta_\mathbf{W}, \theta_\mathbf{b}, \theta_c)$ through gradient descent with learning rate 0.1 trained for 30,000 iterations.
We denote by $\widehat{\tau}$ a vectorised version of the topology projection $\tau$. For flat tiles, $\tau$ simply assigns the normalised floor level (e.g. for a tile $t_{k}$ at level $k \in \{0, \dots, 5\}$ we have $\tau(t_k) = \tfrac{1}{5}k$), and for simplicity ramps are assigned a half a floor value, i.e. for a ramp $t_{1\leftrightarrow 2}$ between floor 1 and 2 $\tau(t_{1\leftrightarrow 2}) = \tfrac{1}{5} \cdot 1.5$.

Intuitively, we seek a linear projection $\theta_\mathbf{W} \in \mathbb{R}^{w \cdot h \times 2}$ in the space $\widehat{\tau}(\worldspace) \subset \mathbb{R}^{w \cdot h}$ such that the distance from some point $\theta_\mathbf{b} \in \mathbb{R}^2$ in the embedding space corresponds to the distance between the target property and some arbitrary learned bias $\theta_c \in \mathbb{R}$. This can naturally lead to finding projections where the property of interest shows circular like placement (since we use Euclidean distance from a point as a predictor), but of course can also look more linear, if $\theta_\mathbf{b}$ is placed very far away from the projection of the data.

For our analysis we used $h_{\rho}(\world) := \mathrm{H}_2(\rho(\world))$ and 
 $h_{\rho_\mathrm{sp}}(\world) := \mathrm{H}_2(\rho_\mathrm{sp}(\world))$ (note that entropy is always positive and thus satisfies the assumptions).
 
 In order to avoid spurious results coming from a rich parameter space (81) compared to number of samples (around 250) we exploit the fact that each of $h_\cdot$ is invariant to rotations and symmetries, and learn a projection that is invariant to them too. To achieve this we simply parametrise $\theta_\mathbf{w}$ (which is itself a 9 by 9 world) to be an average of all 8 rotations/symmetries. Thanks to that, we reduce the number of actual trainable weights in the linear projection from 81 to 15, significantly reducing overfitting. 
 
\begin{figure}[t]
     \centering
     \includegraphics[width=\linewidth]{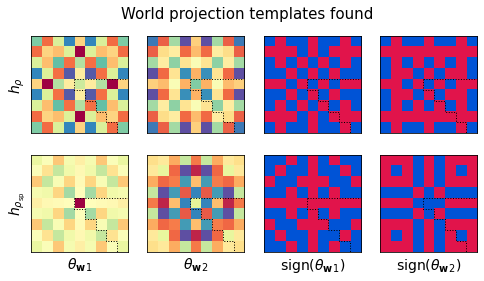}
     \caption{Linear projections $\theta_\mathbf{w}$ found from the process of linearly embedding world topologies. The first two columns represent the 2 projection dimensions, and the following two columns simply show only the sign of each entry to emphasise the pattern. We can see a checkerboard like structure being discovered which naturally translates to navigational complexity/distribution of shortest paths. Dotted lines show the effective size of the learnable parameters, with remaining ones emerging from learning in the space invariant to symmetries.}
     \label{fig:worldtemplate}
 \end{figure}

 We can see templates found in \figref{fig:worldtemplate}, where one can notice checkerboard like patterns being learned. To some extent this is a world-topology navigational complexity analogue to edge detectors from computer vision.

\subsection{Games}
\label{app:gamespace}
The following technical details of game space describe our current instance of XLand. Nothing in the system is constrained to using only the following relations, predicates, number of players, options etc. and can easily be expanded to cover an even richer space.

\subsubsection{Relations}
\paragraph{\texttt{near(a,b)}.} Is true if object \texttt{a} is at most 1 meter away from object \texttt{b} (for reference a player's avatar is 1.65m tall).
In order to measure this we first calculate the center of mass of each object, and find the  point on an object that is closest to the center of mass of the other. We find a mid-point between these two points on objects surfaces. We compute closest points on the surfaces of the objects to this midpoint, and measure the distance between them. This construction is relatively cheap to compute, and takes into consideration the size of the object (as in we do not merely measure distance between centers of mass, but actual body of the object).

\paragraph{\texttt{see(a,b)}.} If \texttt{a} is not a player, then it is evaluated by drawing a line connecting centers of masses of both objects and checking if it is unobstructed. If \texttt{a} is a player then it evaluates to true if and only if \texttt{b} would render in the field of view of \texttt{a}.

\paragraph{\texttt{on(a,b)}.} This relation can only be used on a floor and a player/object. It evaluates to true if and only if the given player/object \texttt{a} is in contact with the upper surface of the given floor colour \texttt{b}.

\paragraph{\texttt{hold(a,b)}.} This relation can only be used on a player and object. It evaluates to true if a player is holding a specific object with its beam. Note that it is technically possible for two agents to hold the same object at the same time.

\begin{figure}[t]
    \centering
    \includegraphics[width=0.47\linewidth]{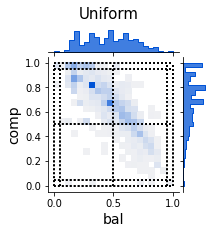}
    \includegraphics[width=0.47\linewidth]{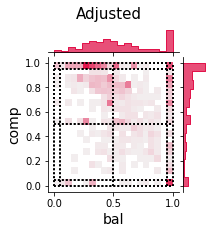}
    \caption{The distributions of competitiveness and balance of games created by different sampling mechanisms. \textbf{(left)}
    We uniformly sample matrices and attach random relations for 1000 games with 1, 2 and 3 options, and see that extreme values are never sampled, and there is a huge correlation between balance and competitiveness, e.g. leading to no fully balanced and fully competitive games. \textbf{(right)} After using our local search procedure to seek values of $b^*,c^*$ in various ranges (rectangular areas with dotted lines) we easily populate each entry including the extreme values.}
    \label{fig:uniformgames}
\end{figure}
\subsubsection{Atomic predicates}
 XLand currently consists of every possible instantiation of the above 4 relations between 9 objects (3 colours -- purple, black and yellow -- and 3 shapes -- pyramid, cube and sphere), 2 player references (me, opponent) and 5 floor colours. This leads to 212 unique atomic predicates listed in Table~\ref{tab:atoms} after taking into account logical symmetry, such as $\goal := \texttt{near(a,b)} \equiv \texttt{near(b,a)} =: \goal'$ in\ the sense that $\reward_\goal = \reward_{\goal'}$. Similarly \texttt{see} is symmetric if both arguments are objects (since objects have no orientation), but is not symmetric if at least one player is involved (since they have a directional vision).
\begin{table*}
\begin{tabular}{ccccccccccc}
     \tiny{\texttt{hold(me,black cube)}} &
\tiny{\texttt{hold(me,black pyramid)}} &
\tiny{\texttt{hold(me,black sphere)}} &
\tiny{\texttt{hold(me,purple cube)}} &
\\
\tiny{\texttt{hold(me,purple pyramid)}} &
\tiny{\texttt{hold(me,purple sphere)}} &
\tiny{\texttt{hold(me,yellow cube)}} &
\tiny{\texttt{hold(me,yellow pyramid)}} &
\\
\tiny{\texttt{hold(me,yellow sphere)}} &
\tiny{\texttt{hold(opponent,black cube)}} &
\tiny{\texttt{hold(opponent,black pyramid)}} &
\tiny{\texttt{hold(opponent,black sphere)}} &
\\
\tiny{\texttt{hold(opponent,purple cube)}} &
\tiny{\texttt{hold(opponent,purple pyramid)}} &
\tiny{\texttt{hold(opponent,purple sphere)}} &
\tiny{\texttt{hold(opponent,yellow cube)}} &
\\
\tiny{\texttt{hold(opponent,yellow pyramid)}} &
\tiny{\texttt{hold(opponent,yellow sphere)}} &
\tiny{\texttt{near(black cube,black pyramid)}} &
\tiny{\texttt{near(black cube,black sphere)}} &
\\
\tiny{\texttt{near(black cube,me)}} &
\tiny{\texttt{near(black cube,opponent)}} &
\tiny{\texttt{near(black cube,purple cube)}} &
\tiny{\texttt{near(black cube,purple pyramid)}} &
\\
\tiny{\texttt{near(black cube,purple sphere)}} &
\tiny{\texttt{near(black cube,yellow cube)}} &
\tiny{\texttt{near(black cube,yellow pyramid)}} &
\tiny{\texttt{near(black cube,yellow sphere)}} &
\\
\tiny{\texttt{near(black pyramid,black sphere)}} &
\tiny{\texttt{near(black pyramid,me)}} &
\tiny{\texttt{near(black pyramid,opponent)}} &
\tiny{\texttt{near(black pyramid,purple cube)}} &
\\
\tiny{\texttt{near(black pyramid,purple pyramid)}} &
\tiny{\texttt{near(black pyramid,purple sphere)}} &
\tiny{\texttt{near(black pyramid,yellow cube)}} &
\tiny{\texttt{near(black pyramid,yellow pyramid)}} &
\\
\tiny{\texttt{near(black pyramid,yellow sphere)}} &
\tiny{\texttt{near(black sphere,me)}} &
\tiny{\texttt{near(black sphere,opponent)}} &
\tiny{\texttt{near(black sphere,purple cube)}} &
\\
\tiny{\texttt{near(black sphere,purple pyramid)}} &
\tiny{\texttt{near(black sphere,purple sphere)}} &
\tiny{\texttt{near(black sphere,yellow cube)}} &
\tiny{\texttt{near(black sphere,yellow pyramid)}} &
\\
\tiny{\texttt{near(black sphere,yellow sphere)}} &
\tiny{\texttt{near(me,purple cube)}} &
\tiny{\texttt{near(me,purple pyramid)}} &
\tiny{\texttt{near(me,purple sphere)}} &
\\
\tiny{\texttt{near(me,yellow cube)}} &
\tiny{\texttt{near(me,yellow pyramid)}} &
\tiny{\texttt{near(me,yellow sphere)}} &
\tiny{\texttt{near(opponent,purple cube)}} &
\\
\tiny{\texttt{near(opponent,purple pyramid)}} &
\tiny{\texttt{near(opponent,purple sphere)}} &
\tiny{\texttt{near(opponent,yellow cube)}} &
\tiny{\texttt{near(opponent,yellow pyramid)}} &
\\
\tiny{\texttt{near(opponent,yellow sphere)}} &
\tiny{\texttt{near(purple cube,purple pyramid)}} &
\tiny{\texttt{near(purple cube,purple sphere)}} &
\tiny{\texttt{near(purple cube,yellow cube)}} &
\\
\tiny{\texttt{near(purple cube,yellow pyramid)}} &
\tiny{\texttt{near(purple cube,yellow sphere)}} &
\tiny{\texttt{near(purple pyramid,purple sphere)}} &
\tiny{\texttt{near(purple pyramid,yellow cube)}} &
\\
\tiny{\texttt{near(purple pyramid,yellow pyramid)}} &
\tiny{\texttt{near(purple pyramid,yellow sphere)}} &
\tiny{\texttt{near(purple sphere,yellow cube)}} &
\tiny{\texttt{near(purple sphere,yellow pyramid)}} &
\\
\tiny{\texttt{near(purple sphere,yellow sphere)}} &
\tiny{\texttt{near(yellow cube,yellow pyramid)}} &
\tiny{\texttt{near(yellow cube,yellow sphere)}} &
\tiny{\texttt{near(yellow pyramid,yellow sphere)}} &
\\
\tiny{\texttt{see(black cube,black pyramid)}} &
\tiny{\texttt{see(black cube,black sphere)}} &
\tiny{\texttt{see(black cube,me)}} &
\tiny{\texttt{see(black cube,opponent)}} &
\\
\tiny{\texttt{see(black cube,purple cube)}} &
\tiny{\texttt{see(black cube,purple pyramid)}} &
\tiny{\texttt{see(black cube,purple sphere)}} &
\tiny{\texttt{see(black cube,yellow cube)}} &
\\
\tiny{\texttt{see(black cube,yellow pyramid)}} &
\tiny{\texttt{see(black cube,yellow sphere)}} &
\tiny{\texttt{see(black pyramid,black sphere)}} &
\tiny{\texttt{see(black pyramid,me)}} &
\\
\tiny{\texttt{see(black pyramid,opponent)}} &
\tiny{\texttt{see(black pyramid,purple cube)}} &
\tiny{\texttt{see(black pyramid,purple pyramid)}} &
\tiny{\texttt{see(black pyramid,purple sphere)}} &
\\
\tiny{\texttt{see(black pyramid,yellow cube)}} &
\tiny{\texttt{see(black pyramid,yellow pyramid)}} &
\tiny{\texttt{see(black pyramid,yellow sphere)}} &
\tiny{\texttt{see(black sphere,me)}} &
\\
\tiny{\texttt{see(black sphere,opponent)}} &
\tiny{\texttt{see(black sphere,purple cube)}} &
\tiny{\texttt{see(black sphere,purple pyramid)}} &
\tiny{\texttt{see(black sphere,purple sphere)}} &
\\
\tiny{\texttt{see(black sphere,yellow cube)}} &
\tiny{\texttt{see(black sphere,yellow pyramid)}} &
\tiny{\texttt{see(black sphere,yellow sphere)}} &
\tiny{\texttt{see(me,black cube)}} &
\\
\tiny{\texttt{see(me,black pyramid)}} &
\tiny{\texttt{see(me,black sphere)}} &
\tiny{\texttt{see(me,opponent)}} &
\tiny{\texttt{see(me,purple cube)}} &
\\
\tiny{\texttt{see(me,purple pyramid)}} &
\tiny{\texttt{see(me,purple sphere)}} &
\tiny{\texttt{see(me,yellow cube)}} &
\tiny{\texttt{see(me,yellow pyramid)}} &
\\
\tiny{\texttt{see(me,yellow sphere)}} &
\tiny{\texttt{see(opponent,black cube)}} &
\tiny{\texttt{see(opponent,black pyramid)}} &
\tiny{\texttt{see(opponent,black sphere)}} &
\\
\tiny{\texttt{see(opponent,me)}} &
\tiny{\texttt{see(opponent,purple cube)}} &
\tiny{\texttt{see(opponent,purple pyramid)}} &
\tiny{\texttt{see(opponent,purple sphere)}} &
\\
\tiny{\texttt{see(opponent,yellow cube)}} &
\tiny{\texttt{see(opponent,yellow pyramid)}} &
\tiny{\texttt{see(opponent,yellow sphere)}} &
\tiny{\texttt{see(purple cube,me)}} &
\\
\tiny{\texttt{see(purple cube,opponent)}} &
\tiny{\texttt{see(purple cube,purple pyramid)}} &
\tiny{\texttt{see(purple cube,purple sphere)}} &
\tiny{\texttt{see(purple cube,yellow cube)}} &
\\
\tiny{\texttt{see(purple cube,yellow pyramid)}} &
\tiny{\texttt{see(purple cube,yellow sphere)}} &
\tiny{\texttt{see(purple pyramid,me)}} &
\tiny{\texttt{see(purple pyramid,opponent)}} &
\\
\tiny{\texttt{see(purple pyramid,purple sphere)}} &
\tiny{\texttt{see(purple pyramid,yellow cube)}} &
\tiny{\texttt{see(purple pyramid,yellow pyramid)}} &
\tiny{\texttt{see(purple pyramid,yellow sphere)}} &
\\
\tiny{\texttt{see(purple sphere,me)}} &
\tiny{\texttt{see(purple sphere,opponent)}} &
\tiny{\texttt{see(purple sphere,yellow cube)}} &
\tiny{\texttt{see(purple sphere,yellow pyramid)}} &
\\
\tiny{\texttt{see(purple sphere,yellow sphere)}} &
\tiny{\texttt{see(yellow cube,me)}} &
\tiny{\texttt{see(yellow cube,opponent)}} &
\tiny{\texttt{see(yellow cube,yellow pyramid)}} &
\\
\tiny{\texttt{see(yellow cube,yellow sphere)}} &
\tiny{\texttt{see(yellow pyramid,me)}} &
\tiny{\texttt{see(yellow pyramid,opponent)}} &
\tiny{\texttt{see(yellow pyramid,yellow sphere)}} &
\\
\tiny{\texttt{see(yellow sphere,me)}} &
\tiny{\texttt{see(yellow sphere,opponent)}} &
\tiny{\texttt{on(black cube,brown floor)}} &
\tiny{\texttt{on(black cube,olive floor)}} &
\\
\tiny{\texttt{on(black cube,orange floor)}} &
\tiny{\texttt{on(black cube,blue floor)}} &
\tiny{\texttt{on(black cube,grey floor)}} &
\tiny{\texttt{on(black cube,white floor)}} &
\\
\tiny{\texttt{on(black pyramid,brown floor)}} &
\tiny{\texttt{on(black pyramid,olive floor)}} &
\tiny{\texttt{on(black pyramid,orange floor)}} &
\tiny{\texttt{on(black pyramid,blue floor)}} &
\\
\tiny{\texttt{on(black pyramid,grey floor)}} &
\tiny{\texttt{on(black pyramid,white floor)}} &
\tiny{\texttt{on(black sphere,brown floor)}} &
\tiny{\texttt{on(black sphere,olive floor)}} &
\\
\tiny{\texttt{on(black sphere,orange floor)}} &
\tiny{\texttt{on(black sphere,blue floor)}} &
\tiny{\texttt{on(black sphere,grey floor)}} &
\tiny{\texttt{on(black sphere,white floor)}} &
\\
\tiny{\texttt{on(me,brown floor)}} &
\tiny{\texttt{on(me,olive floor)}} &
\tiny{\texttt{on(me,orange floor)}} &
\tiny{\texttt{on(me,blue floor)}} &
\\
\tiny{\texttt{on(me,grey floor)}} &
\tiny{\texttt{on(me,white floor)}} &
\tiny{\texttt{on(brown floor,opponent)}} &
\tiny{\texttt{on(brown floor,purple cube)}} &
\\
\tiny{\texttt{on(brown floor,purple pyramid)}} &
\tiny{\texttt{on(brown floor,purple sphere)}} &
\tiny{\texttt{on(brown floor,yellow cube)}} &
\tiny{\texttt{on(brown floor,yellow pyramid)}} &
\\
\tiny{\texttt{on(brown floor,yellow sphere)}} &
\tiny{\texttt{on(olive floor,opponent)}} &
\tiny{\texttt{on(olive floor,purple cube)}} &
\tiny{\texttt{on(olive floor,purple pyramid)}} &
\\
\tiny{\texttt{on(olive floor,purple sphere)}} &
\tiny{\texttt{on(olive floor,yellow cube)}} &
\tiny{\texttt{on(olive floor,yellow pyramid)}} &
\tiny{\texttt{on(olive floor,yellow sphere)}} &
\\
\tiny{\texttt{on(opponent,orange floor)}} &
\tiny{\texttt{on(opponent,blue floor)}} &
\tiny{\texttt{on(opponent,grey floor)}} &
\tiny{\texttt{on(opponent,white floor)}} &
\\
\tiny{\texttt{on(orange floor,purple cube)}} &
\tiny{\texttt{on(orange floor,purple pyramid)}} &
\tiny{\texttt{on(orange floor,purple sphere)}} &
\tiny{\texttt{on(orange floor,yellow cube)}} &
\\
\tiny{\texttt{on(orange floor,yellow pyramid)}} &
\tiny{\texttt{on(orange floor,yellow sphere)}} &
\tiny{\texttt{on(purple cube,blue floor)}} &
\tiny{\texttt{on(purple cube,grey floor)}} &
\\
\tiny{\texttt{on(purple cube,white floor)}} &
\tiny{\texttt{on(purple pyramid,blue floor)}} &
\tiny{\texttt{on(purple pyramid,grey floor)}} &
\tiny{\texttt{on(purple pyramid,white floor)}} &
\\
\tiny{\texttt{on(purple sphere,blue floor)}} &
\tiny{\texttt{on(purple sphere,grey floor)}} &
\tiny{\texttt{on(purple sphere,white floor)}} &
\tiny{\texttt{on(blue floor,yellow cube)}} &
\\
\tiny{\texttt{on(blue floor,yellow pyramid)}} &
\tiny{\texttt{on(blue floor,yellow sphere)}} &
\tiny{\texttt{on(grey floor,yellow cube)}} &
\tiny{\texttt{on(grey floor,yellow pyramid)}} &
\\
\tiny{\texttt{on(grey floor,yellow sphere)}} &
\tiny{\texttt{on(white floor,yellow cube)}} &
\tiny{\texttt{on(white floor,yellow pyramid)}} &
\tiny{\texttt{on(white floor,yellow sphere)}} &
\\
\end{tabular}
\caption{List of all atomic predicates used in the current iteration of XLand. It consists of 212 elements that relate 2 players, 9 movable objects and 5 floors through use of 4 relations. The slab object is never used in atomic predicates.}
\label{tab:atoms}
\end{table*}

\subsubsection{Generating games}
A naive way of generating a game could involve sampling a set of 6 atomic predicates, and then sampling a random matrix $\{-1,0,1\}^{3 \times 6}$ where rows are options, each made up of conjunctions of predicates, with at most 3 non zero entries per row. However, as seen in \figref{fig:uniformgames}
 this creates an extremely non-uniform distribution with respect to balance/competitiveness. In fact it is almost impossible to randomly sample a fully competitive game.

\begin{figure*}[t]
\includegraphics[width=\textwidth]{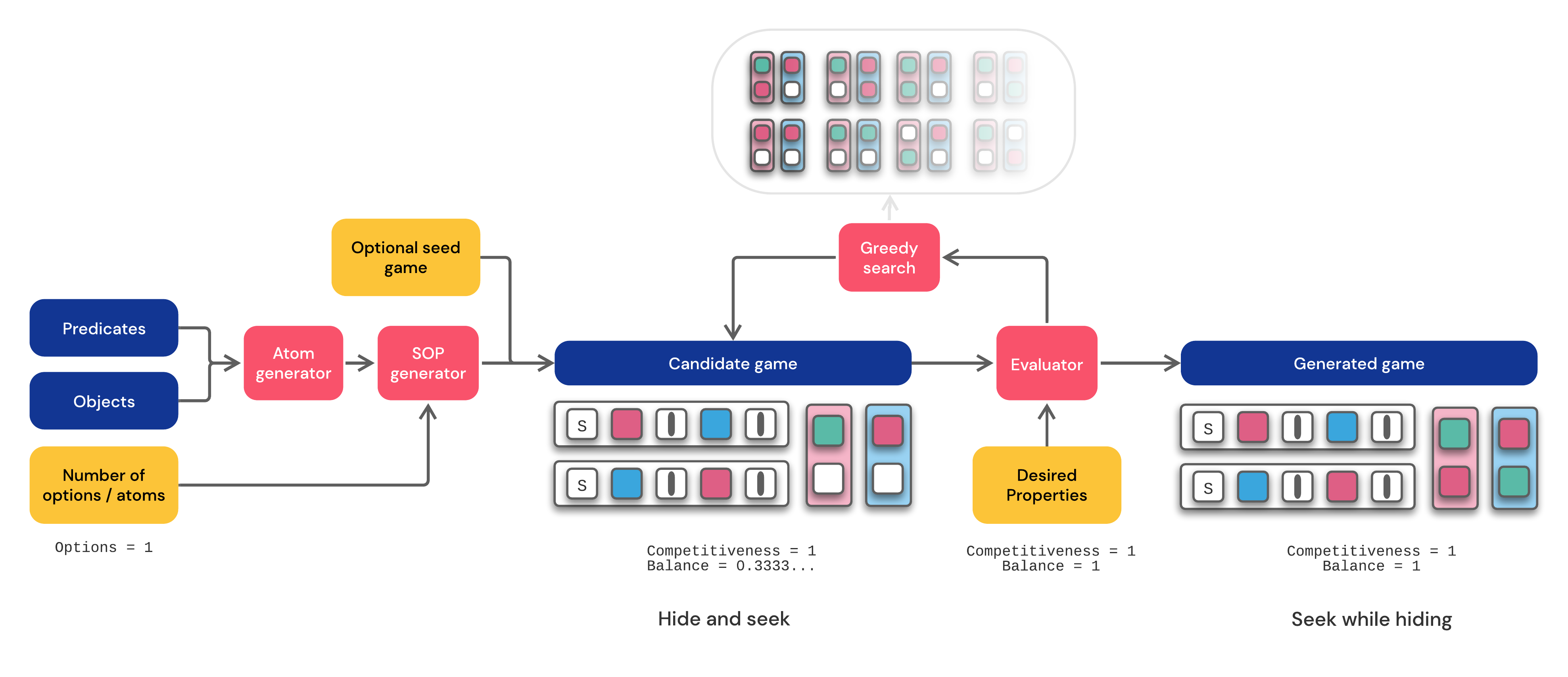}
\caption{
The process of generating of a game with target properties. We see the matrix representation, as well as relations. Greedy local search performs simple modifications to the game matrix, as described in \secref{app:gamespace}.
}
\label{fig:game_generation}
\end{figure*}

Consequently we rely on a local search methodology of game generation (see \figref{fig:game_generation}). In order to generate a game with a given number of options, conjunctions and target balance $b^*$, and competitiveness $c^*$:
\begin{itemize}
    \item We sample a random game matrix with a specified number of options and conjunctions
    \item We sample 3 unique atomic predicates $\predicate_1,\predicate_2,\predicate_3$ , and then sample recolouring $\xi$ and put 
    $
    \predicate_4:=\xi(\predicate_1),\predicate_5:=\xi(\predicate_2),\predicate_6:=\xi(\predicate_3)$. 
    \item If some predicates are identical then we keep resampling $\xi$. This always terminates as there always exists a recolouring that creates 3 new predicates.
    \item We combine matrix and predicates to get $\game$
    \item We compute $b := \balance(\game)\;\;\; c := \comp(\game)$
    \item For a fixed number of iterations, or until $b = b^* \wedge c = c^*$ we:
    \begin{itemize}
        \item Define improvement $\mathrm{imp}(\goal') := \min\{
        |b - b^*| - |\balance(\goal') - b^*|,
        |c - c^*| - |\comp(\goal') - c^*| \}$
        \item We randomly select a goal $\goal$ from $\game = (\goal, \goal')$
        \item We perform a local modification of the goal by trying to change matrix representation of this goal:
        \begin{itemize}
            \item[$\cdot$] Flip one of the 1s to -1s or vice verse
            \item[$\cdot$] Add 1 or -1 in a random option (if it would not invalidate the limit of max 3 conjunctions) 
            \item[$\cdot$] Remove 1 or -1 in a random option (if it would not zero out the option)
            \item[$\cdot$] Copy an option from $\goal'$
            \item[$\cdot$] Copy an option from $\goal'$ and negate it (multiply by -1)
        \end{itemize}
        \item For each of such modifications $\goal''$ we first verify that the game is not trivial, and that no 2 options are repeated, and then compute $\mathrm{imp}(\goal'')$.
        \item If at least one modification led to non-negative improvement ($\goal^*$), we pick the highest one and construct corresponding $\game = (\goal', \goal^*)$, recompute $b$ and $c$ and go to the next iteration.
        \item If all improvement were negative we terminate.
        \end{itemize}
\end{itemize}
The whole process is repeated 10 times and the best game selected (in terms of distance to $b^*, c^*$). Of course this process does not guarantee convergence to the exact value of $b^*$ and $c^*$, in particular some are impossible -- for example, there is no 1 option, 1 predicate game of competitiveness $0.25$ and balance $0.75$. Since generating a game with given properties is expensive, we also utilise the ability to create multiple games with the same characteristics described in the next section.

\subsubsection{Creating alike games}
Given a game $\game$ it is very easy to create another game of exactly the same number of options, conjunctions, and same balance and competitiveness. We randomly choose one of the bijective recolourings $\xi$ of objects, e.g.
\begin{equation*}
\begin{aligned}
    \xi(\texttt{black sphere}) := & \texttt{black sphere}\\
    \xi(\texttt{purple sphere}) := & \texttt{yellow sphere}\\
    \xi(\texttt{yellow sphere}) := & \texttt{purple sphere}\\
    \xi(\texttt{black pyramid}) := & \texttt{black pyramid}\\
    \xi(\texttt{purple pyramid}) := & \texttt{purple pyramid}\\
    \xi(\texttt{yellow pyramid}) := & \texttt{yellow pyramid}\\
    \xi(\texttt{black cube}) := & \texttt{yellow cube}\\
    \xi(\texttt{purple cube}) := & \texttt{purple cube}\\
    \xi(\texttt{yellow cube}) := & \texttt{black cube}
    \end{aligned}
\end{equation*}
and return $\xi(\game)$. It is easy to verify that $\comp(\game) = \comp(\xi(\game))$ (since competitiveness does not depend on semantincs of predicates) and also $\balance(\game) = \balance(\xi(\game))$, since because $\xi \in \Xi$ is a bijection, there also exists $\xi^{-1} \in \Xi$ thus
$$
\balance(\xi(\game)) = \max_{\xi' \in \Xi} \coop(\xi'(\xi(\game))) \geq \coop(\xi^{-1}(\xi(\game))) = \balance(\game).
$$
And at the same time
$$
\balance(\game) = \max_{\xi' \in \Xi} \coop(\xi'(\game)) \geq \coop(\xi(\game)) = \balance(\xi(\game)).
$$
consequently
$$
\balance(\game) =  \balance(\xi(\game)).
$$
Unfortunately, this process does not guarantee that $\xi(\game) \neq \game$ as the recolouring might not affect the game, or might just happen to recolour symmetric parts of the game. In practise, we repeat this process until a new game is found, and terminate it after 100 unsuccessful tries (\emph{e.g.} note that the game of hide and seek will never produce any new games with this method as it does not contain any objects).

\subsubsection{Generation of a 3 player game}
For simplicity, 3 player games are generated in the following way:
\begin{itemize}
    \item We take a 2 player game $(\goal_1, \goal_2)$
    \item We create $\goal_3$ by randomly mixing options from $\goal_1$ and $\goal_2$, and negate every predicate in a given option with 50\% too. 
    \item We randomly permute agents (and corresponding goals).
\end{itemize}


\subsubsection{PCA projection}
In order to obtain a PCA over the game space, we note that 
\begin{equation*}
\begin{aligned}
 \| \game_i - \game_j \|_\gamespace^2 &=  \tfrac{1}{4n^2\cdot N_\predicate^2}\| \gimel(\game_i) - \gimel(\game_j) \|^2 \\ &= \tfrac{1}{4n^2\cdot N_\predicate^2} \left [ \|\gimel(\game_i)\|^2 + \|\gimel(\game_j)\|^2 - 2\langle \gimel(\game_i), \gimel(\game_j)\rangle \right ],
\end{aligned}
\end{equation*}
where $\gimel(\goal)$ is a mapping that outputs a vector of length $N_\predicate$, where $i$th dimension equals 1 if $i$th valuation of predicates is rewarding under $\goal$, and -1 otherwise, and $\gimel(\game) : \gamespace \rightarrow \{-1,1\}^{n \cdot N_\predicate}$ is simply concatenation  of  $\gimel(\goal)$ over goals of the game.
With this construction, the norm of each vector is constant $\|\gimel(\cdot)\|^2 = n\cdot N_\predicate$ and thus
we can compute a $\gimel$-kernel~\citep{scholkopf2001kernel}:
$$
\mathrm{K}_\gimel(\game_i, \game_j) \propto -\| \game_i - \game_j \|_\gamespace^2
$$

This way we can utilise any kernelised method (such as Kernel PCA~\citep{scholkopf1998nonlinear}) to analyse the Game Space without explicitly forming exponentially large $\gimel(\gamespace)$ space (which in our case has over $6 \cdot 10^{65}$ dimensions) and see linear structures in the space in which $\| \cdot \|_\gamespace$ is just the Euclidean distance.

\subsection{Holding out tasks from training}
\label{sec:holdout}
In order to create hold-out games we simply create a single set of unique games (in the sense that no two games $\game \equiv \game'$ even under any recolouring $\xi$), split them between \evalvalid{} and \evaltrain{}. Training games generated are online rejected if there is any collision (including under recolouring) with \evalvalid{} or \evaltrain{}.

For hold-out worlds similarly a set of unique worlds was created (\secref{sec:evalset}) and split between \evalvalid{} and \evaltrain{}. Training worlds generated are online rejected if there is any collision with \evalvalid{} or \evaltrain{}.

Finally, \evaltrain{} and \evalvalid{} task sets are composed by random matching of corresponding world and games. This means that there is no game, nor world (and consequently no task) shared between the two sets or encountered during dynamic training task generation.

\subsection{Reinforcement Learning}
\label{app:rl}
We use V-MPO~\citep{vmpo} with hyperparameters provided in Table~\ref{tab:vmphypers}. We use a batch size of 64 and unroll length 80.
\begin{table}[t]
    \centering
    \begin{tabular}{ccc}
    \toprule
         Name & Value & Evolved with PBT \\
         \midrule
         discount ($\gamma$) & 0.99 & no \\
         learning rate & 1e-4 & yes\\
         baseline cost & 1 & no \\
         target update period & 50 & no \\
         $\epsilon_\alpha$ & 0.01252 & yes \\
         $\epsilon_\mathrm{temp}$ & 0.1 & no\\
         init $\alpha$ & 5 & no\\
         init temp & 1 & no \\
         top k fraction & 0.5 & no\\
         \midrule
         $m_>$ & 7.5 & yes\\
         $m_s$ & 375 & yes\\
         $m_\mathrm{cont}$ & 0 & yes\\
         $m_{>\mathrm{cont}}$ & 0 & yes \\
         $m_\mathrm{solved}$ & 1 & yes\\
         \bottomrule
    \end{tabular}
    \caption{V-MPO and DTG hyperparameters. The hyperparameters that have "yes" in the last column are adjusted using PBT (see~\secref{app:pbt}).}
    \label{tab:vmphypers}
\end{table}
\begin{table*}[t]
    \centering
    \begin{tabular}{ccc}
    \toprule
        Name & Type & Description \\
    \midrule
    \texttt{ACCELERATION} & $\mathbb{R}^3$ & Acceleration in each of 3 axes.\\
    \texttt{RGB} & $[0,255]^{72 \times 96 \times 3}$ & Pixel observations in RGB space with 72 by 96 pixel resolution.\\
    \texttt{HAND IS HOLDING} & $\{0,1\}$ & Flag whether an item is held.\\
    \texttt{HAND DISTANCE} & $\mathbb{R}$ & A distance to a held object if one is held.\\
    \texttt{HAND FORCE} & $\mathbb{R}$ & A force applied to an item held by an agent. \\
    \texttt{LAST ACTION} & $(\mathbf{a}_{t-1,i})_{i=1}^{10}$ & A 10-hot encoding of the previously executed action.\\
    \texttt{GOAL MATRIX} & $\{-1,0,1\}^{6 \times 6}$ & a matrix encoding a goal in DNF form.\\
    \texttt{GOAL ATOMS} & $\mathbb{N}^{6 \times 6}$ & a matrix encoding corresponding atoms from the goal, provided as categoricals.\\
         \bottomrule
    \end{tabular}
    \caption{List of all observations an agent receives as part of $\observation_{t}^{i} := (f_i(\state_t), \goal_i)$. Note, that this does not contain a reward, as the agent policy $\policy$ does not have access to this information, only the value head does during training.}
    \label{tab:obsevations}
\end{table*}

\begin{table*}[t]
\begin{tabular}{ccc}
\toprule
    Name & Possible values & Description  \\
\midrule
\texttt{MOVE FORWARD BACK} &  $\{-1,0,1\}$ & Whether to apply forward/backward or no movement.\\
\texttt{MOVE LEFT RIGHT} &  $\{-1,-0.05,0,0.05,1\}$ & Strafing.\\
\texttt{LOOK LEFT RIGHT} &  $\{-1,-0.2,-0.05,0,0.05,0.2,1\}$ & Left/right rotation.\\
\texttt{LOOK UP DOWN} &  $\{-1,-0.03,0,0.03,1\}$ & Up/down rotation.\\
\texttt{GRAB} &  $\{0,1\}$ & Grab an object.\\
\texttt{USE GADGET} &  $\{0,1\}$ & Use currently equipped gadget.\\
    \bottomrule
\end{tabular}
\caption{The structure of the decomposed action space, consisting of 6 dimensions of discrete actions. Every combination of the values is feasible, leading to 2100 possible actions.}
\label{tab:actionset}
\end{table*}

\subsection{Distillation}
\label{app:distillation}
We follow on-policy distill~\citep{czarnecki2020}, which for a teacher $\policy^\mathrm{teacher}$ and student $\policy$ defines the per timestep $t$ auxiliary loss
$$
\ell_t^\mathrm{distill} := \mathrm{KL} \left[ \stopgradient{ \policy^\mathrm{teacher}_t} \middle \| \policy_t \right ].
$$
In order to further exploit the fact that our goal is not to replicate the teacher but rather use to bootstrap from and provide assistance with exploration, we mask this loss over states when the reward is obtained. In other words, we distill on a timestep if and only if an agent is coming from a non-rewarding state, thus defining \emph{exploration distillation} loss
$$
\ell_t^\mathrm{exp-distill} := (1 - \reward_{t-1}) \cdot \ell_t^\mathrm{distill}.
$$
Note we have binary 0 or 1 rewards $\reward_{t-1}$ only in this work. An agent uses a weight of 4 of this loss for the first 4 billion steps in every generation apart from the first generation (since there is no initial teacher), and the weight changes to 0 afterwards.

\subsection{Network architecture}
\label{app:network}
Unless otherwise specified, every MLP uses the ReLU activation function.
We use Sonnet~\citep{sonnet_cited} implementations of all neural network modules, and consequently follow Sonnet default initialisation schemes as well.

\begin{figure*}[t]
\includegraphics[width=\textwidth]{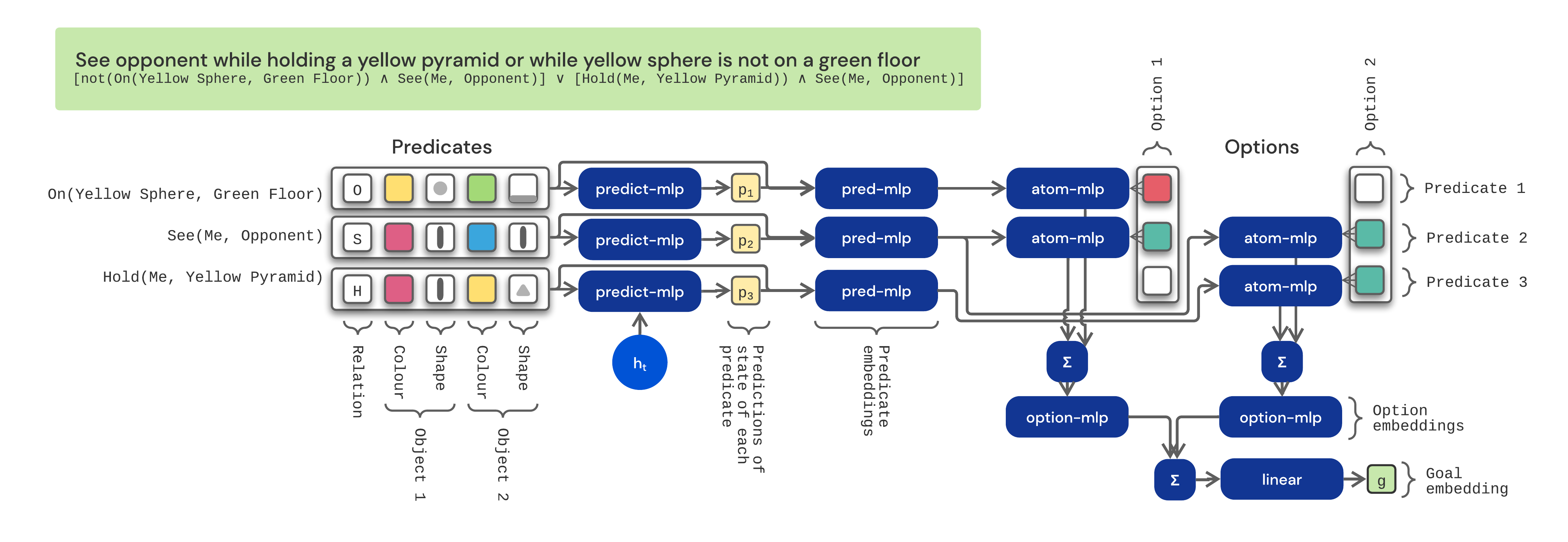}
\caption{The architecture of the goal embedding/prediction modules. Atomic predicates are provided in a 5-hot encoded fashion, since all the relations used take two arguments, each of which can be decomposed to a colour and shape. For player object we simply have a special colour "me" and "opponent". Details of the architecture are provided in~\secref{app:network}.}
\label{fig:goal_network}
\end{figure*}

\paragraph{Torso.} An $\texttt{RGB}$ observation (see Table~\ref{tab:obsevations} for details) is passed to a ResNet~\citep{he2016deep} torso with $[16,32,32]$ channels, each consisting of 2 blocks and output size of 256. Max pooling is used, as well as scalar residual multiplier. Torso produces $\hat{\mathbf{o}}_t$.

\paragraph{Goal embedding.} The goal embedding network is presented in \figref{fig:goal_network}. The predicate embedding network uses a linear layer with 256 outputs. The atom-mlp is a 2-hidden layer MLP with 256 hidden units. The option-mlp is a 2-hidden layer MLP with 256 hidden units.  The final goal embedding is 256 dimensional. We use summation to aggregate across conjunctions in the option, as well as options in the goal. This processing is akin to using a Graph Neural Network~\citep{battaglia2018relational}, where there is one vertex for a goal, connected to one vertex per option, which has edges to predicate vertices, labeled with either 1 or -1 (to denote negation), and each predicate vertex is labeled with its value.

\paragraph{LSTM.} We use a 2 layer LSTM core with 256 neurons each, and skip connections. The LSTM takes as input $\hat{\mathbf{o}}_t$ and produces $\mathbf{h}_t$.

\paragraph{Atom predictor.} The atom predictor uses inputs $\mathbf{h}_t$ and atom embedding $\mathbf{e_a}_t$, concatenated and tiled to form a tensor of size $[\mathrm{num\_atoms}, \mathrm{atom\_features + hidden\_size}]$. We apply a small MLP with 2 hidden layers of sizes 64 and 32 and one output neuron per atom predicted, forming $\mathbf{p}_t$ of size $\mathrm{num\_atoms}$.

\paragraph{GOAT module.} The GOAT module takes $\mathbf{h}_t$, $\mathbf{p}_t$ and $\goal$ as inputs.
In order to apply attention to a flat output of an LSTM, we first linearly upscale it to 1024 dimensions, and then reshape to a 32 by 32 matrix (and analogously the goal embedding is reshaped to have 32 features) and then it is passed to an attention module with a single head, and key and value sizes set to 128. We use a temperature of 0.1 and softmax mixing in the GOAT module. The internal value heads of GOAT use an MLP with 256 hidden units. Note, that these do not accept last reward as input (as opposed to the external value head), since these value heads are used during inference and affect the policy output -- our agent's policy is not conditioned on reward.
The GOAT module produces $\widehat{\mathbf{h}}_t$ as an output.

\paragraph{PopArt.} The PopArt value head~\citep{popart1, hessel2019multi} uses 256 hidden units.

\paragraph{Policy head.} The policy head is a simple MLP, applied to $\widehat{\mathbf{h}}_t$, with 256 hidden units, and 6 linear, softmaxed heads $(\policy_t)_k$, one per action group (see Table~\ref{tab:actionset} for details). We follow a decomposed action space, meaning that we assume the value of each action group is conditionally independent given the current state:
$$
\policy_t[\mathbf{a}_t] := \mathrm{Pr}[\mathbf{a}_t=(\mathbf{a}_t^1, \dots, \mathbf{a}_t^6)] := \prod_{k=1}^6 (\policy_t)_k[\mathbf{a}_t^k] ,
$$
which allows us to efficiently parameterise the joint distribution over the entire space of 2100 actions~\citep{jaderberg2019human}.

\paragraph{External Value Head.} The value head used for RL algorithms, takes, in addition to $\widehat{\mathbf{h}}_t$ the last reward $r_{t-1}$ (concatenated), and produces the value prediction by an MLP with 256 units.

\subsubsection{Auxiliary losses}
All auxiliary losses are computed by treating a trajectory as if it was a sequence-to-sequence problem, without taking into consideration the effects it has on the control policy~\citep{jaderberg2016reinforcement,czarnecki2019distilling}.

We do not weight nor tune auxiliary losses weights, they simply added up together, jointly with the updates coming from the RL system. 
Merging is done on the update level, meaning that RL updates do not take into consideration effect they have on auxiliary losses either.

\paragraph{GOAT.} There are two consistency losses $\ell^\mathbf{V}, \ell^\mathbf{h}$ coming from the GOAT architecture.

\paragraph{Atom prediction.} We use standard cross-entropy loss for multi-label classification.

\paragraph{External value function matching.} Internal value functions of GOAT have an additional alignment loss in the form of the L2 distance between the final GOAT value and the external value function. We do not stop gradients on external value function, meaning that they co-distill.

\subsection{Population Based Training}
\label{app:pbt}

After the initial period of 5e8 steps of guaranteed no evolution events, every 1e8 steps we check whether some agents should be evolved. A pair of agents $\policy_i$ and $\policy_j$ are considered eligible for evolution from $\policy_j$ (parent) to $\policy_i$ (child) if and only if:
\begin{itemize}
    \item $\policy_i$ did not undergo evolution in last 2.5e8 steps,
    \item $\forall_{k \in \{10,20,50\}} \mathrm{perc}(\policy_j|\population_t)[k]  \geq \mathrm{perc}(\policy_i|\population_t)[k]$,
    \item $\mathrm{score}_{ij} := \tfrac{\sum_{k \in \{10,20,50\}} \mathrm{perc}(\policy_j|\population_t)[k]
    }{
    \sum_{k \in \{10,20,50\}} \mathrm{perc}(\policy_i|\population_t)[k]
    } > 1.01.
    $
\end{itemize}
Next, we take a pair where $\mathrm{score}_{ij}$ is maximum and perform the evolution operation. This means that weights and hyperparameters of $\policy_i$ are copied from $\policy_j$, and for each of the hyperparameters, independently with 50\% chance we apply following mutation:
\begin{itemize}
    \item V-MPO hyperparameter $\epsilon_\alpha$ is multiplied or divided by 1.2 (with 50\% chance).
    \item learning rate is multiplied or divided by 1.2 (again with 50\% chance)
    \item $m_\mathrm{cont}$ is increased or decreased by 7.5 (again with 50\% chance), and clipped to $(0,900)$ (range of the return).
    \item $m_\mathrm{cont}$ is multiplied or divided by 1.1 (again with 50\% chance), and clipped to $(0,900)$ (range of the return).
    \item $m_>$ is multiplied or divided by 1.2 (again with 50\% chance), and clipped to $(0,900)$ (range of the return).
    \item $m_\mathrm{solved}$ and $m_{>\mathrm{cont}}$ is increased or decreased by 0.1 (again with 50\% chance), and clipped to $(0,1)$ (corresponding to possible values of our MC estimator with 10 samples).
\end{itemize}
For efficiency reasons, the child agent also inherits the parent's DTG training task set.

\subsection{GOAT}

\paragraph{Placement of the GOAT module.} The GOAT module is placed after the recurrent core, and lacks recurrence itself. The reason is to be able to query \emph{"what would happen if the policy's goal was different"} without having to unroll entire trajectory with a separate goal. In principle one could have $n_o+1$ copies of the agent unrolled over the experience, one conditioned on the full goal, and the remaining $n_o$ on corresponding options. The full goal conditioned unroll would generate the actions and experience, whilst the remaining unrolls would be off-policy. Placing the GOAT module post recurrence avoids this redundancy, and actually forces the agent to explicitly split it's reasoning into goal conditioned and goal invariant (recurrent) parts. Arguably this is also responsible for creating an agent that is capable of easily reacting to a new goal/change of goals mid-episode, despite not being trained to do so.

\paragraph{Feedback loop that affects the policy.} 
Let us imagine that $\mathbf{v}_t^{[1]} > \mathbf{v}_t^{[0]}$, meaning that option 1 has a higher value than the whole game. If this is correct, meaning that $\mathbf{v}_t \approx \mathbf{V}_\pi(\state_t)$ this means that an agent could improve its policy if it was to act as if its goal was changed purely to option 1. Consequently the loss $\ell^\mathbf{h}$, which aligns the internal state with the internal state corresponding to following just option 1, can be seen as a step of policy improvement. We are working under an assumption that $f_\policy$ is a smooth L-Lipschitz function, meaning that for some (hopefully small $L>0$) $\| f_\policy(\widehat{\mathbf{h}}) - f_\policy(\widehat{\mathbf{h}'}) \| \leq L \| \widehat{\mathbf{h}} - \widehat{\mathbf{h}'} | $. One could also add an explicit loss aligning policies themselves, however this would require balancing a loss over distributions over actions (e.g. KL) with the loss over hidden activations (L2). For simplicity we are not doing this in this work.

\subsection{Multi-agent analysis}
\label{app:ma}
For multi-agent analysis we took the final generation of the agent (generation 5) and created equally spaced checkpoints (copies of the neural network parameters) every 10 billion steps, creating a collection of 13 checkpoints.

\subsubsection{Hide and seek}
We the following definition of hide and seek
\begin{equation*}
\begin{aligned}
\goal_1 &:= &&\texttt{see(me,opponent)}\\
\goal_2 &:= &&\texttt{not(see(opponent,me))}\\
\end{aligned}
\end{equation*}
We collected 1000 \evalvalid{} worlds and ran 1 episode per matchup (agent pair) per world.

\subsubsection{Conflict Avoidance}
The game is defined as
\begin{equation*}
\begin{aligned}
\goal_1 &:= &&\texttt{on(purple sphere, orange floor)} \vee \\  &&&\texttt{on(yellow sphere, orange floor)}\\
\goal_2 &:= &&\texttt{on(yellow sphere, grey floor)}\\
\end{aligned}
\end{equation*}
Experiments were conducted on a single world, where agents spawn equally distant to both spheres, and thus their choice of which one to use is not based on distance heuristics they might have learned. They can also see all the target floors initially thus there is no exploration needed.
We collected 1000 episodes per matchup (agent pair).

\subsubsection{Encoding Chicken in Xland}
We encode social dilemmas in Xland using task predicates such that the game options can be grouped into \emph{meta-actions} and the social dilemma rewards correspond to option \emph{multiplicity} that the player has given the opponent meta-action. Table~\ref{tab:chicken_encoding} represents the encoding for the social dilemma game of \emph{Chicken}. Here, the payoff for the meta-actions \emph{Cooperate} and \emph{Defect} are given. The choice of the predicates makes it impossible for both the players to satisfy options corresponding to the Defect meta-action simultaneously (hence the meta-action joint reward is $(0,0)$ for the Defect-Defect joint meta-action), whereas the options corresponding to Cooperate are compatible with any option. In our experiments we used options of same the predicate length and type so that they are similar in complexity to satisfy. The exact game is given as follows:
\begin{equation*}
\begin{aligned}
\widehat{\goal}_\mathrm{c} :=& \texttt{see(opponent,yellow pyramid)} \wedge \\
&\texttt{see(yellow cube,yellow sphere)}\wedge\\
&\texttt{not(see(black pyramid,purple pyramid))}\\
\widehat{\goal}_\mathrm{d_1} :=& \texttt{see(me,yellow pyramid)}\wedge \\
&\texttt{see(black cube,black sphere)}\wedge\\
&\texttt{not(see(opponent,yellow pyramid))}\\
\widehat{\goal}_\mathrm{d_2} :=& \texttt{see(me,yellow pyramid)}\wedge \\
&\texttt{see(purple cube,purple sphere)} \wedge\\
&\texttt{not(see(opponent,yellow pyramid))}\\
\goal_1 := &\widehat{\goal}_\mathrm{c} \vee \widehat{\goal}_\mathrm{d_1} \vee \widehat{\goal}_\mathrm{d_2}\\
\goal_2 :=&\widehat{\goal}_\mathrm{c} \vee \widehat{\goal}_\mathrm{d_1} \vee \widehat{\goal}_\mathrm{d_2}\\
\end{aligned}
\end{equation*}
\begin{table}[t]
    \centering
    \begin{tabular}{ccc}
    \toprule
         Chicken & Cooperate & Defect  \\
         \midrule
         Cooperate &  1,1 & \textbf{1,2} \\
         Defect &  \textbf{2,1} & 0,0 \\
        \bottomrule
    \end{tabular}
    \\
    \begin{tabular}{cc}
    \toprule
         $\goal_1$&  $\goal_2$ \\
         \midrule
         $\predicate_1 \wedge \neg \predicate_5 \wedge \predicate_4
         $
         &
         $\predicate_1 \wedge  \predicate_5 \wedge \neg \predicate_4
         $
         \\
         $\predicate_2 \wedge \neg \predicate_5 \wedge \predicate_4
         $
         &
         $\predicate_2 \wedge  \predicate_5 \wedge \neg \predicate_4
         $ 
         \\
         $\predicate_3 \wedge  \predicate_5 \wedge \neg \predicate_6
         $
         &
         $\predicate_3 \wedge  \predicate_4 \wedge \neg \predicate_6
         $ \\
         \bottomrule
    \end{tabular}
    \caption{\textbf{(Top)} The Chicken social dilemma with Nash Equilibria highlighted in bold font. \textbf{(Bottom)} The corresponding option based encoding using task predicates. $\predicate_1, \predicate_2, \predicate_3$ are unique predicates and $\predicate_4, \predicate_5, \predicate_6$ represent conjunctions with non overlapping predicates.}
    \label{tab:chicken_encoding}
\end{table}
We collected 1000 \evalvalid{} worlds and run 1 episode per matchup (agent pair) per world.

\subsection{Handauthored levels}
We created a fixed set of hand-authored tasks as an additional static evaluation set. These are described in Table~\ref{tab:handauthoredlist} and Table~\ref{tab:handauthoredlist2}.

\begin{table*}[]
    \centering
    \begin{tabular}{lp{8cm}c}
    \toprule
    Name & Description & Agent return $>0$ \\
    \midrule
Capture The Cube & A competitive game where both players must bring the opponent's cube to their cube and base floor in a symmetrical world. & \checkmark \\
\midrule
Catch Em All & A cooperative game where both players must make 5 particular objects near each other, gathering them from across the world. & \checkmark \\
\midrule
Choose Wisely 2p & Each player has the same 3 options to choose from: one of 3 cubes to hold without the other player holding that cube. & \checkmark \\
\midrule
Choose Wisely 3p & The same as above with 3 players. & \checkmark \\
\midrule
Coop or Not & Both players have the one cooperative option and one competitive option to choose from. & \checkmark \\
\midrule
Find Cube & The player must find the cube to be near in a complex world. & \checkmark \\
\midrule
Find Cube With Teaser & The same as above, however the world allows the agent to see the cube but the world does not allow direct navigation to the cube. & $\times$ \\
\midrule
Hide And Seek: Hider  & Asymmetric hide and seek in a simple world with the agent playing the hider. & \checkmark \\
\midrule
Hide And Seek: Seeker  & Same as above but the agent playing the seeker. & \checkmark \\
\midrule
Hold Up High & Both players must hold the yellow pyramid on the highest floor. & \checkmark \\
\midrule
King of the Simplest Hill & Both players must be on the top floor of the world without the other player being on that floor. & \checkmark \\
\midrule
King of The Hill & Same as above but with a more complex world. & \checkmark \\
\midrule
Keep Tagging & The player must stop the other player holding objects or touching a floor. & \checkmark \\
\midrule
Make Follow Easy & The agent must lead the other player (whose policy is to follow) to a particular floor colour. & $\times$ \\
\midrule
Make Follow Hard & Same as above however the target floor is higher and has a smaller area. & $\times$ \\
\midrule
Mount Doom & In a world with a large central mountain, the agent must get to the top without the other player getting to the top. & \checkmark \\
\midrule
Mount Doom 2 & Same as above but the other player starts at the top. & \checkmark \\
\midrule
Navigation With Teaser & Similar to Find Cube With Teaser, however the agent must navigate to a target floor rather than the cube.  & \checkmark \\
\midrule
Nowhere To Hide & The players start on opposite towers, and the agent must stop the other player (noop policy) from touching the tower floor. & \checkmark \\
\midrule
Object Permanence Black Cube & The agent starts and can see a yellow cube on the left and a black cube on the right. The agent must choose which path to take (which means the agent loses sight of the cubes) to reach the target cube (black in this case). & \checkmark \\
\midrule
Object Permanence Yellow Cube & Same as above but the target cube is yellow. & \checkmark \\
\midrule
One Pyramid Capture The Pyramid & Same world as Capture the Cube, however both players must take the single yellow pyramid to their base floor. & \checkmark \\
\midrule
Race To Clifftop With Orb  & Both players must stand on the top of a cliff edge holding the yellow sphere, without the other player standing on the cliff. & \checkmark \\
\bottomrule
\end{tabular}
\caption{List of hand authored tasks. The last column shows if the agent participates in the specific task (whether it ever reaches a rewarding state). Continues in Table~\ref{tab:handauthoredlist2}.}
\label{tab:handauthoredlist}
\end{table*}

\begin{table*}[]
    \centering
    \begin{tabular}{lp{8cm}c}
    \toprule
    Name & Description & Agent return $>0$ \\
\midrule
Ridge Fencing & Similar to King of the Hill, however the target floor is a narrow ridge. & \checkmark \\
\midrule
Sheep Herder & The agent must make the other player stand near a set of target objects. & $\times$ \\
\midrule
Solve AGI & The agent must take the black sphere and put it on the target floor against another player that is trying to oppose this. & \checkmark \\
\midrule
Stay Where You Spawn & The player doesn't have to do anything, it is rewarded if it stays in its initial position. & \checkmark \\
\midrule
Stop Rolling Freeze Gadget & In world composed of one big slope, the agent must stop the purple sphere from touching the bottom floor without holding it. & \checkmark \\
\midrule
Stop Rolling Tag Gadget & Same as above, except the agent has a tag gadget rather than freeze gadget. & \checkmark \\
\midrule
Stop Rolling Tag Gadget Easy & Same as above, except the slope is less steep. & \checkmark \\
\midrule
Tag Fiesta 2p & Both players have a goal to make the other player not touch any floors. All players have the tag gadget. & \checkmark \\
\midrule
Tag Fiesta 3p & Same as above but a 3-player version. & \checkmark \\
\midrule
Tool Use Climb 1 & The agent must use the objects to reach a higher floor with the target object. & $\times$ \\
\midrule
Tool Use Gap 1 & The agent must reach an object but there is a gap in the floor. & \checkmark \\
\midrule
Tool Use Gap 2 & Same as above but the gap is bigger. & $\times$ \\
\midrule
Who Gets The Block & Both players want to be holding the same cube but on different coloured floors. & \checkmark \\
\midrule
XFootball & Both players want the black sphere to be touching the floor at opposite ends of the world. & \checkmark \\
\midrule
XRPS Counter Black & A version of XRPS~\secref{sec:game-diversity} in which the agent can choose from three rock-paper-scissor like options. The other player always chooses to hold the black sphere. & \checkmark \\
\midrule
XRPS Counter Purple & Same as above but where the other player always chooses to hold the purple sphere. & \checkmark \\
\midrule
XRPS Counter Yellow & Same as above but where the other player always chooses to hold the yellow sphere. & \checkmark \\
\midrule
XRPS With Tag Gadget & Full rock-paper-scissor like game. & \checkmark \\
\midrule
XRPS 2 With Tag Gadget & Same as above but with different predicates representing the options. & \checkmark \\
\bottomrule
\end{tabular}
\caption{List of all hand authored tasks continued. The last column shows if the agent participates in the specific task (whether it ever reaches a rewarding state).}
\label{tab:handauthoredlist2}
\end{table*}

\subsection{Representation analysis}
\label{app:internals}

To conduct the representation analysis we gathered 3000 trajectories coming from \evalvalid{} tasks. Next, we randomly selected 60 timestamps $\hat{t}_i \in (1, 900)$ and used them to subsample the trajectories, resulting in 180,000 states $\state_i$ with corresponding recorded activations of $\widehat{h}_i$ (GOAT module), $\mathbf{h}_i$ (LSTM state) and the goal embedding.

We train a Kohonen Network composed of 900 neurons, arranged in a $30 \times 30$ grid covering a unit circle through a transformation of the lattice over $[-1, 1]^2$
$$
\mathfrak{k}(x,y) :=\left (x \cdot \sqrt{1-\tfrac{1}{2}y^2},
                      y \cdot \sqrt{1-\tfrac{1}{2}x^2} \right),
$$
for 5000 iterations, using stochastic gradient descent with batches of size 100 and an exponentially decaying learning rate $\mathrm{lr}_k := 0.1 \cdot \exp(1 - \tfrac{k}{5000})$. The initial neurons positions are given by the minibatch k-means clustering~\citep{lloyd1982least} using \texttt{k-means++} heuristic initialisation~\citep{arthur2006k}, batch size of 100 and 100 iterations. 

\begin{figure}[t]
    \centering
    \includegraphics[width=0.5\linewidth]{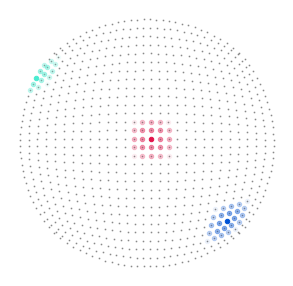}
    \caption{Visualisation of the Kohonen Network used in our analysis, composed of 900 Kohonen Neurons. Three neurons are called out in colour, with their receptive field (neurons that have non zero update weight) colour-coded with the colour intensity representing the weight.}
    \label{fig:kohonen_grid}
\end{figure}

We use $\mathrm{d}_\mathrm{max} = 3$, with the visualisation of the emerging Kohonen Network and local receptive fields shown in \figref{fig:kohonen_grid}.

We use the following definitions of properties of state $\state_t$ when playing with a goal $\goal$:
\begin{itemize}
    \item \textbf{Early in the episode}. True if and only if $t < \hat{t}_{30}$ (approximately half of the episode).
    \item \textbf{Agent is holding}. True if and only if $\state_t$ represents a situation in which the agent is holding an object.
    \item \textbf{High baseline}. True if and only if $\mathbf{v}_t > 7.5$ (after Popart normalisation).
    \item \textbf{Rewarding state}. True if and only if $\reward_\goal(\state_t) = 1$.
    \item \textbf{Knows it's rewarded}. True if and only if $\reward_\goal(\state_t) = 1$ and the active option has all its atoms predicted correctly.
    \item \textbf{Knows entire state}. True if and only if every atom prediction is correct.
    \item \textbf{One missing atom}. True if and only if $\reward_\goal(\state_t) = 0$ and there exists $\state'$ such that $\| \predicate(\state_t) - \predicate(\state') \| = 1$ and $\reward_\goal(\state') = 1$.
    \item \textbf{Many missing atoms}. True if and only if $\reward_\goal(\state_t) = 0$ and for every $\state'$ such that $\reward_\goal(\state') = 1$ we have $\| \predicate(\state_t) - \predicate(\state') \| > 1$.
    \item \textbf{One option}. Whether $\goal$ consists of a single option.
    \item \textbf{Many options}. Whether $\goal$ consists of more than one option.
\end{itemize}
When plotting a Kohonen map for a specific hidden state we subtract the average activity in our dataset and use colour using a diverging colour map. We do this for visual clarity to avoid displaying a pattern (bias) that is constantly there.

\paragraph{Kohonen Neurons}
Once the Kohonen Neuron $\mathfrak{h}$ for a property $p$ has been identified, we define the following classifier:
$$
\hat{p}(x) := \mathrm{KDE}[\mathfrak{h}_i: p(\state_i)](x) > \mathrm{KDE}[\mathfrak{h}_i: \neg p(\state_i)](x)
$$
where $\mathrm{KDE}[A](x)$ is the kernel density estimation density of a set $A$ evaluated at $x$. We use a Gaussian kernel and Silverman's rule to estimate its bandwidth~\citep{silverman2018density}.

\end{document}